\documentclass[fleqn,10pt]{wlscirep}

\usepackage{ctex}

\usepackage{graphicx}
\usepackage{amsmath}
\usepackage{multirow}
\usepackage{adjustbox}
\usepackage{subcaption}
\usepackage{pifont}
\usepackage{xr}
\usepackage{float}
\usepackage{nameref}
\usepackage{xcolor}
\usepackage{longtable}
\usepackage{booktabs}
\usepackage{array}
\usepackage{appendix}
\usepackage{hyperref}
\usepackage[nameinlink,noabbrev]{cleveref}
\usepackage{makecell}
\usepackage[table]{xcolor}
\usepackage{threeparttable}
\usepackage{diagbox}

\crefname{figure}{Fig.}{Figs.}
\crefname{table}{Table}{Tables}
\crefname{section}{Section}{Sections}
\crefname{appendix}{Supplementary}{Supplementary}

\title{Auditable agentic AI for evidence-grounded thyroid ultrasound diagnosis and reporting}

\author[1,$\dagger$]{Haifan Gong}
\author[1,$\dagger$]{Shiyu Chen}
\author[2,$\dagger$]{Bodong Wang}
\author[3,$\dagger$]{Yuqi Wang}
\author[4]{Shijie Wang}
\author[5]{Guoliang You}
\author[1]{Xinyu Xiong}
\author[6]{Haowei Wang}
\author[2]{Mingzhi Mao}
\author[4]{Dexing Kong}
\author[7,*]{Qinghua Liu}
\author[8,*]{Wei Lou}
\author[9,*]{Fei Chen}
\author[1,*]{Guanbin Li}

\affil[1]{School of Computer Science and Engineering, Sun Yat-sen University, Guangzhou, China}

\affil[2]{School of Software Engineering, Sun Yat-sen University, Zhuhai, China}

\affil[3]{Independent Researcher, Jersey City, NJ, USA}

\affil[4]{School of Mathematical Sciences, Zhejiang University, Hangzhou, China}

\affil[5]{Department of Radiology, Perelman School of Medicine, University of Pennsylvania, Philadelphia, PA, USA}

\affil[6]{Department of Pathology, Zhujiang Hospital, Southern Medical University, Guangzhou, China}

\affil[7]{Department of Health Management, Zhujiang Hospital, Southern Medical University, Guangzhou, China}

\affil[8]{College of Mathematical Medicine, Zhejiang Normal University, Jinhua, China}

\affil[9]{Department of Thyroid Surgery, Zhujiang Hospital, Southern Medical University, Guangzhou, China}

\affil[*]{\textbf{Corresponding authors:} Qinghua Liu~(\href{mailto:13760633321@163.com}{13760633321@163.com}), Wei Lou~(\href{mailto:louwei@zjnu.edu.cn}{louwei@zjnu.edu.cn}), Fei Chen~(\href{mailto:gzchenfei@126.com}{gzchenfei@126.com}), and Guanbin Li~(\href{mailto:liguanbin@mail.sysu.edu.cn}{liguanbin@mail.sysu.edu.cn})}
\affil[$\dagger$]{These authors contributed equally to this work}

\begin{abstract}
Thyroid ultrasound diagnosis requires coordinated lesion localization, measurement, risk stratification and reporting, yet most AI systems address these tasks in isolation and provide limited support for clinical review. We present ThyroidXAgent, a clinician-interactive agentic AI system that coordinates specialized diagnostic tools and stores their outputs as an auditable case-level evidence record. The system was developed using OpenThyroidDB, a multicentre, multitask resource integrating approximately 0.3 million ultrasound images and 24,000 paired reports, and was evaluated on 28,458 non-overlapping test cases, including 8,721 cases from 35 centres in the private NHC-MISD-TUS cohort. Across heterogeneous datasets, ThyroidXAgent achieved a mean Dice score of 87.21\% for nodule segmentation and a mean AUROC of 0.9466 for benign–malignant classification. The same workflow supported lymph-node metastasis prediction and follicular versus papillary thyroid carcinoma classification, with AUROCs of 0.864 and 0.805, respectively. For report generation, evidence-grounded assembly outperformed multimodal language-model baselines across three cohorts. ThyClinScore, a lesion-level clinical semantic metric introduced here, showed the strongest correlation with a location-aware language-model judge. ThyroidXAgent improved physician classification accuracy, increased report diagnostic consistency from 70.3\% to 86.2\%, and reduced segmentation and reporting time by 35.9\% and 27.4\%, respectively. These findings support auditable, clinician-correctable agentic AI for thyroid ultrasound diagnosis and reporting.
\end{abstract}

\begin{document}
\flushbottom
\maketitle

\section{Introduction}

Thyroid ultrasound diagnosis depends on a sequence of lesion-level observations and clinical decisions \cite{Alexander2022Lancet,Grani2024NatRevEndocrinol}. A clinically useful examination localizes and measures thyroid nodules \cite{Alexander2022Lancet,Grani2024NatRevEndocrinol}, characterizes sonographic features \cite{Grani2024NatRevEndocrinol,Tessler2017ACRTIRADS,Hoang2018AJRInterobserver}, assigns risk using systems such as the Thyroid Imaging Reporting and Data System (TI-RADS) \cite{Tessler2017ACRTIRADS}, determines whether fine-needle aspiration is indicated \cite{Alexander2022Lancet,Tessler2017ACRTIRADS}, integrates cytology when available \cite{Cibas2017Bethesda} and communicates the findings in a structured report \cite{Grani2024NatRevEndocrinol,Tessler2017ACRTIRADS}. Each step introduces variability. Descriptors such as margins and echogenic foci \cite{Tessler2017ACRTIRADS,Hoang2018AJRInterobserver} show substantial reader dependence, and small changes in these features can alter biopsy or follow-up recommendations \cite{Tessler2017ACRTIRADS,Hoang2018AJRInterobserver}. These requirements make thyroid ultrasound a workflow-level problem for clinical AI \cite{Topol2019NatMedHighPerformance,Rajpurkar2022NatMedAIHealth}: useful systems must support prediction, preserve the evidence behind each step \cite{Chen2022HumanCenteredXAI} and allow clinicians to revise that evidence when needed \cite{Wekenborg2025RealWorldHAI,Giddings2024ClinicianPatientInteraction}.

Most thyroid ultrasound AI systems have focused on individual components of this workflow, including segmentation~\cite{gong2021multi,gong2023thyroid,sun2025clip}, classification~\cite{gong2022less,Peng2021LancetDigitalHealthThyNet,Chen2022RadiologyTIRADS,Yao2025NPJDigitMedThyGPT}, and report generation~\cite{li_ultrasound_2024,Tanno2025ClinicianVLM,li_towards_2025}. Multicenter systems \cite{Peng2021LancetDigitalHealthThyNet,Wang2024LancetDigitalHealthFNAB} and feature-aligned multimodal models \cite{Chen2022RadiologyTIRADS,Yao2025NPJDigitMedThyGPT,Tanno2025ClinicianVLM} have connected image predictions to risk descriptors \cite{Chen2022RadiologyTIRADS,Yao2025NPJDigitMedThyGPT} or management recommendations \cite{Peng2021LancetDigitalHealthThyNet,Yao2025NPJDigitMedThyGPT}. Recent studies have extended thyroid AI to fine-needle aspiration cytology \cite{Wang2024LancetDigitalHealthFNAB}, lateral lymph-node metastasis prediction \cite{Shen2025NatCommunLLNM} and rare thyroid cancer subtype classification \cite{Dai2025NatCommunThyroidSubtype}. Many systems still present their outputs as endpoints: a mask, probability, label or report-like text. The intermediate evidence that supports these outputs is often unavailable for clinical review \cite{Chen2022HumanCenteredXAI,Wekenborg2025RealWorldHAI}, correction \cite{Wekenborg2025RealWorldHAI,Giddings2024ClinicianPatientInteraction} or reuse across downstream tasks \cite{Chen2022HumanCenteredXAI,Tikhomirov2024LostCognitive,you2026learning}. This endpoint-oriented design makes it difficult to determine whether an AI result is supported by appropriate lesion localization, measurement, sonographic features and report statements.

This limitation reflects a broader challenge in medical AI. High-impact clinical AI studies increasingly emphasize workflow integration \cite{Vasey2022NatMedDECIDEAI,Wekenborg2025RealWorldHAI,dreyer2025mechanistic}, human-AI collaboration \cite{Topol2019NatMedHighPerformance,Patel2019HumanMachinePartnership,Leibig2022RadiologistsAI,Yu2024AIAssistanceRadiologists,Chen2024WorkloadCollaboration,Everett2026ToolTeammate,Strong2026HumanAICollaboration}, and evidence beyond retrospective performance \cite{Wiens2019NatMedDoNoHarm,Topol2019NatMedHighPerformance,Rajpurkar2022NatMedAIHealth,Vasey2022NatMedDECIDEAI}. The cognitive consequences of AI-supported clinical work \cite{Tikhomirov2024LostCognitive}, clinician interaction with algorithmic recommendations \cite{Giddings2024ClinicianPatientInteraction} and the transition of AI from a tool to a clinical teammate \cite{Zou2025LancetAgenticTeammates,Everett2026ToolTeammate} have also become central considerations. Generalist medical AI \cite{Moor2023NatureGMAI,Kohane2024InjectingAI,Katz2024GPTResidents,Zhou2026NEJMAIMedVersa} and multimodal foundation models \cite{Moor2023NatureGMAI,Zhou2026NEJMAIMedVersa,Tu2025ConversationalAI,McDuff2025DifferentialDiagnosis,deltadahl2025deep,pontikos2025next} extend this ambition to flexible inputs and outputs across tasks. For thyroid ultrasound, however, generality must be connected to specialty-specific requirements: lesion-level measurement \cite{Alexander2022Lancet,Tessler2017ACRTIRADS}, sonographic feature attribution \cite{Tessler2017ACRTIRADS,Chen2022RadiologyTIRADS}, anatomical context \cite{Grani2024NatRevEndocrinol}, guideline-aligned management \cite{Alexander2022Lancet,Tessler2017ACRTIRADS,Peng2021LancetDigitalHealthThyNet} and auditable reporting \cite{Chen2022HumanCenteredXAI,Tanno2025ClinicianVLM,Wekenborg2025RealWorldHAI}. We therefore treat the case-level evidence record, rather than a single prediction endpoint, as the central object of AI assistance.

Agent-based workflows \cite{Qiu2024NatMachIntellAgenticSystems,Zou2025LancetAgenticTeammates,Moritz2025NatBMECoordinatedAgents,Ferber2026AutonomousAgents,Collaco2026AgenticAIReview,kong2025ai} provide one way to implement this evidence-centerd formulation. In this setting, the agent's role is coordination rather than direct image interpretation \cite{Qiu2024NatMachIntellAgenticSystems,Zou2025LancetAgenticTeammates,Moritz2025NatBMECoordinatedAgents}: it acts as a workflow controller that plans case-specific analysis \cite{wang_plan-and-solve_2023,Schmidgall2026AgentClinic,Liu2026AgentBenchmark}, routes inputs to specialized tools \cite{yao_react:_2022,Moritz2025NatBMECoordinatedAgents,Ferber2026AutonomousAgents}, maintains intermediate state \cite{Tian2026AutonomousWorkflow} and exposes structured evidence for human review \cite{Chen2022HumanCenteredXAI,Wekenborg2025RealWorldHAI,Zou2025LancetAgenticTeammates}. Medical-agent benchmarks increasingly emphasize these capabilities in interactive settings \cite{Jiang2025NEJMAIMedAgentBench,Schmidgall2026AgentClinic,Liu2026AgentBenchmark} that require retrieval, action execution and workflow-level reasoning \cite{yao_react:_2022,Jiang2025NEJMAIMedAgentBench,Tian2026AutonomousWorkflow}. For thyroid ultrasound, the relevant evidence objects include images, lesion masks and measurements \cite{gong2021multi,gong2023thyroid,zhang2025tn5000}, radiomic descriptors and risk estimates \cite{van2017computational,Chen2022RadiologyTIRADS,Yao2025NPJDigitMedThyGPT}, report clauses \cite{rebuff_data2text_2020,li_ultrasound_2024,Tanno2025ClinicianVLM}, uncertainty signals \cite{farquhar2024detecting} and clinician corrections \cite{Yu2024AIAssistanceRadiologists,Chen2024WorkloadCollaboration,Wekenborg2025RealWorldHAI}. The agent is therefore useful insofar as it can coordinate these objects into an auditable clinical workflow.

Here we present ThyroidXAgent, a clinician-interactive agentic system that reframes thyroid ultrasound AI around an auditable case-level evidence record rather than a collection of isolated prediction endpoints. Instead of directly interpreting ultrasound images with a general-purpose multimodal model, ThyroidXAgent acts as a workflow controller that plans case-specific analyses, routes inputs to specialized tools and maintains structured intermediate evidence, including lesion masks, measurements, class probabilities, radiomic descriptors, uncertainty signals and report clauses. These evidence objects remain inspectable and editable by clinicians, and corrections can be propagated to subsequent analysis and reporting steps. We evaluate this formulation across nodule segmentation, benign-malignant classification, malignant-lesion stratification and case-level report generation using 40 heterogeneous multicentre datasets and clinician reader studies. We further introduce ThyClinScore, a lesion-level semantic metric that evaluates whether generated reports preserve clinically relevant evidence rather than surface-level wording alone. ThyroidXAgent improved produced more clinically consistent reports and reduced clinician workload while retaining an editable evidence trace. Together, these results establish evidence-centred orchestration as an alternative to both standalone predictive models and unconstrained end-to-end medical agents.

\begin{figure}[p]
    \centering
    \includegraphics[width=0.9\textwidth]{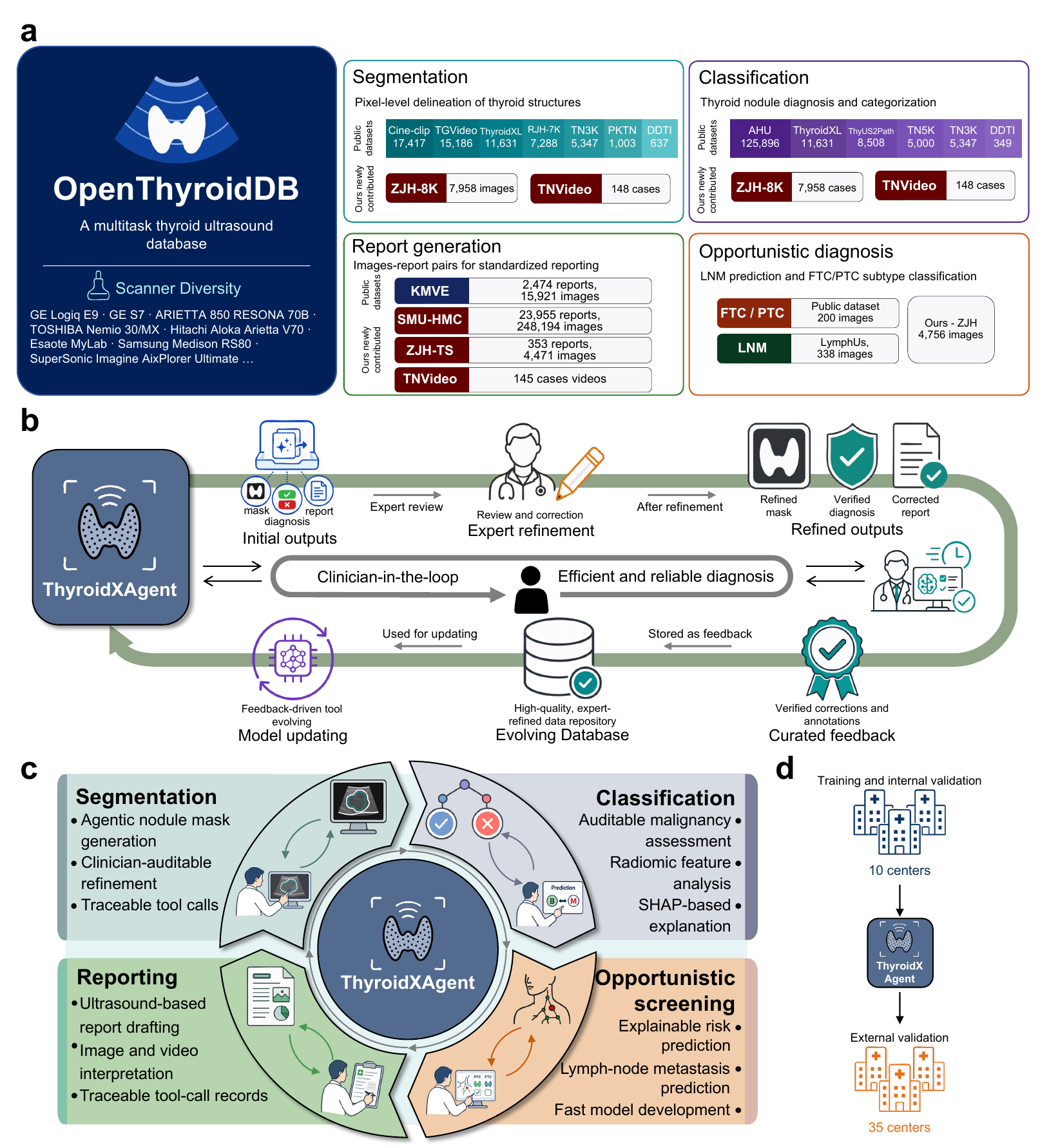}
    \caption{Multicenter thyroid ultrasound data and the ThyroidXAgent evidence workflow. \textbf{a}, OpenThyroidDB integrates curated public thyroid ultrasound resources and institutionally governed clinical cohorts into a full-spectrum resource for segmentation, benign-malignant classification, report generation and malignant-lesion stratification across diverse scanners and acquisition settings. \textbf{b}, The clinician-feedback workflow stores AI-generated masks, predictions, report clauses and clinician corrections as case-level evidence, allowing intermediate outputs to be reviewed, revised and reused in later steps. \textbf{c}, ThyroidXAgent acts as a workflow controller for four evidence-producing workflows: expert-refined nodule segmentation, clinician-verified malignancy classification, evidence-grounded structured reporting, and advanced diagnosis for lymph-node metastasis assessment and PTC/FTC subtype analysis. \textbf{d}, Multicenter validation design, with model development/internal validation followed by independent external validation.}
    \label{fig:introduction_overview}
\end{figure}

\begin{figure}[htbp]
    \centering
    \includegraphics[width=0.9\textwidth]{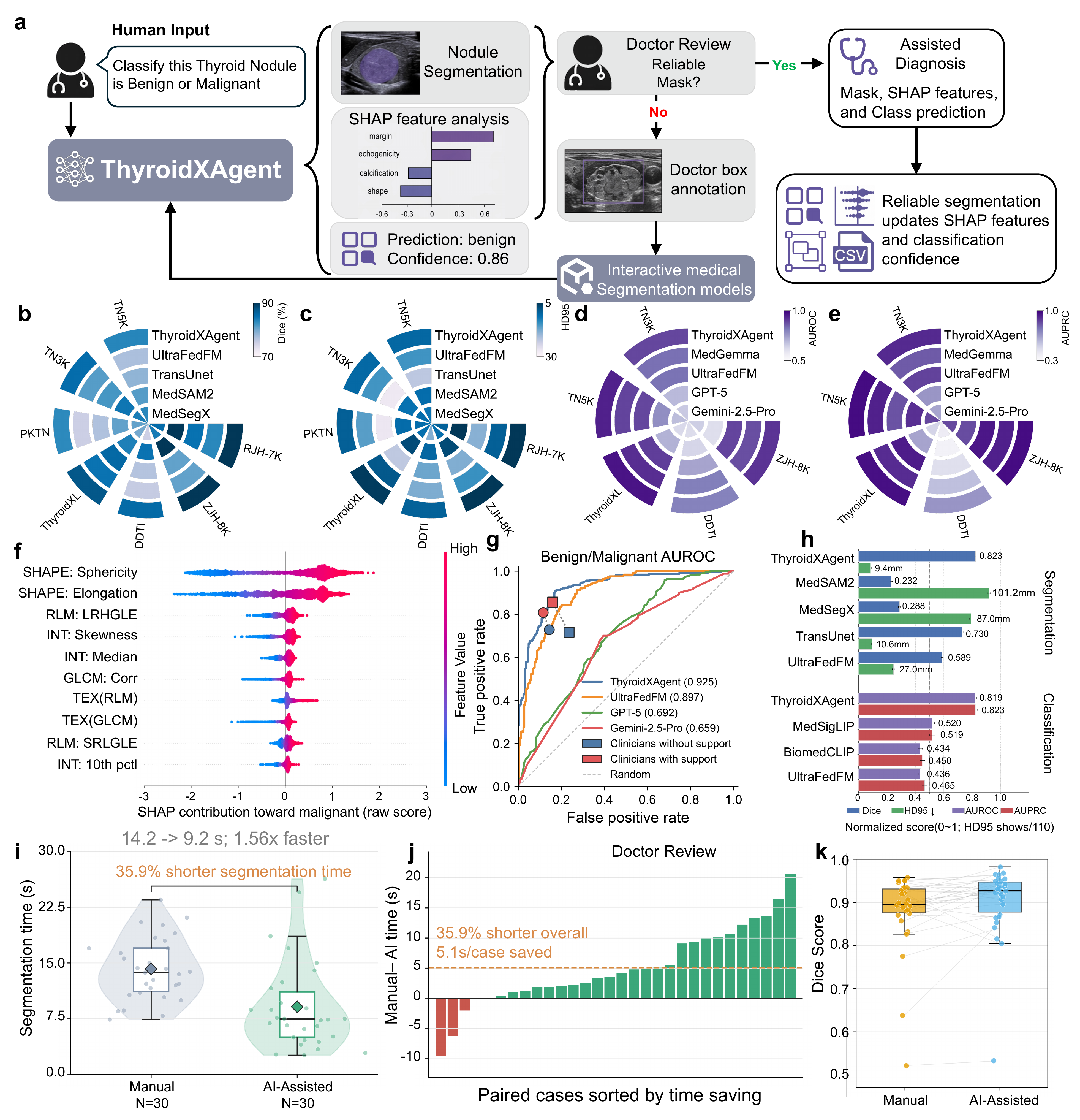}
    \caption{Agent-routed evidence for thyroid nodule segmentation, benign-malignant classification and clinician review. \textbf{a}, Interactive review workflow: clinicians assess the predicted nodule mask, accept the classification if the mask is reliable, or refine it through box annotation and interactive segmentation; SHAP-based feature attributions are then recomputed from the corrected mask. \textbf{b-e}, Segmentation and classification performance across heterogeneous thyroid ultrasound benchmarks, evaluated using Dice, HD95, AUROC and AUPRC. \textbf{f}, Cohort-level SHAP beeswarm analysis for benign-malignant classification. \textbf{g}, ROC curves on the 500-image physician comparison set, showing ThyroidXAgent, representative AI baselines and clinician operating points before and after ThyroidXAgent support; both clinicians improved with evidence support. \textbf{h}, Pooled performance on the NHC-MISD-TUS private external test set across nine models and four metrics: nodule Dice, HD95, binary AUROC and AUPRC. Bars show point estimates with 95\% CIs, and the dashed line indicates the classification chance level of 0.5. \textbf{i}, Segmentation time for manual and AI-assisted workflows. \textbf{j}, Ranked within-case time savings. \textbf{k}, Paired Dice distributions showing preserved segmentation quality with improved efficiency.}
    \label{fig:ThyroidXAgent_SegCls_performance}
\end{figure}

\section{Results}
\subsection{ThyroidXAgent organizes thyroid ultrasound diagnosis as an auditable case-level evidence workflow}

We developed ThyroidXAgent to organize thyroid ultrasound diagnosis as a case-level evidence workflow (Fig.~\ref{fig:introduction_overview}). For each examination, the agent routes the case through tool-callable steps, including nodule segmentation, measurement, benign-malignant classification, radiomics extraction, malignant-lesion stratification and report generation. The workflow stores tool outputs in a shared evidence record that can be inspected, corrected and reused across downstream tasks.

OpenThyroidDB provides the multicenter data resource underlying this workflow-level formulation. The segmentation and classification analyses included seven cohorts from six clinical centres across China, Vietnam and Colombia, acquired on at least eight ultrasound platforms (Supplementary Table~\ref{tab:dataset_summary}). Four cohorts served as internal training data. TN3K~\cite{gong2021multi}, collected at Zhujiang Hospital, Southern Medical University, Guangzhou, comprised 4,633 training, 100 validation and 614 test images acquired on GE Logiq E9, ARIETTA 850 and RESONA 70B scanners. TN5K~\cite{zhang2025tn5000}, from the Cancer Hospital, Chinese Academy of Medical Sciences, Beijing, comprised 3,500 training, 500 validation and 1,000 test images acquired on GE Logiq E9 and GE S7 scanners with 5-12 or 8-15~MHz probes. ThyroidXL~\cite{duong2025thyroidxl}, from the Vietnam National Hospital of Endocrinology, Hanoi, comprised 9,441 training, 100 validation and 2,090 test images acquired on a Hitachi Aloka Arietta V70 scanner. PKTN~\cite{sun2025clip}, from Peking University First Hospital, Beijing, comprised 703 training, 150 validation and 150 test images; the acquisition device was not disclosed. Three cohorts provided independent external test sets. DDTI~\cite{pedraza2015open}, from IDIME, Bogot\'{a}, Colombia, contributed 637 images (637 for segmentation, 349 of which also carry benign-malignant classification labels) acquired on TOSHIBA Nemio 30 and Nemio MX scanners with a 12~MHz probe. RJH-7K~\cite{shusharina2021segmentation}, from Ruijin Hospital, Shanghai Jiao Tong University School of Medicine, Shanghai, contributed 7,288 segmentation-only images acquired on multiple scanners whose models were not specified. ZJH-8K, collected at Zhujiang Hospital, Southern Medical University, Guangzhou, contributed 7,958 images (426 benign cases with 3,202 images and 723 malignant cases with 4,756 images) used for both segmentation and classification; the acquisition device was not disclosed. Overall, the segmentation and classification benchmark comprised 38{,}864 images (18{,}277 training, 850 validation, 19{,}737 test) from seven cohorts spanning six centres.

The report-generation analyses included four cohorts from three centres. SMU-HMC comprised 23,955 reports from 21,954 patients, with 248,194 associated ultrasound images. A patient-level held-out set of 400 patients, comprising 5,984 images, was reserved for internal testing. After excluding all examinations from these internally held-out patients, the remaining SMU-HMC cohort was used to develop the relevant component models, and 7,147 quality-controlled reports were selected to construct the retrieval template library. The KMVE analysis cohort~\cite{li_ultrasound_2024} comprised 2,457 cases and 4,914 image assignments, partitioned according to the original dataset split into 1,719 training, 246 validation and 492 test cases; its training partition was additionally used for template-library construction. External evaluation included two complementary cohorts. ZJH-TS provided an independent external-centre test set of 150 reports selected from the original collection of 353 reports, after excluding reports dominated by postoperative findings. TNVideo provided an independently assembled case-level cohort for the external reader study, comprising 148 ultrasound examinations, of which 145 had annotation-derived diagnostic labels. Overall, the report-generation analyses comprised four cohorts from three centres, including an internal SMU-HMC test set, an independent external-centre test set and an external reader-study cohort.

For malignant-lesion stratification, the 4,756 malignant images from ZJH-8K (also an external test set for segmentation and classification) served as the primary training set, of which 20 cases (183 images) were held out for validation. For lateral lymph-node metastasis (LNM) prediction, 180 images from LymphUs Center~1 were additionally included as training data, and the 158 images from LymphUs Center~2 served as the independent external test set. For follicular (FTC) versus papillary (PTC) thyroid carcinoma subtype classification, the 200 public images released by Dai \textit{et al.}~\cite{dai2025improving} served as the external test set. Within the ZJH-8K malignant cohort, 533 cases (3,394 images) were classified as CN0 and 190 cases (1,362 images) as CN1 for lymph-node status, and 656 cases (4,312 images) were classic PTC and 67 cases (444 images) were follicular variant for subtype.

Collectively, these resources cover heterogeneous acquisition settings, file formats, annotation types and clinical tasks, including nodule segmentation, benign-malignant classification, report generation and advanced malignant-lesion analysis (Fig.~\ref{fig:introduction_overview}a and Supplementary Table~\ref{tab:dataset_comparison}). Using this resource, ThyroidXAgent coordinates four evidence-producing workflow branches: expert-refined nodule segmentation, clinician-verified malignancy classification, expert-edited report generation and advanced diagnosis for lateral lymph-node metastasis and PTC/FTC subtype analysis (Fig.~\ref{fig:introduction_overview}b,c). Each branch contributes structured evidence to the same case-level record, including masks, measurements, class probabilities, radiomic descriptors, feature attributions, uncertainty signals and report clauses. This design allows intermediate outputs to be reused across tasks. A corrected mask can support radiomics extraction, a malignancy estimate can inform the report impression, and structured report evidence can be reviewed and edited rather than accepted as opaque text.

The resulting workflow makes auditability a property of the diagnostic process rather than a post hoc explanation attached to a final answer. Clinicians can inspect generated masks, correct segmentation errors, review SHAP- or Grad-CAM-based explanations, edit report statements and return corrected outputs to the evidence store. The subsequent results evaluate this formulation across automatic image analysis, clinician correction, malignant-lesion stratification, clinical semantic report scoring and evidence-grounded report assembly.

\subsection{Agent-routed evidence improves cross-dataset segmentation and classification}

We first evaluated the two image-analysis tasks that anchor the downstream workflow: nodule segmentation and benign-malignant classification. For each case, ThyroidXAgent collected candidate masks and class probabilities from DINOv3-based experts, selected or fused outputs using case-level quality signals, extracted radiomic features from the selected lesion mask and stored confidence, disagreement and tabular predictions as structured evidence (Supplementary Fig.~\ref{fig:ThyroidXAgent_for_seg_and_cls}). This design tests whether tool routing and evidence consolidation improve robustness across heterogeneous ultrasound datasets, where dataset bias remains a major source of performance degradation~\cite{liu2024decade}.

On seven segmentation test sets, including the independent DDTI, RJH-7K and ZJH-8K cohorts, ThyroidXAgent achieved a mean Dice coefficient of 87.21\% and a mean 95th-percentile Hausdorff distance (HD95) of 6.90~mm (Fig.~\ref{fig:ThyroidXAgent_SegCls_performance}b,c and Supplementary Table~\ref{tab:seg_performance}). It obtained the highest Dice score on six of seven test sets and the lowest HD95 on all test sets, indicating improved boundary robustness across heterogeneous acquisition conditions. The strongest baseline, MedSAM2~\cite{ma2025medsam2}, achieved a mean Dice of 85.66\%, with the largest gap on the external ZJH-8K cohort (86.29\% versus 94.30\% for ThyroidXAgent). UltraFedFM~\cite{jiang2025pretraining}, a medical imaging foundation model, reached 79.66\%.

For benign-malignant classification, ThyroidXAgent achieved a mean AUROC of 0.9466 and a mean AUPRC of 0.8361 across five test sets, including the independent DDTI and ZJH-8K cohorts (Fig.~\ref{fig:ThyroidXAgent_SegCls_performance}d,e and Supplementary Table~\ref{tab:cls_performance}). Specialized image models showed weaker cross-dataset consistency; for example, RepViT~\cite{wang2023repvit} reached AUROC of 0.777 on ThyroidXL but 0.556 on TN3K. General-purpose vision-language models~\cite{dong2022survey} also underperformed; GPT-5~\cite{openai2025gpt5systemcard} reached AUROC of 0.611-0.774 across test sets, and Gemini-2.5-Pro~\cite{comanici_gemini_2025} reached 0.616-0.687 (Supplementary Table~\ref{tab:cls_performance}). These results support a division of labour in which domain-specific image and radiomics tools generate the evidence, while the agent routes and consolidates tool outputs. We further validated on NHC-MISD-TUS, an independent private multicentre test set (Fig.~\ref{fig:ThyroidXAgent_SegCls_performance}h and Supplementary Tables~\ref{tab:nodule_dice}-\ref{tab:binary_auprc}). ThyroidXAgent led on both nodule segmentation (Dice 82.31\%, HD95 9.41~mm) and binary classification (AUROC 0.819, AUPRC 0.823). By contrast, MedSAM2 and MedSegX failed on segmentation (Dice <0.3), and zero-shot vision-language baselines (BiomedCLIP~\cite{zhang2024biomedclip}, MedSigLIP~\cite{sellergren2025medgemma}) performed at or below chance on classification. GPT-5 and Gemini-2.5-Pro could not be evaluated on this private intranet dataset.

Beyond prediction accuracy, we examined whether the structured evidence provided interpretable signals for clinician review. Cohort-level SHAP profiles~\cite{lundberg2017unified} showed that morphology-related radiomic descriptors, especially Sphericity and Elongation, dominated benign-malignant classification, whereas texture and intensity features contributed complementary information (Fig.~\ref{fig:ThyroidXAgent_SegCls_performance}f). Representative cases confirmed that accurate segmentation produced SHAP attributions and Grad-CAM maps aligned with visible nodule characteristics, whereas poor segmentation degraded these explanations (Supplementary Fig.~\ref{fig:BM_cases}).

To assess whether this evidence supports clinician decision-making, we conducted a blinded 500-image physician comparison (Fig.~\ref{fig:ThyroidXAgent_SegCls_performance}g). ThyroidXAgent achieved AUROC of 0.9256 and AUPRC of 0.9250. In the same comparison, UltraFedFM~\cite{jiang2025pretraining} reached AUROC of 0.880, GPT-5~\cite{openai2025gpt5systemcard} 0.660 and Gemini-2.5-Pro~\cite{comanici_gemini_2025} 0.619. When clinicians were provided with ThyroidXAgent's structured evidence, including SHAP-based feature attributions and nodule segmentation boundaries, both clinicians improved across all metrics. For clinician 1, accuracy rose from 79.2\% to 84.6\% (+5.4\%), precision from 83.5\% to 87.5\%, recall from 72.8\% to 80.8\%, specificity from 85.6\% to 88.4\%, and F1 from 0.778 to 0.840, while the false-positive rate fell from 14.4\% to 11.6\%. Clinician 2 improved more markedly: accuracy from 74.0\% to 84.8\% (+10.8\%), precision from 75.2\% to 84.3\%, recall from 71.6\% to 85.6\%, specificity from 76.4\% to 84.0\%, and F1 from 0.734 to 0.849, with the false-positive rate decreasing from 23.6\% to 16.0\%; clinician 2 thereby narrowly surpassed clinician 1 as the higher-performing reader.

\subsection{Clinician correction reduces segmentation time while preserving quality}

The same evidence representation supported clinician-interactive segmentation review. Clinicians inspected predicted masks, corrected segmentation errors when needed and returned the refined masks to the case-level evidence store. SHAP-based feature attributions were then recomputed using the corrected mask, allowing downstream classification evidence to reflect clinician refinement (Fig.~\ref{fig:ThyroidXAgent_SegCls_performance}a).

AI assistance reduced mean segmentation time from 14.21~s to 9.11~s per image, a 1.6-fold speedup, while preserving segmentation quality (Fig.~\ref{fig:ThyroidXAgent_SegCls_performance}i-k). AI-assisted Dice (0.903) matched or exceeded manual Dice (0.879) in approximately two-thirds of paired cases. These results indicate that the intermediate evidence layer can reduce repetitive annotation work while keeping the segmentation boundary available for clinician correction.

\begin{figure}[htbp]
    \centering
    \includegraphics[width=\textwidth]{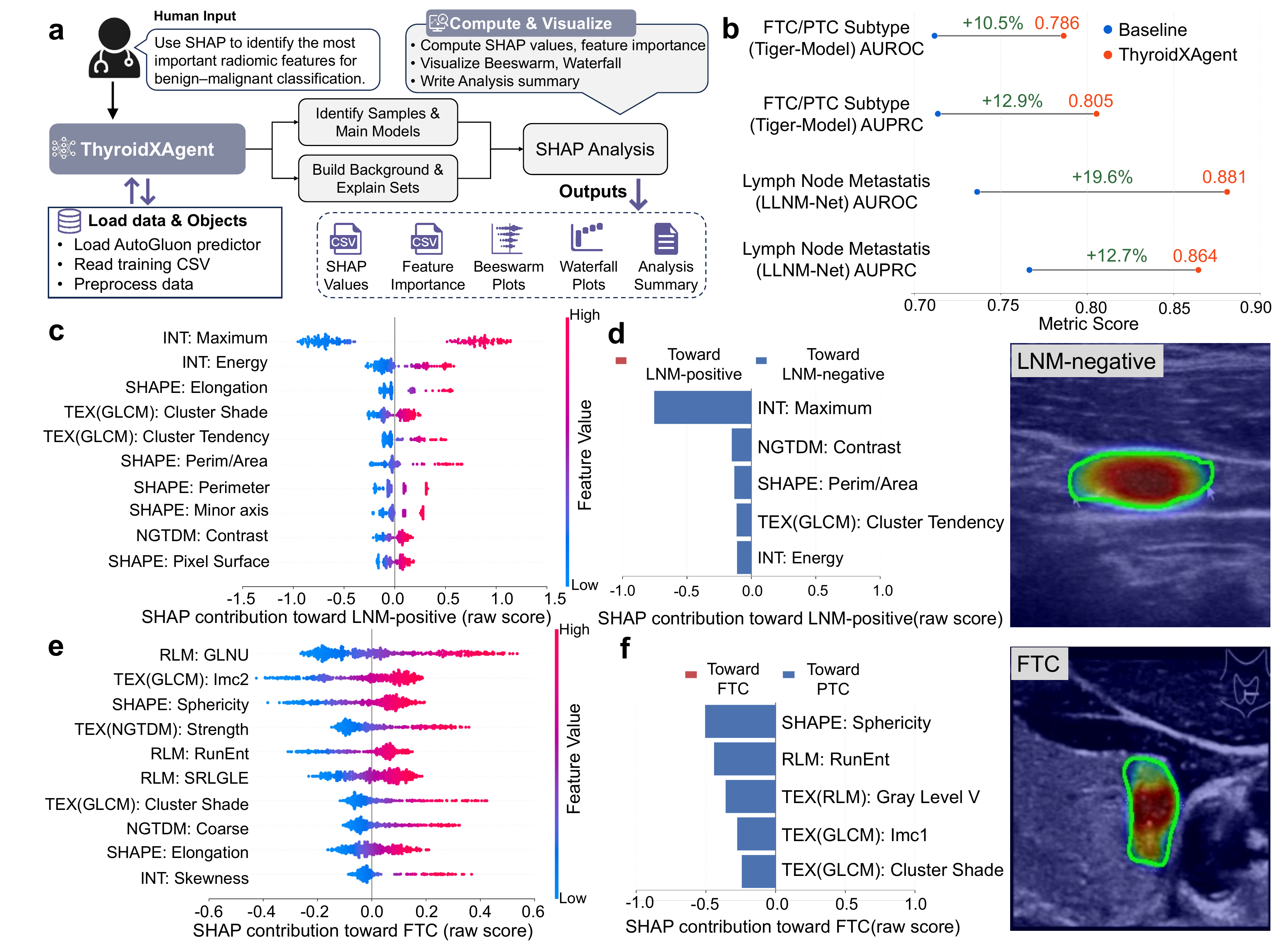}
    \caption{Shared ThyroidXAgent tools support malignant-lesion stratification with task-specific radiomic attributions. \textbf{a}, Workflow for SHAP-based interpretation of malignant-lesion tasks. \textbf{b}, Performance comparison between ThyroidXAgent and the corresponding specialist baselines for FTC/PTC subtype classification and lymph node metastasis prediction, reported as AUROC and AUPRC; percentages denote the relative improvement of ThyroidXAgent over each baseline. \textbf{c,d}, Global and representative local SHAP analyses for lymph node metastasis prediction. \textbf{e,f}, Global and representative local SHAP analyses for FTC/PTC subtype classification, showing stronger contributions from texture heterogeneity and shape descriptors.}
    \label{fig:ThyroidXAgent_Malignant_Image_tasks}
\end{figure}

\subsection{Shared agent tools generate task-specific evidence for malignant-lesion stratification}

We next tested whether the shared ThyroidXAgent tools could be redirected to clinically distinct malignant-lesion stratification tasks after the primary thyroid nodule assessment. Lateral lymph-node metastasis (LNM) prediction informs surgical planning, whereas follicular (FTC) versus papillary (PTC) thyroid carcinoma subtype discrimination informs treatment strategy and follow-up. In a conventional development pipeline, each task would require a separate workflow for preprocessing, feature extraction, prediction, interpretation and reporting. In ThyroidXAgent, the segmentation, radiomics extraction, tabular classification and routing logic were reused, with task-specific classifier fine-tuning and task instructions changed.

This reuse enabled rapid adaptation to new clinical questions while preserving the same evidence structure. ThyroidXAgent achieved AUROC of 0.864 for LNM prediction on the 158-image LymphUs Center~2 test set and 0.805 for FTC/PTC subtype classification on the 200-image Dai \textit{et al.} test set~\cite{dai2025improving} (Fig.~\ref{fig:ThyroidXAgent_Malignant_Image_tasks}b and Supplementary Table~\ref{tab:Malignant_images_tasks_performance}), outperforming the specialist baselines LLNM-Net~\cite{Shen2025NatCommunLLNM} (0.767) for LNM and Tiger-Model~\cite{Dai2025NatCommunThyroidSubtype} (0.714) for FTC/PTC.

The attribution profiles changed with the clinical task (Fig.~\ref{fig:ThyroidXAgent_Malignant_Image_tasks}a,c-f). LNM prediction relied more on lymph-node position and size features, including distance to the thyroid capsule and lesion area. FTC/PTC subtype classification relied more on texture heterogeneity and shape descriptors. These task-dependent attribution patterns indicate that ThyroidXAgent generated radiomic evidence according to the requested clinical question, rather than simply reusing the benign-malignant decision rule.

\begin{figure}[htbp]
    \centering
    \includegraphics[width=\textwidth]{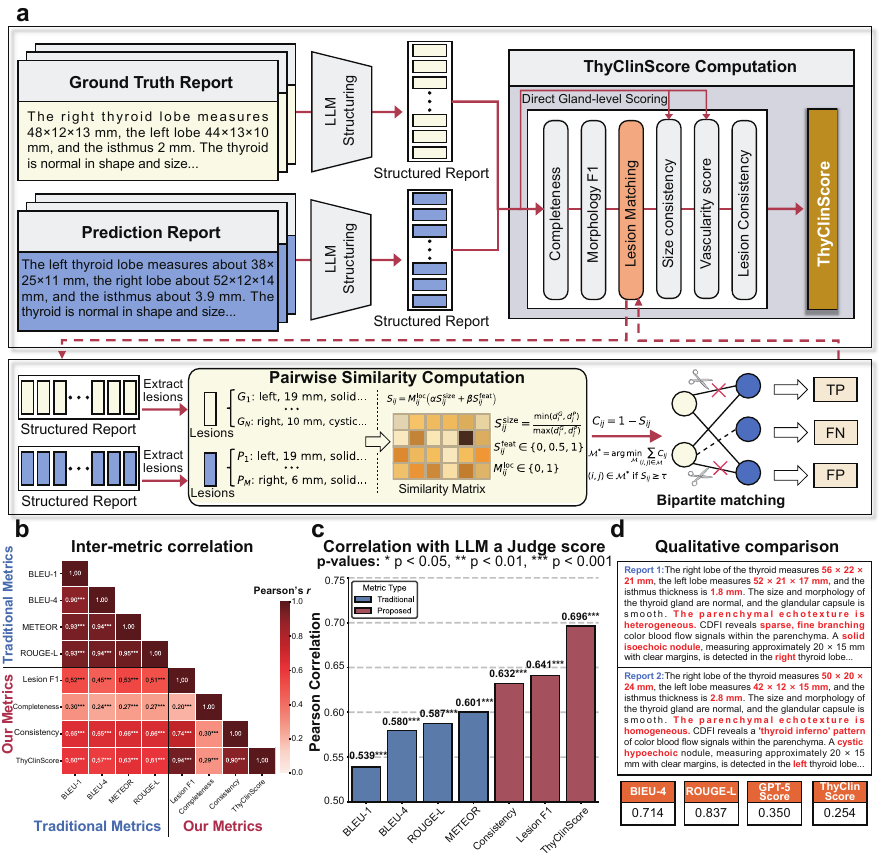}
    \caption{ThyClinScore for clinical semantic evaluation of thyroid ultrasound reports. \textbf{a}, ThyClinScore first structures the ground-truth and predicted reports into lesion-level entries, matches corresponding lesions, and scores clinically relevant attributes, including size, vascularity, morphology, lesion-level F1 and completeness. \textbf{b}, Pearson correlation matrix comparing conventional natural-language generation metrics with clinical semantic metrics. Overlap-based metrics were highly correlated with one another, whereas the clinical semantic metrics captured complementary report-quality dimensions. \textbf{c}, Pearson correlations between each metric and a location-aware LLM judge. ThyClinScore showed the strongest correlation among the evaluated metrics; asterisks denote statistical significance (* \(p<0.05\), ** \(p<0.01\), *** \(p<0.001\)). \textbf{d}, Qualitative comparison showing that reports with similar wording overlap can differ in clinically important lesion attributes, which is reflected by ThyClinScore.}
    \label{fig:thyclinscore}
\end{figure}

\begin{figure}[htbp]
    \centering
    \includegraphics[width=\textwidth]{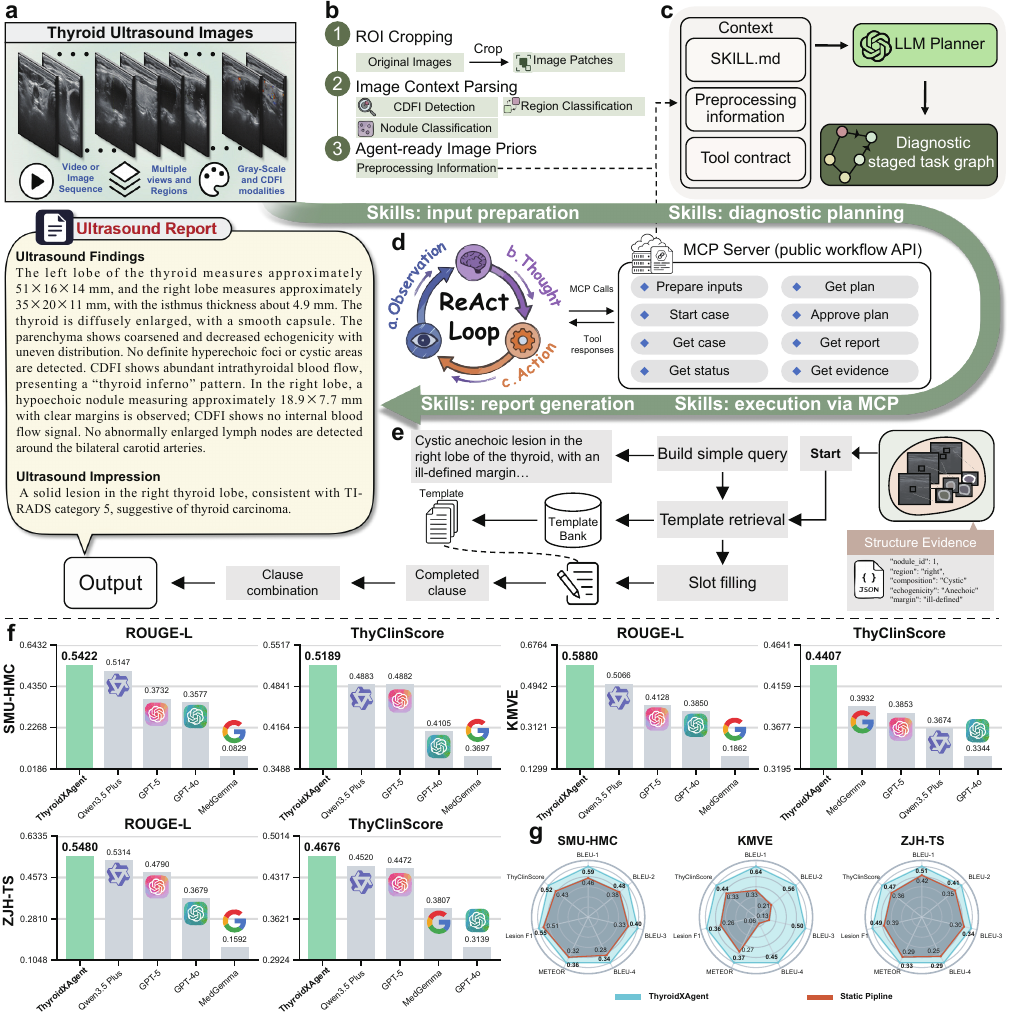}
    \caption[Evidence-grounded report assembly in ThyroidXAgent]{
    Evidence-grounded report assembly in ThyroidXAgent. \textbf{a}, Case-level thyroid ultrasound input, including video or image sequences, multiple views and anatomical regions, and greyscale and CDFI modalities. \textbf{b}, Input-preparation skills crop regions of interest, parse image context, detect CDFI information, classify nodules and anatomical regions, and convert the resulting preprocessing outputs into agent-ready image priors. \textbf{c}, Diagnostic planning combines the skill instructions, preprocessing information and tool contract with an LLM planner to generate a staged diagnostic task graph. \textbf{d}, A ReAct-style execution loop observes intermediate evidence, reasons over the next step and calls tools through an MCP server that exposes the public workflow API, including case initialization, plan approval, status checking, evidence retrieval and report generation. \textbf{e}, Structured evidence is transformed into report text through query construction, template retrieval, slot filling and clause combination, so that report clauses remain linked to measurements, lesion descriptors, risk estimates and other case-level evidence. \textbf{f}, Report-generation performance comparison on SMU-HMC, KMVE and ZJH-TS. \textbf{g}, Radar plots comparing ThyroidXAgent with the static pipeline across lexical and clinical semantic metrics.
    }
    \label{fig:thyroidxagent_report_generation}
\end{figure}

\subsection{ThyClinScore captures lesion-level report errors missed by overlap metrics}

We then addressed the evaluation of thyroid ultrasound reports. Conventional natural-language generation metrics, such as BLEU~\cite{papineni_bleu:_2001}, ROUGE~\cite{lin_rouge_2004}, and METEOR~\cite{lavie_meteor:_2007}, primarily reward surface overlap and can miss clinically important disagreements in lesion location, size, vascularity, morphology or impression. We therefore developed ThyClinScore, a clinical semantic metric that structures ground-truth and generated reports, matches lesion entries and scores clinically relevant attributes (Fig.~\ref{fig:thyclinscore}a).

ThyClinScore captured report-quality dimensions that were complementary to wording overlap. Overlap-based metrics were strongly correlated with one another, whereas the clinical semantic metrics captured distinct lesion-level and feature-level information (Fig.~\ref{fig:thyclinscore}b). Using a Pearson correlation-based evaluation approach similar to that of Li \textit{et al.}~\cite{li_towards_2025}, ThyClinScore showed the strongest correlation with a location-aware LLM judge among the evaluated metrics (Pearson's \(r=0.696\), \(p<0.001\); Fig.~\ref{fig:thyclinscore}c). Qualitative examples further show that reports with similar wording overlap can differ in clinically important attributes, which is reflected by the ThyClinScore components (Fig.~\ref{fig:thyclinscore}d).

\subsection{Evidence-grounded report assembly improves reporting consistency and efficiency}

For report generation, ThyroidXAgent converted the case-level evidence record into structured report text. The agent first used multi-view and multimodal thyroid ultrasound inputs to construct image priors, invoked diagnostic tools through planning and execution, and then assembled report clauses from structured facts using BM25 template retrieval, slot filling and clause combination (Fig.~\ref{fig:thyroidxagent_report_generation}a-g). This design links report statements to intermediate evidence, including gland measurements, nodule location, lesion size, sonographic descriptors, vascularity, lymph-node findings and diagnostic impressions, rather than generating unconstrained free text. The report-generation workflow was packaged as a reusable skill and exposed to external agents through a Workflow MCP server. This interface provided high-level case operations, including input preparation, case initiation, plan review, approval, report retrieval and evidence retrieval, while low-level model calls remained internal to the workflow. Auxiliary tool performance is summarized in Supplementary Table~\ref{tab:auxiliary_tools}. Clinicians can review the structured evidence, edit generated statements and return corrected report content to the case-level evidence store.

On conventional natural-language generation metrics, ThyroidXAgent achieved the strongest overall performance across SMU-HMC, KMVE~\cite{li_ultrasound_2024} and ZJH-TS (Fig.~\ref{fig:thyroidxagent_report_generation}f and Supplementary Table~\ref{tab:report_generation_nlg_ci}). On SMU-HMC, BLEU-1, BLEU-4 and ROUGE$_L$ reached 0.5961, 0.3405 and 0.5450, respectively. On KMVE, ThyroidXAgent ranked first on all reported overlap metrics, with BLEU-1 of 0.6209, BLEU-4 of 0.4465, METEOR of 0.3596 and ROUGE$_L$ of 0.5826. On ZJH-TS, ThyroidXAgent achieved the highest BLEU-1, BLEU-4 and ROUGE$_L$ values, reaching 0.5134, 0.2942 and 0.5529, respectively.

We further compared the adaptive tool-routing workflow with a fixed rule-based tool-calling pipeline (Fig.~\ref{fig:thyroidxagent_report_generation}g and Supplementary Table~\ref{tab:static_rule_controller_radar_metrics}). Across all three report-generation test sets, ThyroidXAgent produced larger multi-metric profiles than the static pipeline. ThyClinScore increased from 0.4293 to 0.5238 on SMU-HMC, from 0.3346 to 0.4465 on KMVE and from 0.3648 to 0.4775 on ZJH-TS.

Clinical semantic evaluation showed that the reports retained clinically relevant information beyond wording similarity (Fig.~\ref{fig:thyclinscore} and Supplementary Table~\ref{tab:report_generation_clinical_ci}). ThyroidXAgent achieved the highest ThyClinScore on SMU-HMC (0.5238), KMVE (0.4465) and ZJH-TS (0.4775). It also achieved the highest lesion-level F1 and report completeness on SMU-HMC and ZJH-TS, whereas KMVE showed a different submetric profile in which several individual components were led by other baselines. Together with the metric-correlation analysis (Fig.~\ref{fig:thyclinscore}b,c), these results indicate that clinical semantic scoring captures report-quality information not represented by conventional overlap metrics.

\begin{figure}[p]
    \centering
    \includegraphics[width=\textwidth]{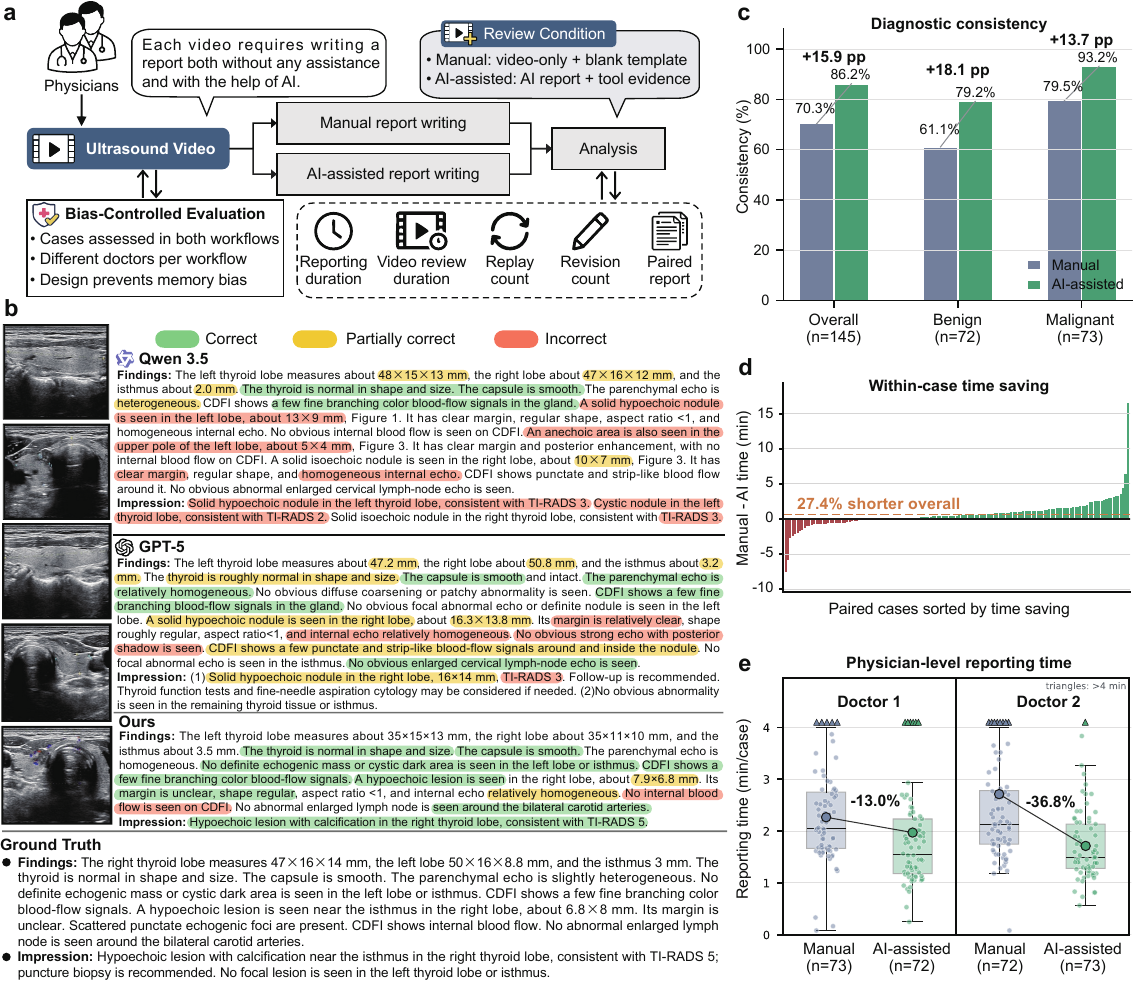}
    \caption{Reader study of AI-assisted thyroid ultrasound reporting. \textbf{a}, Cross-over reader-study design. Each ultrasound video was interpreted under both manual and AI-assisted conditions by different physicians, reducing recall bias while enabling paired case-level comparisons. \textbf{b}, Representative malignant thyroid nodule case comparing reports generated by Qwen 3.5, GPT-5 and ThyroidXAgent with the reference report. Text spans are annotated as correct, partially correct or incorrect according to medical-semantic concordance; ThyroidXAgent shows closer agreement with the reference in this example. \textbf{c}, Annotation-based diagnostic-direction consistency of manual and AI-assisted reports, shown overall and stratified by benign and malignant cases. \textbf{d}, Case-level reporting-time reduction, defined as manual minus AI-assisted reporting time and ranked across paired cases. \textbf{e}, Physician-level reporting time under the manual and AI-assisted conditions.}
    \label{fig:reader_study}
\end{figure}

Finally, we assessed human-AI cooperation in a cross-over reader study. Two physicians wrote reports for 145 thyroid ultrasound videos under manual and AI-assisted workflows, with each case evaluated in both workflows by different physicians to reduce memory bias (Fig.~\ref{fig:reader_study}a). In the 145 cases with annotation-derived diagnostic-direction labels, AI-assisted reports showed higher consistency than manual reports overall and within benign and malignant subsets (Fig.~\ref{fig:reader_study}c). AI assistance reduced mean reporting time from 2.5 to 1.8 min per case, a 27.4\% reduction, with similar time savings for both readers (Fig.~\ref{fig:reader_study}d,e). A representative malignant case shows that the structured evidence supported statements on gland morphology, nodule location, measurements, sonographic features and diagnostic impression, while leaving partially correct or incorrect statements available for clinician review (Fig.~\ref{fig:reader_study}b).

\section{Discussion}

This study reframes thyroid ultrasound AI as an auditable case-level evidence workflow rather than a collection of isolated prediction tasks. The principal contribution of ThyroidXAgent is not a new standalone segmentation, classification or report-generation model, but a shared evidence architecture in which lesion masks, measurements, probabilities, radiomic descriptors, uncertainty signals and report clauses remain visible, editable and reusable across the diagnostic process. Across heterogeneous datasets, this formulation improved nodule segmentation and benign-malignant classification, supported adaptation to additional malignant-lesion tasks, enhanced evidence-grounded reporting and reduced clinician workload. Auditability is therefore embedded in the workflow itself rather than appended retrospectively to a final prediction.

ThyroidXAgent also defines a constrained and clinically practical role for agentic AI. Rather than asking a general-purpose vision-language model to interpret ultrasound images directly, the agent plans case-specific analyses, routes inputs to specialized tools, maintains case state and exposes intermediate results for review. Image interpretation and quantitative measurement remain assigned to task-specific models, whereas the agent provides coordination and evidence management across them. This division of labour extends emerging medical-agent frameworks centred on planning, tool use and coordinated clinical workflows~\cite{Qiu2024NatMachIntellAgenticSystems,Zou2025LancetAgenticTeammates,Moritz2025NatBMECoordinatedAgents}. Its value lies less in unrestricted autonomy than in making heterogeneous model outputs coherent, traceable and correctable.

The shared evidence record further distinguishes ThyroidXAgent from a conventional fixed pipeline. A corrected lesion mask can be propagated to measurement, radiomics, classification and reporting, while anatomical context, malignancy estimates and lesion descriptors can be reused when constructing the diagnostic impression. The same evidence-producing tools could consequently be redirected towards lymph-node metastasis prediction and thyroid carcinoma subtype classification without rebuilding the complete workflow. Report generation provides a particularly stringent demonstration of this design: rather than producing unconstrained free text, ThyroidXAgent assembles report statements from structured clinical evidence. ThyClinScore complements this approach by evaluating lesion matching, clinically relevant attributes and report completeness, thereby capturing errors in location, size or morphology that may be missed by conventional lexical-overlap metrics.

The reader studies indicate that this evidence-centred formulation can support human-AI cooperation without requiring clinicians to accept an opaque recommendation. AI assistance reduced segmentation and reporting time while preserving meaningful points of clinical control, including lesion-boundary correction, evidence inspection and report editing. These findings should not be interpreted as evidence for replacing clinical judgement. Instead, they suggest that agentic systems can reduce repetitive work while keeping consequential intermediate outputs available for verification and correction. The effectiveness of such systems will ultimately depend not only on predictive accuracy, but also on whether clinicians can identify errors, understand their downstream consequences and efficiently intervene.

Several limitations remain. The evaluations were retrospective, and prospective multicentre studies are needed under real acquisition conditions, changing scanner settings and institution-specific reporting practices. The reader studies included a limited number of physicians and primarily assessed workflow feasibility and efficiency rather than patient-level clinical benefit. ThyroidXAgent also remains dependent on the validity and calibration of its component tools: routing cannot compensate for systematically biased segmentation, incomplete metadata or poorly generalized classifiers, although visible intermediate evidence may make these failures easier to detect~\cite{Wiens2019NatMedDoNoHarm,Obermeyer2019ScienceBias}. More generalizable tool design may further improve cross-domain robustness~\cite{gong2025domain,gong2026intermediate}. Finally, the template-based reporting strategy may require local adaptation, and learning from clinician corrections will require governance for data quality, privacy, provenance, model versioning and distribution shift. Future work should therefore determine whether evidence-level corrections improve downstream clinical decisions prospectively and whether the shared evidence architecture can support additional imaging tasks without sacrificing reliability or interpretability.

\section{Methods}

\subsection{Datasets and task definitions}
For image segmentation and benign-malignant classification, OpenThyroidDB integrates seven public and institutional ultrasound sources comprising 38{,}864 images (18{,}277 training, 850 validation, 19{,}737 test; Supplementary Table~\ref{tab:dataset_summary}). TN3K~\cite{gong2021multi}, TN5K~\cite{zhang2025tn5000}, ThyroidXL~\cite{duong2025thyroidxl} and PKTN~\cite{sun2025clip} served as internal cohorts for nodule segmentation training; TN3K, TN5K and ThyroidXL additionally provided benign-malignant classification labels. DDTI~\cite{pedraza2015open}, RJH-7K~\cite{shusharina2021segmentation} and ZJH-8K (collected at Zhujiang Hospital, Southern Medical University) served as independent external test sets: DDTI and ZJH-8K for both tasks, RJH-7K for segmentation only. To construct the expert pool, we merged training portions across datasets into stacked training sets, with the largest containing 18,277 images. The included cohorts differed markedly in sample size, class balance and lesion characteristics (Supplementary Fig.~\ref{fig:SegCls_statistics}).

For malignant-lesion stratification, two additional tasks were evaluated: lateral lymph-node metastasis (LNM) prediction and follicular (FTC) versus papillary (PTC) thyroid carcinoma subtype classification. The training data for both tasks comprised the 4,756 malignant images from ZJH-8K, of which 20 cases (183 images) were held out for validation. Within this cohort, 533 cases (3,394 images) were CN0 and 190 cases (1,362 images) CN1 for lymph-node status, and 656 cases (4,312 images) were classic PTC and 67 cases (444 images) follicular variant; all labels were confirmed by post-surgical histopathology. For LNM prediction, 180 images from LymphUs Center~1~\cite{mohammadi2026lymphus,abbasian2023diagnosis} were added to the training set (total 4,753 images), and the 158 images from LymphUs Center~2 served as the independent external test set. LymphUs is a multicenter open-access database of patients with histologically confirmed PTC and binary LNM labels confirmed by fine-needle aspiration biopsy, acquired using a Samsung Medison RS80 scanner (5-12~MHz L5-12/60 transducer) at Center~1 and a SuperSonic Imagine AixPlorer Ultimate scanner (4-15~MHz SL15-4 transducer) at Center~2. For FTC-versus-PTC classification, the training and validation splits were unchanged, and the 200 public images from Dai et al.~\cite{dai2025improving} served as the external test set.

\subsection{Agent workflow controller and evidence store}
ThyroidXAgent decomposes each case into tool calls and structured intermediate outputs. The workflow contains an expert pool for image segmentation and classification, a radiomics branch, a tabular prediction branch, post hoc explanation modules, anatomical-context parsers, measurement tools and report-generation modules. The LLM router operates on structured summaries rather than raw ultrasound images. During inference, it receives candidate masks, class probabilities, confidence estimates, radiomic descriptors and metadata such as image resolution, device and data source. It emits a strict JSON decision that records the selected output, supporting evidence and uncertainty signals. These outputs are normalized into a case-level evidence store containing masks, measurements, class probabilities, radiomic descriptors, explanation objects, warnings and report clauses. The evidence store is used for final prediction, report assembly and clinician review, and clinician corrections can be written back to the same record for subsequent use.

\subsection{Segmentation, classification and radiomics}
ThyroidXAgent coordinates a heterogeneous expert pool, an LLM router and a radiomics branch for image segmentation and classification: the expert pool generates candidate masks and class probabilities, the router selects the most reliable candidate based on quality metrics and case-level metadata, and the radiomics branch provides an independent classification signal. For image segmentation and classification, the expert pool is designed as a heterogeneous ensemble rather than a single model to reduce sensitivity to dataset bias~\cite{torralba2011unbiased,liu2024decade} (Supplementary Fig.~\ref{fig:ThyroidXAgent_for_seg_and_cls}). Each expert typically uses a DINOv3-based backbone~\cite{simeoni2025dinov3} with a task-specific lightweight head, though task-specific external models can also be assembled for specialized classification targets. To encourage complementary generalization profiles, these experts were trained under varying configurations along three dimensions: stacked-training composition, input resolution (128, 224 and 448 pixels), and whether DINOv3 pretrained weights were loaded or the backbone was trained from scratch. For the segmentation branch, backbone dilation rates were additionally varied to produce experts with different receptive-field profiles. This heterogeneity exposes individual experts to progressively broader data distributions, so that the router can select the most reliable candidate for each case rather than relying on a single model's bias. The effect of these stacked-training configurations on cross-dataset performance is reported in Supplementary Table~\ref{tab:stacked_performance}.

Within this pool, the segmentation branch uses a U-Net-style decoder with skip fusion to preserve fine boundary detail, and is optimized with a combined loss that balances pixel-level supervision with region-level overlap:
\[
\mathcal{L}_{\text{seg}} = \mathcal{L}_{\text{wBCE}}(m, \hat{m}) + \mathcal{L}_{\text{IoU}}(m, \hat{m}),
\]
where $m$ denotes the ground-truth mask, $\hat{m}$ the prediction, $\mathcal{L}_{\text{wBCE}}$ the class-weighted binary cross-entropy and $\mathcal{L}_{\text{IoU}}$ the intersection-over-union loss. The classification branch pools backbone features using global average and max pooling, followed by a compact attention head. To mitigate class imbalance, generalized logit adjustment (GLA)~\cite{menon2020long} is applied to the classification logits:
\[
\tilde{z}_c = z_c + \tau\log(\pi_c),
\]
where $z_c$ is the logit for class $c$, $\pi_c$ is the empirical class prior estimated from the training set, and $\tau$ is adaptively set based on the class imbalance ratio. Binary classification tasks use BCE on the adjusted logits, whereas multi-class tasks use cross-entropy on the adjusted logits. Each expert produces a class probability $p_i$; the per-expert confidence $c_i$ is taken as the maximum softmax output of the classification head for expert $i$.

Once these per-expert outputs are available, task-specific quality metrics are computed across them to inform the selection: morphological plausibility such as area, circularity and compactness, and inter-model agreement measured by pairwise IoU and HD95, for segmentation; and prediction uncertainty such as entropy and margin, and class consensus, for classification. The LLM router then selects the best mask and classification result by reasoning over these quality metrics, per-expert confidence estimates and case-level metadata, rather than by simple confidence maximization or majority voting. Depending on the configured ensemble size, the router selects either the single best expert or a subset of experts for weighted ensemble fusion. This routing design allows the system to adapt its selection to the acquisition conditions of each case, rather than relying on a fixed model ranking.

To provide a classification signal independent of the image-based experts, the radiomics branch processes the selected mask. The mask is first refined via connected-component analysis to remove isolated noisy regions when multiple disconnected components are present. Two-dimensional PyRadiomics descriptors~\cite{van2017computational}, including shape, intensity and texture features, are then extracted from the refined mask-image pair, yielding the radiomic descriptor vector. These descriptors are passed to AutoGluon-tabular classifiers~\cite{erickson2020autogluon}, which ensemble multiple tabular models under automated hyperparameter optimization. The resulting tabular class prediction is stored alongside the router's selection in the case-level evidence store, providing a complementary interpretive signal for clinician review. After selection, SHAP analysis~\cite{lundberg2017unified} is applied to the tabular classifier to estimate global and local feature contributions, while Grad-CAM~\cite{selvaraju2017gradcam} is applied to the selected segmentation model to visualize the image regions driving its mask prediction. These post hoc explanations are stored in the evidence store for clinician inspection. Importantly, the same segmentation, radiomics, tabular classification and explanation workflow is reused across benign-malignant classification and additional malignant-lesion stratification tasks, including LNM prediction and FTC/PTC subtype classification; only task-specific classifier fine-tuning, the task description supplied to the router, and, where applicable, external model assembly are changed. This reuse allows the agent workflow to be redirected to new clinical questions without rebuilding the diagnostic pipeline.

\subsection{ThyClinScore}
ThyClinScore evaluates thyroid ultrasound reports as a structured clinical semantic agreement task. It measures both report completeness and semantic consistency with the reference report. Semantic consistency is assessed at two levels: gland-level agreement for thyroid measurements, parenchymal morphology and gland-level vascularity; and lesion-level agreement for lesion detection, lesion size, lesion descriptors and lesion-level vascularity. Size and vascularity can therefore be scored at either level when the corresponding fields are available, whereas morphology mainly captures concept-level agreement in gland and parenchymal descriptions.

For each case, the reference report \(R^{\mathrm{gt}}\) and generated report \(R^{\mathrm{pred}}\) were converted into a shared schema containing thyroid measurements, parenchymal findings, lesion attributes, lymph-node findings and diagnostic impressions. Measurements were standardized in millimetres, categorical fields were restricted to predefined thyroid ultrasound descriptors, and absent information was represented as null. The schema retained clinically relevant information, including lesion location, size, composition, echogenicity, margin, shape, echogenic foci, vascularity and TI-RADS category. These fields supported subsequent lesion-level matching and semantic scoring.

At the lesion level, reference and predicted lesions were first matched before lesion detection and attribute agreement were evaluated. Let \(G=\{g_i\}_{i=1}^{m}\) denote reference lesions and \(P=\{p_j\}_{j=1}^{n}\) denote predicted lesions. For each candidate pair, we computed
\[
S_{ij}=M^{\mathrm{loc}}_{ij}
\left(\alpha_sS^{\mathrm{size}}_{ij}+\beta_fS^{\mathrm{feat}}_{ij}\right),
\qquad
S^{\mathrm{size}}_{ij}=\frac{\min(d_i,d_j)}{\max(d_i,d_j)},
\]
where \(d_i\) and \(d_j\) are maximum lesion diameters, \(M^{\mathrm{loc}}_{ij}\in\{0,1\}\) is a hard anatomical-location gate, and \(S^{\mathrm{feat}}_{ij}\in\{0,0.5,1\}\) scores lesion-composition agreement. Explicit left-right mismatches were assigned \(M^{\mathrm{loc}}_{ij}=0\). We solved the bipartite assignment using \(c_{ij}=1-S_{ij}\) and retained pairs above a predefined threshold. Matched pairs were treated as true positives, unmatched reference lesions as false negatives and unmatched predicted lesions as false positives, from which lesion-level precision, recall, F1 and false discovery rate were calculated.

Numerical measurements were scored using mean relative error (MRE), applied to thyroid lobe and isthmus measurements at the gland level and lesion dimensions at the lesion level. For paired dimensions \(K\),
\[
\mathrm{MRE}_{K}=\frac{1}{|K|}\sum_{k\in K}
\frac{|p_k-g_k|}{\max(|g_k|,\epsilon)},\qquad
s_{\mathrm{size}}(\mathrm{MRE}_{K};\tau)=2^{-(\mathrm{MRE}_{K}/\tau)^2},
\]
where \(\epsilon\) provides numerical stability and \(\tau\) controls tolerance to size error. Vascularity was evaluated at either level when available. It was discretized into four grades, from absent flow to markedly increased flow, and scored as
\[
s_{\mathrm{vasc}}=\max\left(0,1-\frac{|v^{\mathrm{gt}}-v^{\mathrm{pred}}|}{3}\right).
\]
Morphological consistency was computed at the concept level for gland and parenchymal descriptions. A thyroid-specific lexicon mapped report phrases to concepts covering echogenicity, texture, margin, shape, calcification, posterior acoustic features and composition. For concept sets \(C^{\mathrm{gt}}\) and \(C^{\mathrm{pred}}\), morphology agreement was scored by concept F1. For matched lesions, categorical descriptors were scored by mean accuracy over reference-present fields, including composition, echogenicity, margin, shape and echogenic foci.

Gland-level and lesion-level assessments were combined into a clinical consistency score \(C\), which aggregates thyroid-size agreement, lesion-size agreement, vascularity agreement, lesion-detection F1, lesion-feature accuracy and morphology agreement. Components not applicable to either report were excluded from the denominator, whereas reference-present fields missing from the generated report contributed zero:
\[
C=\frac{\sum_{q\in\mathcal{Q}}w_qs_q}{\sum_{q\in\mathcal{Q}}w_q},
\]
where \(s_q\) denotes a valid component score and \(w_q\) its predefined weight. Report completeness \(B\) was defined as the weighted fraction of required information present in the generated structured report:
\[
B=\frac{\sum_{r\in\mathcal{R}}u_rb_r}{\sum_{r\in\mathcal{R}}u_r}.
\]
The completeness groups were thyroid measurements, parenchyma, lesions, impression and lymph-node description; \(b_r\) indicates presence of group \(r\), and \(u_r\) denotes its predefined weight. The final ThyClinScore was
\[
\mathrm{ThyClinScore}=
\left[\lambda B+(1-\lambda)C\right]
\left[\eta+(1-\eta)F_L\right],
\]
for cases with reference lesions, where \(\lambda\) balances completeness and clinical consistency, and \(\eta\) controls the minimum lesion-detection gate. For reports without reference lesions, the gate was omitted. This design rewards complete and semantically consistent reports while penalizing missed or hallucinated lesions.


\subsection{Report generation}
For the reporting branch, thyroid ultrasound reporting was formulated as a case-level evidence-to-report task rather than as single-image captioning. Given an image set \(\mathcal{I}=\{I_j\}_{j=1}^{N}\), which could include video frames, static images, multiple anatomical views, greyscale ultrasound and colour Doppler images, the workflow generated a structured report \(R\) and an accompanying evidence trace. The report covered thyroid gland morphology, nodule-level findings, lymph-node findings when present and the diagnostic impression. The trace recorded the intermediate observations supporting each report component.

Input preparation converted heterogeneous case files into image priors for subsequent planning. Images were standardized and processed by auxiliary modules for region-of-interest cropping, anatomical-context parsing, colour Doppler identification, nodule-presence triage and pixel-spacing estimation when required. The resulting priors encoded region, view, modality, Doppler status and nodule likelihood. They were used to guide downstream analysis, rather than inserted directly into report text.

The reporting workflow was packaged as a single reusable skill. The skill defined the reporting objective, evidence schema, staged diagnostic process and constraints for evidence-grounded writing. External agents invoked this skill through a Workflow Model Context Protocol (MCP) server, which exposed high-level case operations for input preparation, case initiation, plan review and approval, and report and evidence retrieval. Low-level operations, including segmentation, classification, measurement and captioning, remained internal to the workflow. This separated a stable public interface from the image-processing and model-execution details needed to construct reliable evidence. It also made the resulting report traceable for clinician review.

After input preparation, ThyroidXAgent used a planner-executor design~\cite{wang_plan-and-solve_2023}. The planner received the skill instructions, tool contract and image priors, and generated a case-specific staged task graph covering gland assessment, nodule analysis, lymph-node assessment, evidence fusion and report generation. This graph provided global diagnostic constraints, while allowing case-specific execution. In the deployable workflow, the plan could be inspected and approved before model execution. The executor then followed the approved graph and used ReAct-style local decision-making~\cite{yao_react:_2022} to select images and internal tools according to intermediate observations. For example, cases without nodule priors could bypass nodule-feature classification, whereas lateral-neck images could trigger lymph-node screening before report assembly.

Selected model operations were executed by the internal diagnostic runtime. In deployment, the Workflow MCP process remained lightweight and delegated GPU inference to a separate tool service. The runtime could call preprocessing modules, gland captioning, spacing prediction, thyroid and nodule measurement, nodule segmentation, nodule-feature classification, malignancy classification, cervical lymph-node screening and nodule-level fusion. Their outputs were normalized into a shared evidence object before language generation. Gland-level evidence included thyroid lobe and isthmus measurements, parenchymal morphology and vascularity. Nodule-level evidence included anatomical location, size, composition, echogenicity, margin, shape, echogenic foci, vascularity, segmentation-derived measurements and risk-related predictions. Lymph-node evidence summarized cervical-region screening results. When multiple views described the same lesion, the fusion stage consolidated compatible findings into case-level nodule entries and retained warnings for incomplete or conflicting outputs.

Report text was generated by controlled data-to-text assembly~\cite{rebuff_data2text_2020}. To reduce factual hallucination risk~\cite{farquhar2024detecting}, ThyroidXAgent used training-free BM25 template retrieval, slot filling and clause assembly rather than unconstrained free-text decoding. Corresponding template libraries were constructed from the SMU-HMC and KMVE datasets. The SMU-HMC was derived from 7,147 quality-controlled reports after excluding all reports from the 400 patients reserved for internal testing. The KMVE was constructed from all 1,719 reports in its training partition, with the validation and test partitions excluded. During template construction, whole reports were decomposed into five clinically defined clause categories: thyroid measurement, gland morphology, nodule findings, lymph-node findings and ultrasound impression. During inference, the evidence object assembled through tool execution was partitioned into corresponding evidence blocks and converted into category-specific BM25 queries. Retrieved templates served as linguistic frames; slot filling inserted case-specific measurements, descriptors, TI-RADS-relevant information, risk categories and other tool-derived evidence, and clause assembly combined the completed findings and impression clauses into the final report.

Notably, KMVE contains only the ultrasound findings section and does not provide original measurement values. The template library constructed from KMVE therefore retained the original structure and linguistic style of the KMVE findings and was used in accordance with its original findings-only evaluation protocol. Because tool execution was decoupled from report realization, the retrieval libraries could be exchanged in a plug-and-play manner to align generated reports with centre- or corpus-specific writing conventions, without retraining the upstream image-analysis tools or modifying the underlying evidence schema. The same procedure can be used to construct updated retrieval libraries from new report sources (Supplementary Fig.~\ref{fig:report_generation_multimodal_interface}c).

The final artefact consisted of the report text and its supporting evidence trace. Clinicians or external agents could inspect the diagnostic plan, verify the evidence used for each statement, review warnings and edit the generated report. Corrected reports could then be returned to the case-level evidence store for subsequent review and model updating. This design kept report generation constrained by structured clinical evidence while preserving a reviewable path from image inputs to final report statements.

\subsection{Reader studies and statistical analysis}
For segmentation review, clinicians corrected AI-generated masks, and correction time and paired Dice scores were compared with manual segmentation. For report writing, two physicians evaluated 145 thyroid ultrasound videos with annotation-derived diagnostic labels in a cross-over design. Each case was interpreted under both manual and AI-assisted conditions, but by different physicians, to reduce memory and recall bias. In the manual condition, physicians reviewed the ultrasound video and completed a blank structured-report template. In the AI-assisted condition, they reviewed an AI-generated draft together with the supporting tool-derived evidence and revised the report as required. We recorded reporting time, video-review time and the final submitted report for each assessment. Reporting efficiency was assessed at both the case and physician levels, with case-level time saving defined as the manual minus AI-assisted reporting time.

For annotation-based diagnostic-direction analysis, frame-level benign and malignant bounding-box annotations were aggregated into case-level labels. Cases containing any malignant annotation were classified as malignant or suspicious, whereas cases containing only benign annotations were classified as benign. This yielded 145 labelled cases, comprising 72 benign and 73 malignant or suspicious cases. A prespecified rule-based procedure assigned each submitted report to a benign or low-risk, malignant or suspicious, or ambiguous direction using TI-RADS categories and diagnostic terms. Directional consistency was defined by agreement with the annotation-derived case label and was summarized overall and within the benign and malignant or suspicious strata. For statistical analysis, performance was summarized using the primary metric for each task: Dice and HD95 for segmentation, AUROC and AUPRC for classification, conventional natural-language generation metrics for report wording, and ThyClinScore and its submetrics for report semantics. Confidence intervals in the supplementary tables were estimated by nonparametric bootstrap resampling of the test set. Correlations between report metrics and the location-aware LLM judge were evaluated using two-sided Pearson correlation tests.

\section{Data availability}
The public datasets used in this study are available from their original sources, as cited in the Methods and Supplementary Table~\ref{tab:dataset_summary}. The publicly released data associated with this study are available through ThyroidOpenDB at \url{https://huggingface.co/datasets/MedXAgent/ThyroidOpenDB}.
The study was approved by the Medical Ethics Committee of Zhujiang Hospital, Southern Medical University (approval no. 2026-KY-081-01).


The NHCMISD dataset was collected from real-world clinical ultrasound cases provided by the National Health Commission Medical Imaging Standard Database, for which the co-author, Prof. Dexing Kong, has authorized access. All data usage complied with relevant institutional and regulatory requirements.

\section{Code availability}
The source code for ThyroidXAgent is publicly available on GitHub at \url{https://github.com/MedXAgent/ThyroidXAgent}. The website will be made available after acceptance. The trained model weights are available on Hugging Face at \url{https://huggingface.co/MedXAgent/ThyroidXAgent}. The repositories contain the resources required to reproduce the reported computational analyses.

\section{Acknowledgements}
This work was supported in part by the National Natural Science Foundation of China (Grant No. 62322608), the Zhejiang Provincial Natural Science Foundation of China (Grant No. LQN26F020029), and the Natural Science Foundation of Guangdong Province (Grant No. 2024A1515010255).

\section{Author contributions}

\begin{itemize}
    \item \textbf{Conceptualization:} H.G. and G.L.
    \item \textbf{System development:} S.C., B.W. and X.X.
    \item \textbf{Computational evaluation:} S.C., B.W. and S.W.
    \item \textbf{Visualization:} S.C., B.W. and H.G.

    \item \textbf{Reader study:} H.W., Q.L. and F.C.
    \item \textbf{Data curation and organization:} H.G., S.C., B.W., Y.W., G.Y., H.W., M.M., D.K., Q.L., W.L. and F.C.
    \item \textbf{Writing--original draft:} H.G., Y.W., S.C. and B.W.
    \item \textbf{Writing--review and editing:} All authors.
    \item \textbf{Supervision:} Q.L., W.L., F.C. and G.L.
\end{itemize}

\section{Confilts}
The authors has no confilt of interests.

\clearpage
\begin{appendices}

\renewcommand{\thefigure}{S\arabic{figure}}
\renewcommand{\thetable}{S\arabic{table}}
\setcounter{figure}{0}
\setcounter{table}{0}

\section*{Supplementary information}
\phantomsection
\addcontentsline{toc}{section}{Supplementary information}

\newcommand{\suppcontentsline}[3]{%
  \noindent\hyperref[#1]{\textbf{#2}}\enspace #3\dotfill\pageref{#1}\par
}

\subsubsection*{Supplementary figures}

\begin{figure}[p]
    \centering
    \includegraphics[width=\textwidth,height=0.72\textheight,keepaspectratio]{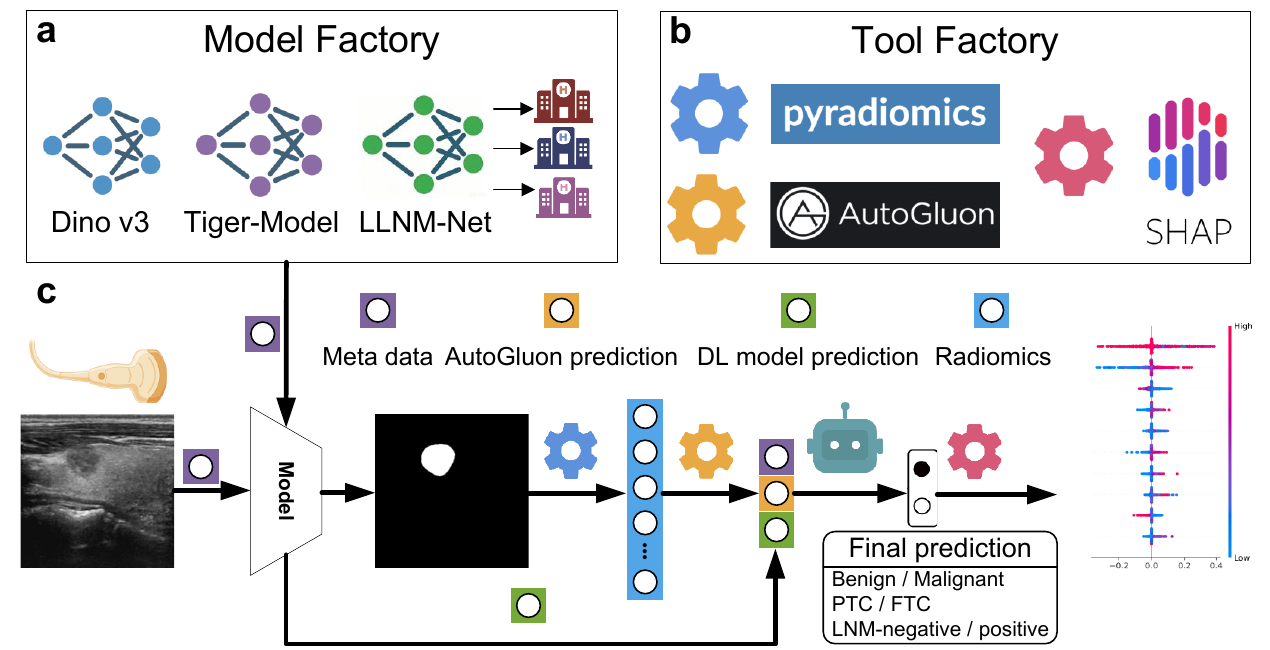}
    \caption{ThyroidXAgent architecture for thyroid nodule segmentation and classification. DINOv3-based experts generate candidate segmentation masks and malignancy probabilities. The radiomics branch extracts PyRadiomics features from the selected lesion mask, applies an AutoGluon classifier and computes SHAP attributions. The LLM router integrates the candidate outputs with image metadata, including resolution, device and data source, to select the final prediction and its supporting evidence.}
    \label{fig:ThyroidXAgent_for_seg_and_cls}
\end{figure}

\begin{figure}[p]
    \centering
    \includegraphics[width=\textwidth,height=0.72\textheight,keepaspectratio]{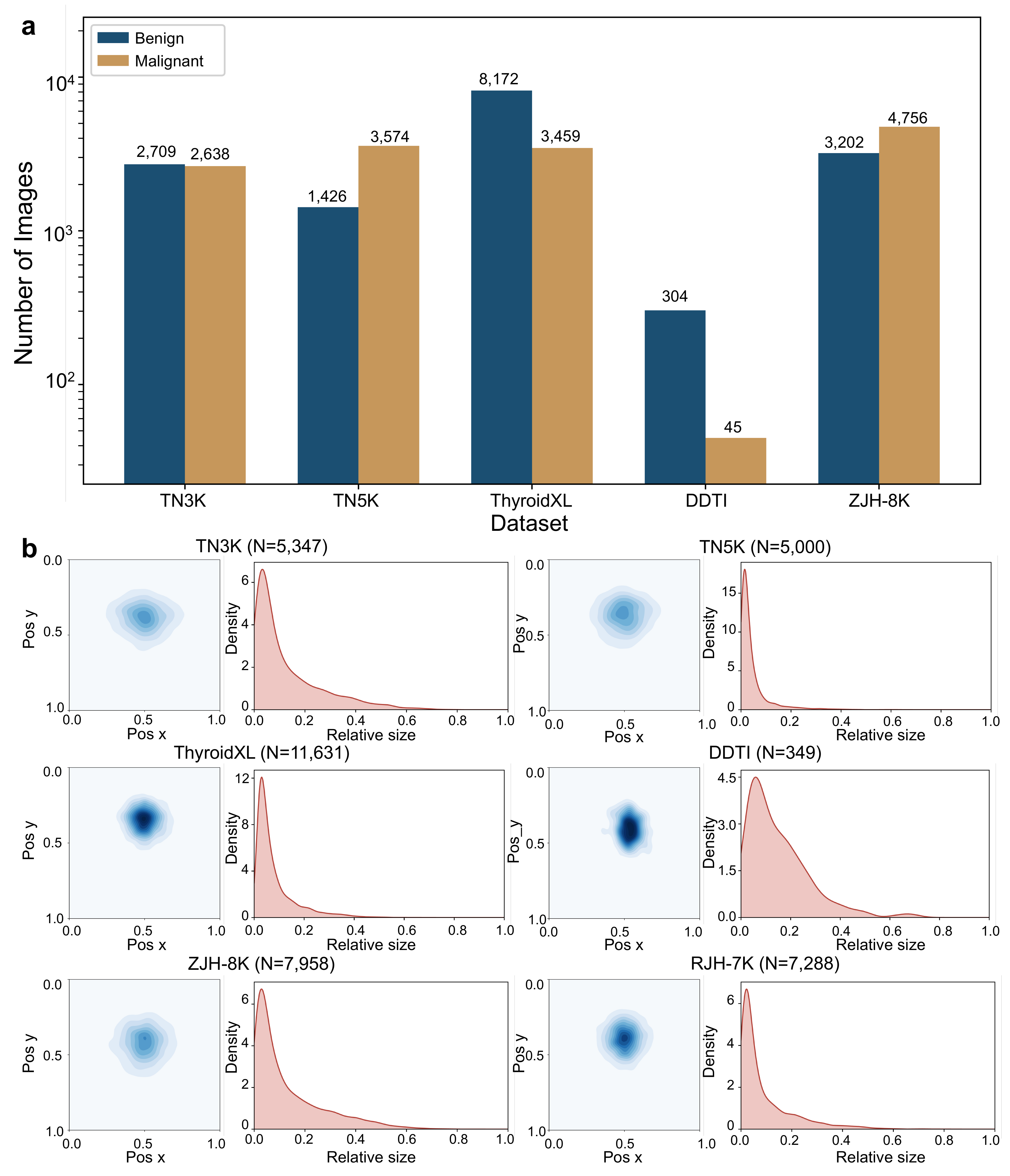}
    \caption{Dataset composition and lesion distributions across thyroid ultrasound cohorts. Top, grouped bar chart comparing the numbers of benign and malignant images in TN3K, TN5K, ThyroidXL, DDTI and ZJH-8K on a logarithmic y axis, highlighting marked variation in cohort size and class balance across cohorts. Bottom, for each of the five datasets, two-dimensional kernel density estimates of normalized lesion-mask centroid positions (left) and relative lesion size distributions, defined as mask area divided by image area (right). All size distributions share a common x-axis range, and all spatial maps use a common density scale, enabling direct cross-dataset comparison of where lesions appear within the ultrasound frame and how large they tend to be.}
    \label{fig:SegCls_statistics}
\end{figure}

\suppcontentsline{fig:ThyroidXAgent_for_seg_and_cls}{\hyperref[fig:ThyroidXAgent_for_seg_and_cls]{Figure S1.}}{\hyperref[fig:ThyroidXAgent_for_seg_and_cls]{ThyroidXAgent architecture for thyroid nodule segmentation and classification}}
\suppcontentsline{fig:SegCls_statistics}{\hyperref[fig:SegCls_statistics]{Figure S2.}}{\hyperref[fig:SegCls_statistics]{Dataset composition and lesion distributions across thyroid ultrasound cohorts}}

\begin{figure}[p]
    \centering
    \includegraphics[width=\textwidth,height=0.72\textheight,keepaspectratio]{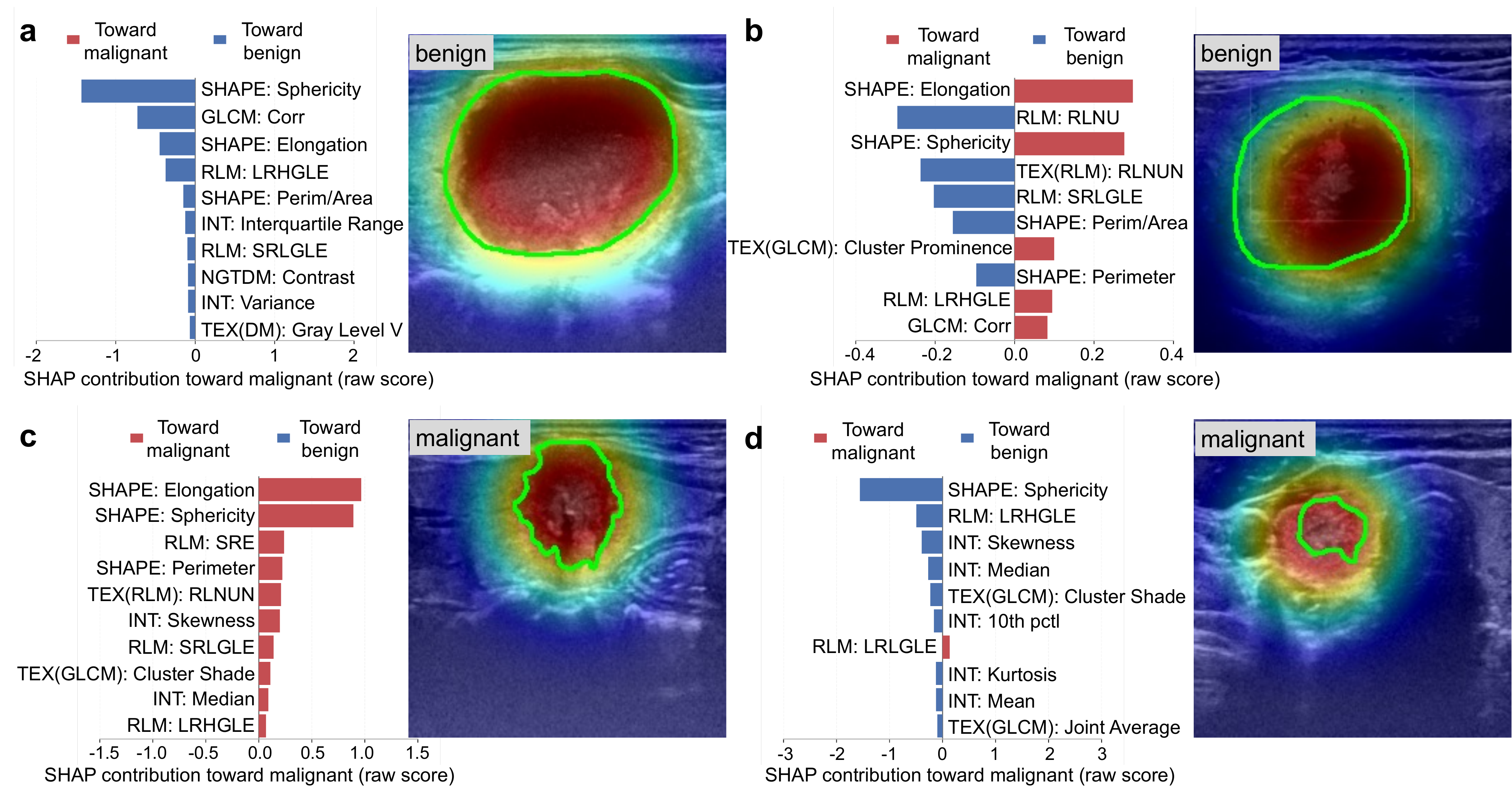}
    \caption{Case-level explanations for thyroid nodule classification. Left, SHAP values for the most influential radiomic features; red and blue indicate contributions towards malignant and benign predictions, respectively. Right, ultrasound images with segmentation contours and Grad-CAM maps from the selected segmentation model. Representative benign and malignant cases with accurate and inaccurate segmentation are shown.}
    \label{fig:BM_cases}
\end{figure}

\suppcontentsline{fig:BM_cases}{\hyperref[fig:BM_cases]{Figure S3.}}{\hyperref[fig:BM_cases]{Case-level explanations for thyroid nodule classification}}

\begin{figure}[p]
    \centering
    \includegraphics[width=\textwidth,height=0.72\textheight,keepaspectratio]{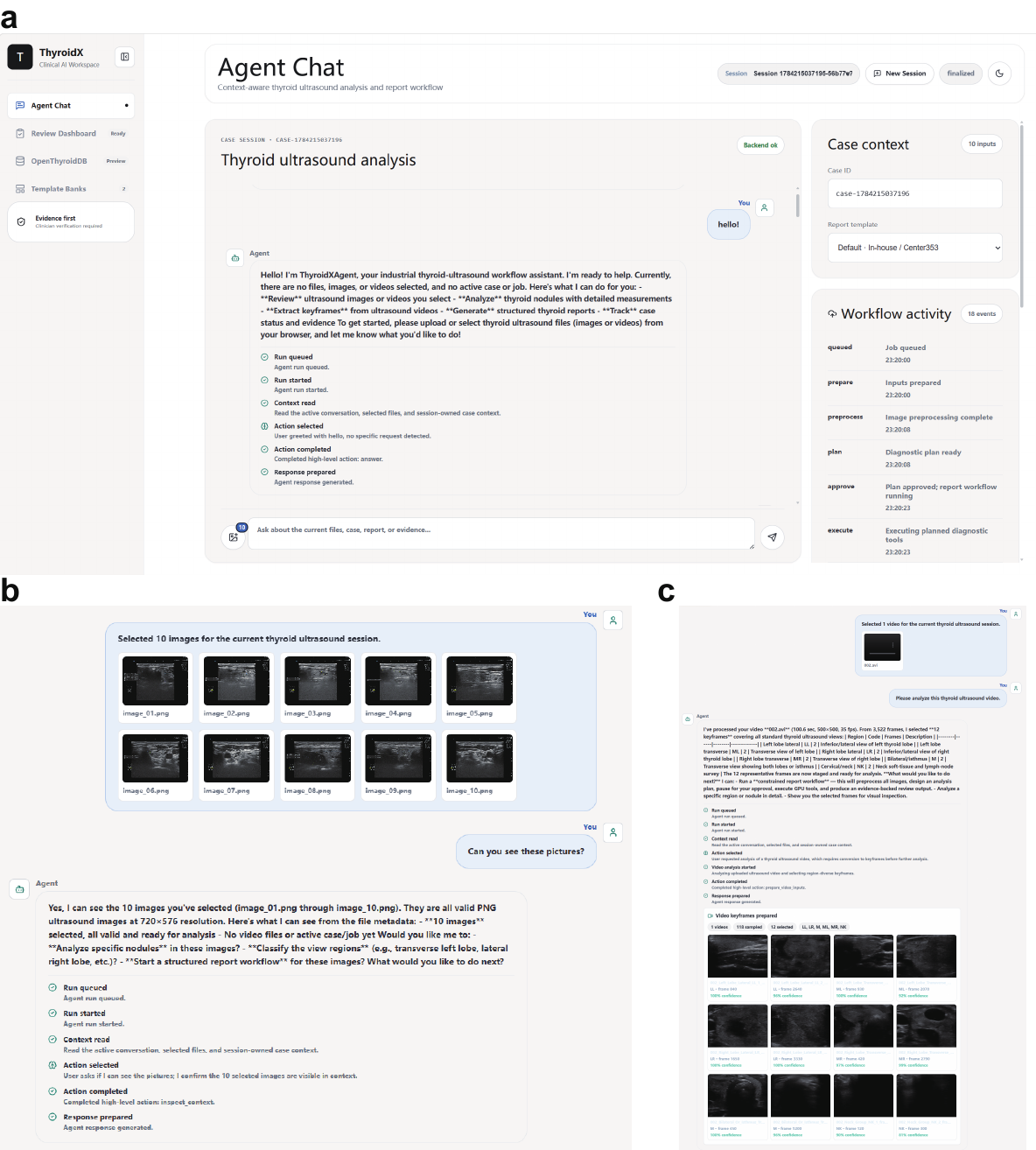}
    \caption{Multimodal input processing and case interaction in ThyroidXAgent. \textbf{a}, Agent Chat interface for initiating and monitoring case-level thyroid ultrasound analysis, with the case context and workflow activity displayed alongside the conversation. \textbf{b}, Visual perception of case-level, multiview thyroid ultrasound inputs, including image selection, anatomical-region recognition and organization of image-context priors. \textbf{c}, Video-input processing, in which an ultrasound video is sampled into representative frames and organized with anatomical-region predictions and image-context priors for downstream analysis.}
    \label{fig:report_generation_multimodal_interface}
\end{figure}

\begin{figure}[p]
    \centering
    \includegraphics[width=\textwidth,height=0.72\textheight,keepaspectratio]{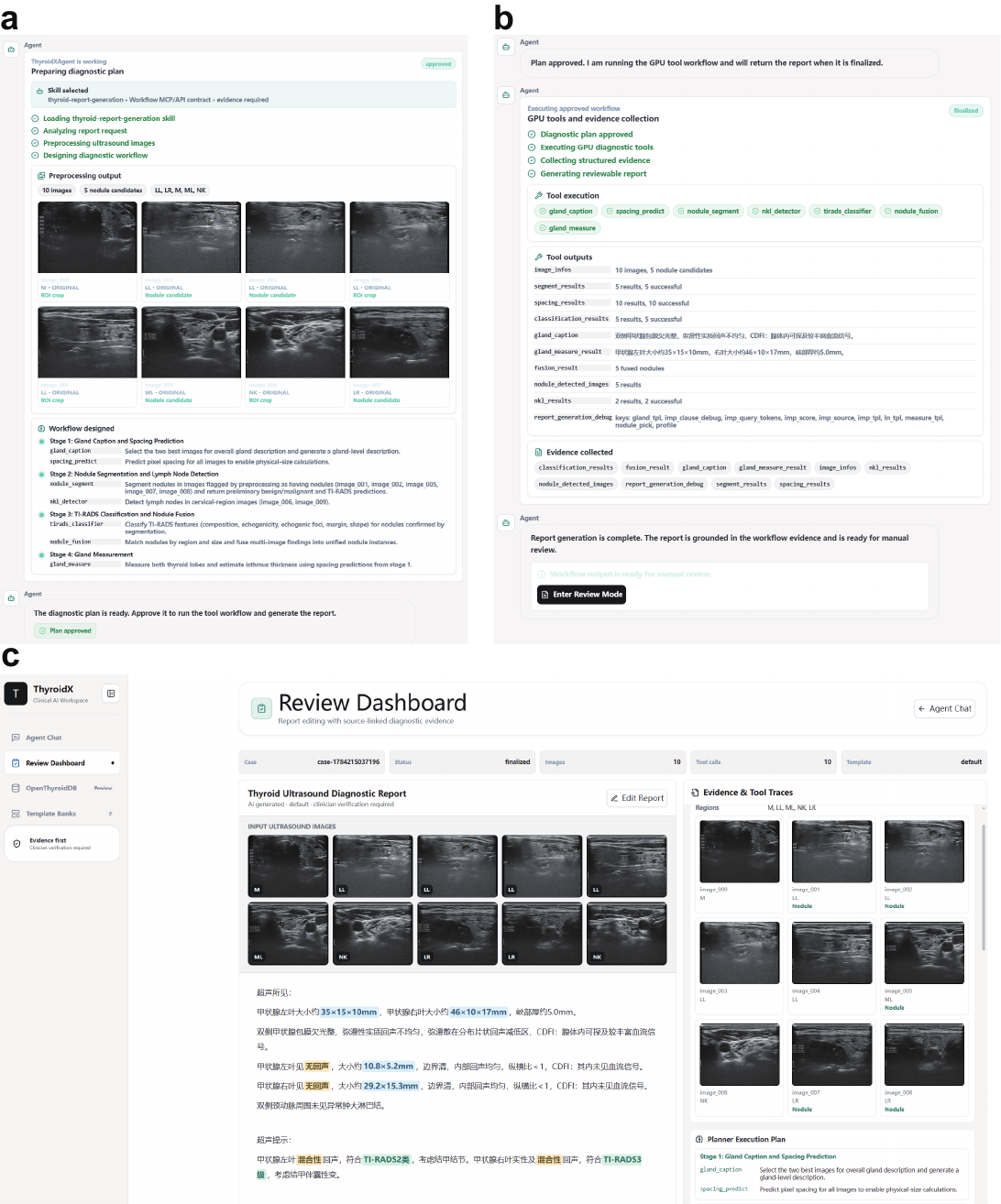}
    \caption{Planning, tool execution and clinician review in the report-generation workflow. \textbf{a}, Case-specific planning, in which preprocessing outputs, workflow instructions and diagnostic objectives are converted into a staged plan for review and approval. \textbf{b}, Execution after plan approval, showing internal tool calls, intermediate outputs, evidence collection and progression from the approved plan to a reviewable report. \textbf{c}, Report Review Dashboard displaying the generated report alongside source images, structured evidence and tool traces for clinician inspection and editing.}
    \label{fig:report_generation_planning_execution_review}
\end{figure}

\begin{figure}[p]
    \centering
    \includegraphics[width=\textwidth,height=0.72\textheight,keepaspectratio]{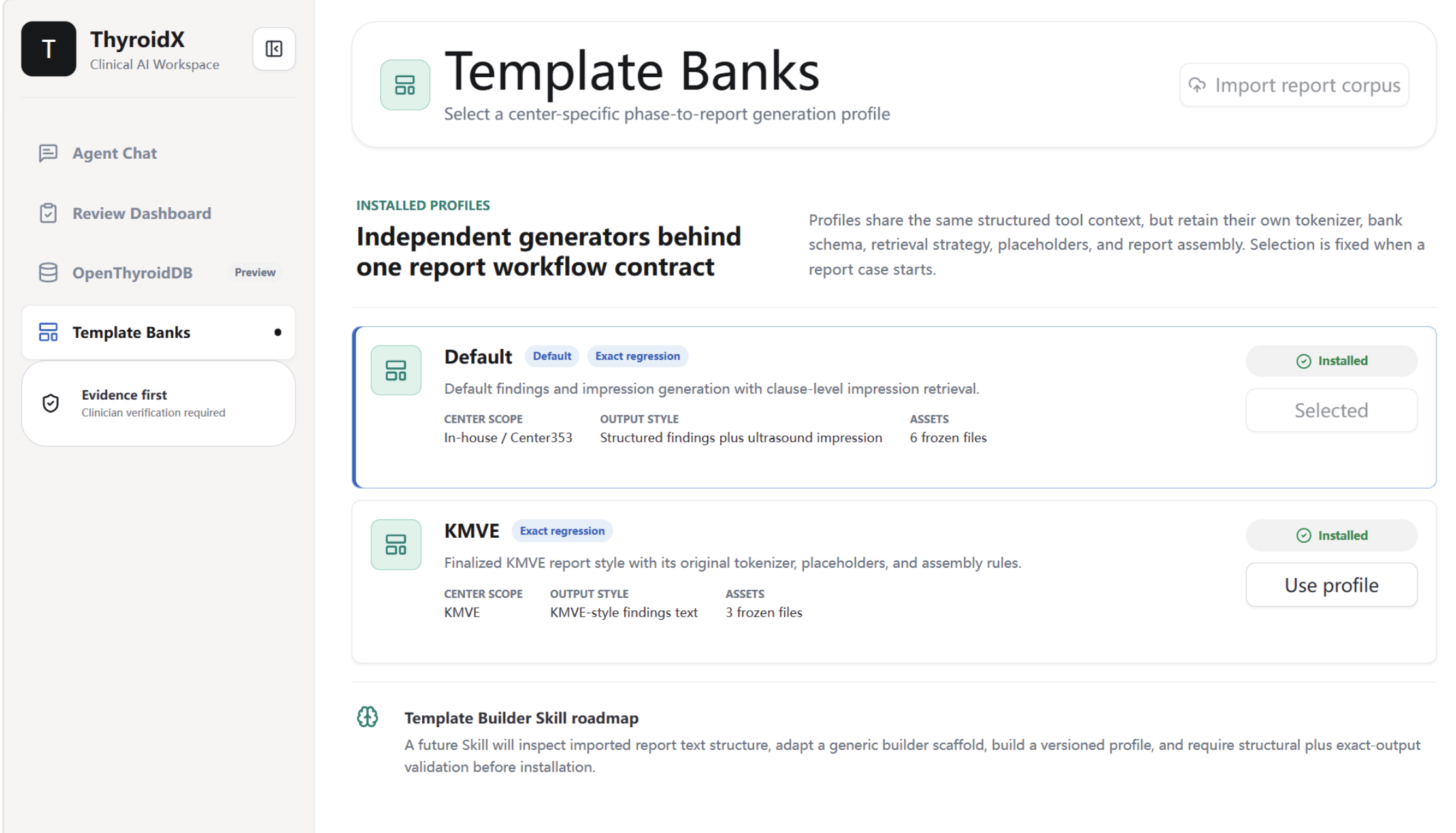}
    \caption{Template-bank construction for centre-specific report generation. The interface enables users to select an existing centre-specific report-generation profile or import a new report corpus. Generic template-construction scripts derive dataset-specific template banks to support adaptation to additional reporting data.}
    \label{fig:report_generation_template_bank}
\end{figure}

\begin{figure}[p]
    \centering
    \includegraphics[width=\textwidth,height=0.72\textheight,keepaspectratio]{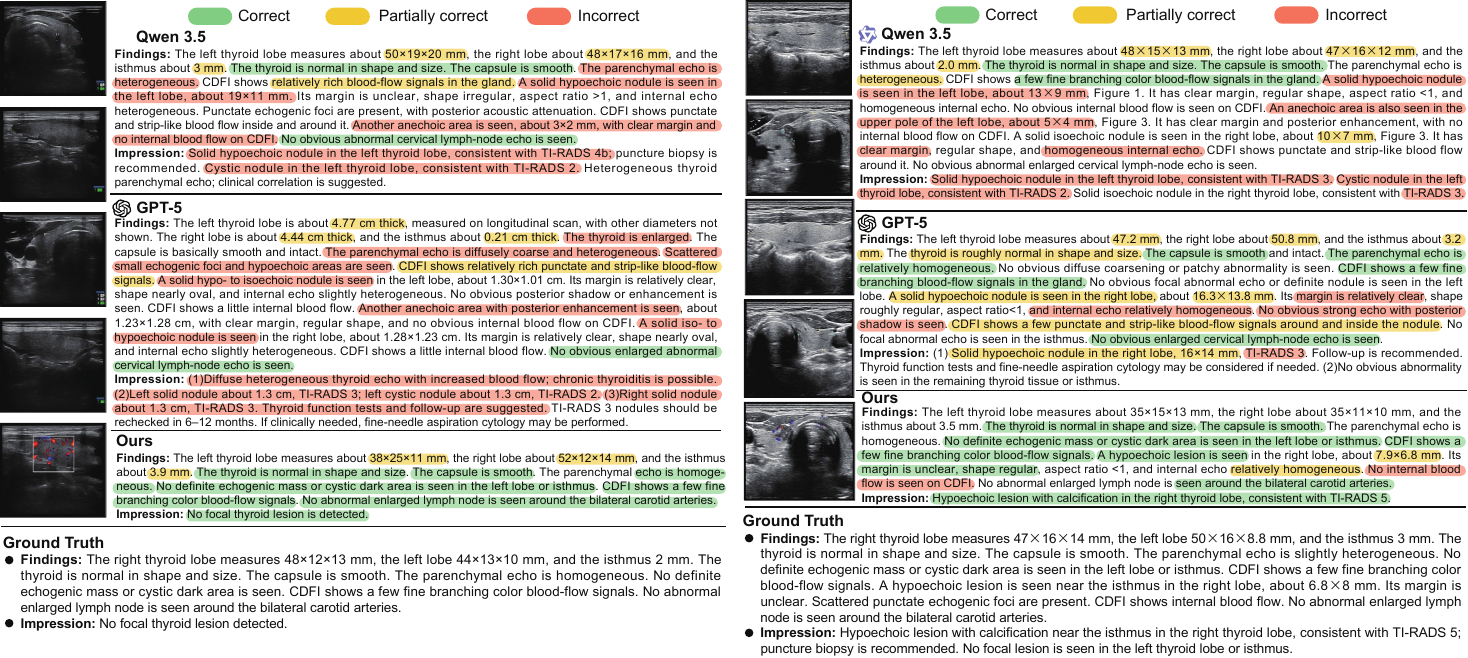}
    \caption{Qualitative comparison of thyroid ultrasound report generation. Representative benign and malignant cases compare reports generated by Qwen 3.5, GPT-5 and ThyroidXAgent with the reference reports. Text spans are annotated as clinically correct, partially correct or incorrect. The malignant example corresponds to Fig.~\ref{fig:reader_study}b; the benign example provides an additional complementary case.}
    \label{fig:RG_CaseReview}
\end{figure}

\suppcontentsline{fig:report_generation_multimodal_interface}{\hyperref[fig:report_generation_multimodal_interface]{Figure S4.}}{\hyperref[fig:report_generation_multimodal_interface]{Multimodal input processing and case interaction in ThyroidXAgent}}
\suppcontentsline{fig:report_generation_planning_execution_review}{\hyperref[fig:report_generation_planning_execution_review]{Figure S5.}}{\hyperref[fig:report_generation_planning_execution_review]{Planning, tool execution and clinician review in the report-generation workflow}}
\suppcontentsline{fig:report_generation_template_bank}{\hyperref[fig:report_generation_template_bank]{Figure S6.}}{\hyperref[fig:report_generation_template_bank]{Template-bank construction for centre-specific report generation}}
\suppcontentsline{fig:RG_CaseReview}{\hyperref[fig:RG_CaseReview]{Figure S7.}}{\hyperref[fig:RG_CaseReview]{Qualitative comparison of thyroid ultrasound report generation}}

\subsubsection*{Supplementary tables}

\suppcontentsline{tab:dataset_summary}{\hyperref[tab:dataset_summary]{Table S1.}}{\hyperref[tab:dataset_summary]{Composition and data splits of the multicentre thyroid ultrasound benchmark}}
\suppcontentsline{tab:dataset_comparison}{\hyperref[tab:dataset_comparison]{Table S2.}}{\hyperref[tab:dataset_comparison]{Characteristics of public and institutional thyroid ultrasound datasets}}

\suppcontentsline{tab:seg_performance}{\hyperref[tab:seg_performance]{Table S3.}}{\hyperref[tab:seg_performance]{Cross-dataset generalization for thyroid nodule segmentation}}
\suppcontentsline{tab:cls_performance}{\hyperref[tab:cls_performance]{Table S4.}}{\hyperref[tab:cls_performance]{Cross-dataset generalization for benign-malignant thyroid nodule classification}}
\suppcontentsline{tab:Malignant_images_tasks_performance}{\hyperref[tab:Malignant_images_tasks_performance]{Table S5.}}{\hyperref[tab:Malignant_images_tasks_performance]{Performance on malignant thyroid lesion stratification}}
\suppcontentsline{tab:stacked_performance}{\hyperref[tab:stacked_performance]{Table S6.}}{\hyperref[tab:stacked_performance]{Effects of cumulative training-data integration on segmentation and classification}}
\suppcontentsline{tab:gland_dice}{\hyperref[tab:gland_dice]{Table S7.}}{\hyperref[tab:gland_dice]{Centre-specific Dice for thyroid gland segmentation in NHC-MISD-TUS}}
\suppcontentsline{tab:gland_hd95}{\hyperref[tab:gland_hd95]{Table S8.}}{\hyperref[tab:gland_hd95]{Centre-specific HD95 for thyroid gland segmentation in NHC-MISD-TUS}}
\suppcontentsline{tab:nodule_dice}{\hyperref[tab:nodule_dice]{Table S9.}}{\hyperref[tab:nodule_dice]{Centre-specific Dice for thyroid nodule segmentation in NHC-MISD-TUS}}
\suppcontentsline{tab:nodule_hd95}{\hyperref[tab:nodule_hd95]{Table S10.}}{\hyperref[tab:nodule_hd95]{Centre-specific HD95 for thyroid nodule segmentation in NHC-MISD-TUS}}
\suppcontentsline{tab:binary_auroc}{\hyperref[tab:binary_auroc]{Table S11.}}{\hyperref[tab:binary_auroc]{Centre-specific AUROC for benign-malignant classification in NHC-MISD-TUS}}
\suppcontentsline{tab:binary_auprc}{\hyperref[tab:binary_auprc]{Table S12.}}{\hyperref[tab:binary_auprc]{Centre-specific AUPRC for benign-malignant classification in NHC-MISD-TUS}}

\suppcontentsline{tab:auxiliary_tools}{\hyperref[tab:auxiliary_tools]{Table S13.}}{\hyperref[tab:auxiliary_tools]{Performance of preprocessing and executor tools in ThyroidXAgent}}
\suppcontentsline{tab:report_generation_nlg_ci}{\hyperref[tab:report_generation_nlg_ci]{Table S14.}}{\hyperref[tab:report_generation_nlg_ci]{Lexical performance of thyroid ultrasound report generation across datasets}}
\suppcontentsline{tab:report_generation_clinical_ci}{\hyperref[tab:report_generation_clinical_ci]{Table S15.}}{\hyperref[tab:report_generation_clinical_ci]{Clinical semantic performance of thyroid ultrasound report generation across datasets}}
\suppcontentsline{tab:static_rule_controller_radar_metrics}{\hyperref[tab:static_rule_controller_radar_metrics]{Table S16.}}{\hyperref[tab:static_rule_controller_radar_metrics]{Static-pipeline metrics used for report-generation radar plots}}
\suppcontentsline{tab:report_generation_tool_ablation}{\hyperref[tab:report_generation_tool_ablation]{Table S17.}}{\hyperref[tab:report_generation_tool_ablation]{Ablation of tool integration for thyroid ultrasound report generation}}

\clearpage
\clearpage

\clearpage

\clearpage

\clearpage

\clearpage

\clearpage

\clearpage

\begin{table}[p]
\centering
\caption{Composition and data splits of the multicentre thyroid ultrasound benchmark. Numbers and percentages show the images contributed by each dataset to the full benchmark (n=38{,}864) and to the training (n=18{,}277), validation (n=850) and test (n=19{,}737) cohorts. DDTI, RJH-7K and ZJH-8K are independent external test cohorts. DDTI comprises 637 images, of which 349 carry benign-malignant classification labels. ZJH-8K serves as an external test set for segmentation and classification; its 4{,}756 malignant images additionally serve as the primary training set for malignant-lesion stratification (Supplementary Table~\ref{tab:Malignant_images_tasks_performance}).}
\label{tab:dataset_summary}
\renewcommand{\arraystretch}{1.2}
\setlength{\tabcolsep}{4pt}
\footnotesize
\resizebox{\textwidth}{!}{%
\begin{tabular}{llllcccc}
\toprule
\textbf{Dataset} &
\textbf{Task} &
\textbf{Center} &
\textbf{Ultrasound device} &
\textbf{Total} &
\makecell[c]{\textbf{Train}} &
\makecell[c]{\textbf{Valid}} &
\makecell[c]{\textbf{Test}} \\
\midrule
TN3K~\cite{gong2021multi} & \makecell[c]{Segmentation,\\Classification} & \makecell[l]{Zhujiang Hospital,\\Southern Medical\\University, Guangzhou,\\China} & \makecell[l]{GE Logiq E9,\\ARIETTA 850,\\RESONA 70B} & 5{,}347 (13.76\%) & 4{,}633 (25.35\%) & 100 (11.76\%) & 614 (3.11\%) \\
TN5K~\cite{zhang2025tn5000} & \makecell[c]{Segmentation,\\Classification} & \makecell[l]{Cancer Hospital,\\Chinese Academy of\\Medical Sciences,\\Beijing, China} & \makecell[l]{GE Logiq E9,\\GE S7 (5-12 MHz\\or 8-15 MHz)} & 5{,}000 (12.87\%) & 3{,}500 (19.15\%) & 500 (58.82\%) & 1{,}000 (5.07\%) \\
ThyroidXL~\cite{duong2025thyroidxl} & \makecell[c]{Segmentation,\\Classification} & \makecell[l]{Vietnam National\\Hospital of\\Endocrinology,\\Hanoi, Vietnam} & Hitachi Aloka Arietta V70 & 11{,}631 (29.93\%) & 9{,}441 (51.66\%) & 100 (11.76\%) & 2{,}090 (10.59\%) \\
PKTN~\cite{sun2025clip} & Segmentation & \makecell[l]{Peking University\\First Hospital,\\Beijing, China} & -- & 1{,}003 (2.58\%) & 703 (3.85\%) & 150 (17.65\%) & 150 (0.76\%) \\
\midrule
DDTI~\cite{pedraza2015open} & \makecell[c]{Segmentation,\\Classification} & \makecell[l]{IDIME,\\Bogot\'{a}, Colombia} & \makecell[l]{TOSHIBA Nemio 30,\\TOSHIBA Nemio MX\\(12 MHz probe)} & 637 (1.64\%) & -- & -- & \makecell[c]{637 (3.23\%) seg\\349 cls} \\
RJH-7K~\cite{shusharina2021segmentation} & Segmentation & \makecell[l]{Ruijin Hospital,\\Shanghai Jiao Tong\\University School of\\Medicine, Shanghai,\\China} & \makecell[l]{Different machines,\\not specified} & 7{,}288 (18.75\%) & -- & -- & 7{,}288 (36.93\%) \\
ZJH-8K & \makecell[c]{Segmentation,\\Classification} & \makecell[l]{Zhujiang Hospital,\\Southern Medical\\University, Guangzhou,\\China} & -- & 7{,}958 (20.48\%) & -- & -- & 7{,}958 (40.32\%) \\
\bottomrule
\end{tabular}%
}
\end{table}
\clearpage

\begin{table}[p]
\centering
\caption{Characteristics of public and institutional thyroid ultrasound datasets. Dataset size, data split, file format, task, geographical source and ultrasound scanner information are summarized for the included resources.}
\label{tab:dataset_comparison}
\resizebox{\textwidth}{!}{%
\begin{tabular}{@{}llllllll@{}}
\toprule
\textbf{Dataset} &
\textbf{Dataset Size} &
\textbf{Development/Validation} &
\textbf{Evaluation} &
\textbf{File Format} &
\textbf{Task} &
\textbf{Location} &
\textbf{Ultrasonic Imaging Device} \\
\midrule

TGVideo~\cite{wunderling2017comparison} &
15,186 (16 cases) &
15,186 &
N/A &
\begin{tabular}[c]{@{}l@{}}
Image: DICOM\\
Mask: DICOM
\end{tabular} &
Segmentation &
Germany &
GE Logiq E9 \\

\midrule
DDTI~\cite{pedraza2015open} &
637 &
N/A &
637 &
\begin{tabular}[c]{@{}l@{}}
Image: PNG\\
Mask: PNG\\
Label: CSV
\end{tabular} &
\begin{tabular}[c]{@{}l@{}}
Segmentation\\
Classification
\end{tabular} &
Colombia &
\begin{tabular}[c]{@{}l@{}}
TOSHIBA Nemio 30\\
TOSHIBA Nemio MX
\end{tabular} \\

\midrule
TN3K~\cite{gong2021multi,gong2022less,gong2023thyroid} &
5,347 &
4,733 &
614 &
\begin{tabular}[c]{@{}l@{}}
Image: JPG\\
Mask: JPG\\
Label: CSV
\end{tabular} &
\begin{tabular}[c]{@{}l@{}}
Segmentation\\
Classification
\end{tabular} &
Guangzhou, China &
\begin{tabular}[c]{@{}l@{}}
GE Logiq E9\\
ARIETTA 850\\
RESONA 70B
\end{tabular} \\

\midrule
TN5K~\cite{zhang2025tn5000} &
5,000 &
4,000 &
1,000 &
\begin{tabular}[c]{@{}l@{}}
Image: JPG\\
Label: XML
\end{tabular} &
\begin{tabular}[c]{@{}l@{}}
Detection\\
Classification
\end{tabular} &
Beijing, China &
\begin{tabular}[c]{@{}l@{}}
GE Logiq E9\\
GE S7
\end{tabular} \\

\midrule
ThyUS2Path~\cite{hou2024ultrasonography} &
8,508 &
5,457 &
3,051 &
\begin{tabular}[c]{@{}l@{}}
Image: JPG\\
Label: CSV
\end{tabular} &
Classification &
Zhejiang, China &
Esaote MyLab (Portable) \\

\midrule
Cine-clip~\cite{stanford2024thyroid} &
\begin{tabular}[c]{@{}l@{}}
17,412 frames\\
192 cases\\
avg. 90 frames/case
\end{tabular} &
N/A &
N/A &
\begin{tabular}[c]{@{}l@{}}
Image: HDF5\\
Label: CSV
\end{tabular} &
\begin{tabular}[c]{@{}l@{}}
Segmentation\\
Classification
\end{tabular} &
California, USA &
N/A \\

\midrule
AHU~\cite{yang2025annotated} &
\begin{tabular}[c]{@{}l@{}}
1,833 cases\\
125,896 images
\end{tabular} &
N/A &
N/A &
\begin{tabular}[c]{@{}l@{}}
Image: JPG\\
Label: Folder-level
\end{tabular} &
Classification &
China (web scraping) &
Heterogeneous \\

\midrule
ThyroidXL~\cite{duong2025thyroidxl} &
11,631 &
9,541 &
2,090 &
\begin{tabular}[c]{@{}l@{}}
Image: PNG\\
Mask: PNG\\
Label: TXT
\end{tabular} &
\begin{tabular}[c]{@{}l@{}}
Segmentation\\
Classification\\
Detection
\end{tabular} &
Vietnam &
Hitachi Aloka Arietta V70 \\

\midrule
PKTN~\cite{sun2025clip} &
1,003 &
N/A &
N/A &
\begin{tabular}[c]{@{}l@{}}
Image: JPG\\
Mask: JPG
\end{tabular} &
Segmentation &
Beijing, China &
N/A \\

\midrule
KMVE~\cite{li_ultrasound_2024} &
\begin{tabular}[c]{@{}l@{}}
2,457 cases\\
4,914 image assignments
\end{tabular} &
\begin{tabular}[c]{@{}l@{}}
1,719 training cases\\
246 validation cases
\end{tabular} &
492 test cases &
\begin{tabular}[c]{@{}l@{}}
Image: JPEG\\
Report: JSON
\end{tabular} &
Report Generation &
Beijing, China &
N/A \\

\midrule
SMU-HMC &
\begin{tabular}[c]{@{}l@{}}
23,955 reports\\
248,194 images
\end{tabular} &
\begin{tabular}[c]{@{}l@{}}
23,555 reports\\
242,210 images
\end{tabular} &
\begin{tabular}[c]{@{}l@{}}
400 reports\\
5,984 images
\end{tabular} &
\begin{tabular}[c]{@{}l@{}}
Image: PNG\\
Report: JSON
\end{tabular} &
\begin{tabular}[c]{@{}l@{}}
Report Generation\\
Component Development
\end{tabular} &
Guangzhou, China &
Heterogeneous \\

\midrule
ZJH-TS &
\begin{tabular}[c]{@{}l@{}}
353 reports\\
4,471 images
\end{tabular} &
N/A &
\begin{tabular}[c]{@{}l@{}}
150 reports\\
2,034 images
\end{tabular} &
\begin{tabular}[c]{@{}l@{}}
Image: PNG\\
Report: JSON
\end{tabular} &
\begin{tabular}[c]{@{}l@{}}
Report Generation\\
External Validation
\end{tabular} &
Guangzhou, China &
Heterogeneous \\

\midrule
TNVideo &
148 cases &
N/A &
145 labelled cases &
Video: AVI &
\begin{tabular}[c]{@{}l@{}}
Segmentation\\
Reader Study
\end{tabular} &
Guangzhou, China &
N/A \\

\bottomrule
\end{tabular}%
}
\end{table}
\clearpage

\begin{table*}[p]
\centering
\begin{threeparttable}

\caption{Cross-dataset generalization for thyroid nodule segmentation. Dice coefficient and 95th-percentile Hausdorff distance (HD95) are reported with 95\% confidence intervals.}
\label{tab:seg_performance}

\footnotesize
\setlength{\tabcolsep}{5.5pt}
\renewcommand{\arraystretch}{1.08}

\begin{tabular}{lccccccc}
\toprule
\textbf{Model}
& \textbf{TN3K}
& \textbf{ThyroidXL}
& \textbf{PKTN}
& \textbf{TN5K}
& \textbf{DDTI}
& \textbf{ZJH-8K}
& \textbf{RJH-7K} \\
\midrule
\multicolumn{8}{l}{\textit{Dice (\%) $\uparrow$}} \\
\midrule
TransUnet~\cite{chen2024transunet}
& $81.84 \pm 1.62$
& $85.75 \pm 0.57$
& $76.89 \pm 3.56$
& $78.54 \pm 1.51$
& $76.58 \pm 1.62$
& $80.72 \pm 0.97$
& $84.83 \pm 0.37$ \\
MedSegX~\cite{zhang2025generalist}
& $83.93 \pm 0.79$
& $79.98 \pm 0.36$
& $80.63 \pm 0.42$
& $83.10 \pm 0.48$
& $75.12 \pm 1.68$
& $84.06 \pm 0.39$
& $85.40 \pm 0.18$ \\
MedSAM2~\cite{ma2025medsam2}
& $84.47 \pm 1.02$
& $86.94 \pm 0.36$
& \textbf{83.46 $\pm$ 2.60}
& $83.03 \pm 1.29$
& $84.72 \pm 1.26$
& $86.29 \pm 0.73$
& $90.72 \pm 0.21$ \\
UltraFedFM~\cite{jiang2025pretraining}
& $81.18 \pm 1.46$
& $84.70 \pm 0.53$
& $75.31 \pm 1.12$
& $77.13 \pm 1.38$
& $75.57 \pm 1.67$
& $80.64 \pm 0.84$
& $83.10 \pm 0.33$ \\
\rowcolor{lightgray}
\textbf{ThyroidXAgent}
& \textbf{85.28 $\pm$ 1.28}
& \textbf{87.58 $\pm$ 0.44}
& $82.99 \pm 2.10$
& \textbf{83.26 $\pm$ 1.34}
& \textbf{85.62 $\pm$ 1.07}
& \textbf{94.30 $\pm$ 0.38} 
& \textbf{91.46 $\pm$ 0.14} \\

\midrule
\multicolumn{8}{l}{\textit{HD95 (mm) $\downarrow$}} \\
\midrule
TransUnet~\cite{chen2024transunet}
& $27.27 \pm 5.52$
& $22.42 \pm 1.34$
& $26.88 \pm 9.66$
& $22.32 \pm 3.43$
& $17.12 \pm 1.55$
& $18.37 \pm 0.75$
& $18.81 \pm 0.74$ \\
MedSegX~\cite{zhang2025generalist}
& $10.95 \pm 0.64$
& $11.07 \pm 0.32$
& $10.83 \pm 0.70$
& $11.76 \pm 0.76$
& $18.39 \pm 1.65$
& $10.96 \pm 0.35$
& $9.37 \pm 0.18$ \\
MedSAM2~\cite{ma2025medsam2}
& $11.51 \pm 1.53$
& $5.46 \pm 0.44$
& $10.56 \pm 3.64$
& $10.94 \pm 1.12$
& $10.06 \pm 1.21$
& $6.79 \pm 0.57$
& $2.92 \pm 0.17$ \\
UltraFedFM~\cite{jiang2025pretraining}
& $14.98 \pm 2.10$
& $8.10 \pm 0.58$
& $16.08 \pm 1.67$
& $14.96 \pm 1.65$
& $18.12 \pm 1.47$
& $8.69 \pm 0.80$
& $9.06 \pm 0.38$ \\
\rowcolor{lightgray}
\textbf{ThyroidXAgent}
& \textbf{10.31 $\pm$ 1.70}
& \textbf{5.43 $\pm$ 0.53}
& \textbf{9.01 $\pm$ 3.58}
& \textbf{10.12 $\pm$ 1.23}
& \textbf{9.24 $\pm$ 1.07}
& \textbf{2.25 $\pm$ 0.39}
& \textbf{1.92 $\pm$ 0.08} \\

\bottomrule
\end{tabular}
\end{threeparttable}
\end{table*}
\clearpage

\begin{table*}[p]
\centering
\begin{threeparttable}

\caption{Cross-dataset generalization for benign-malignant thyroid nodule classification. Area under the receiver operating characteristic curve (AUROC) and area under the precision-recall curve (AUPRC) are reported with 95\% confidence intervals.}
\label{tab:cls_performance}

\footnotesize
\setlength{\tabcolsep}{5.5pt}
\renewcommand{\arraystretch}{1.08}

\begin{tabular}{lccccc}
\toprule
\textbf{Method}
& \textbf{TN3K}
& \textbf{ThyroidXL}
& \textbf{TN5K}
& \textbf{DDTI}
& \textbf{ZJH-8K} \\
\midrule
\multicolumn{6}{l}{\textit{AUROC $\uparrow$}} \\
\midrule
ResNet-50~\cite{he2016deep}
& $0.7674 \pm 0.0394$
& $0.9044 \pm 0.0118$
& $0.9322 \pm 0.0168$
& $0.6704 \pm 0.0842$
& $0.6704 \pm 0.0842$ \\
RepViT~\cite{wang2023repvit}
& $0.5556 \pm 0.0463$
& $0.7774 \pm 0.0188$
& $0.6603 \pm 0.0375$
& $0.6162 \pm 0.0804$
& $0.8538 \pm 0.0185$\\
LSNet~\cite{wang2025lsnet}
& $0.8095 \pm 0.0333$
& $0.9178 \pm 0.0114$
& $0.9091 \pm 0.0201$
& $0.7581 \pm 0.0658$
& $0.8631 \pm 0.0201$\\
UltraFedFM~\cite{jiang2025pretraining}
& $0.8461 \pm 0.0697$
& $0.9239 \pm 0.0104$
& $0.9298 \pm 0.0175$
& $0.7518 \pm 0.1712$
& $0.9115 \pm 0.0140$\\
MedGemma~\cite{sellergren2025medgemma}
& $0.8492 \pm 0.0305$
& $0.9371 \pm 0.0095$
& $0.9442 \pm 0.0156$
& \textbf{0.8255 $\pm$ 0.0650}
& $0.8976 \pm 0.0166$\\
Qwen3-VL-8B-Instruct~\cite{bai2025qwen3}
& $0.8237 \pm 0.0328$
& $0.9050 \pm 0.0115$
& $0.9214 \pm 0.0187$
& $0.7361 \pm 0.0692$
& $0.8659 \pm 0.0189$\\
GPT-5~\cite{openai2025gpt5systemcard}
& $0.6924 \pm 0.0421$
& $0.7059 \pm 0.0469$
& $0.7737 \pm 0.0996$
& $0.6346 \pm 0.0914$
& $0.6109 \pm 0.0515$\\
Gemini-2.5-Pro~\cite{comanici_gemini_2025}
& $0.6587 \pm 0.0455$
& $0.6246 \pm 0.0640$
& $0.6873 \pm 0.0691$
& $0.6156 \pm 0.1308$
& $0.6493 \pm 0.0516$\\
\rowcolor{lightgray}
\textbf{ThyroidXAgent}
& \textbf{0.8692 $\pm$ 0.0349}
& \textbf{0.9676 $\pm$ 0.0066}
& \textbf{0.9472 $\pm$ 0.0152}
& $0.7991 \pm 0.0741$
& \textbf{0.9175 $\pm$ 0.0167}\\
\midrule
\multicolumn{6}{l}{\textit{AUPRC $\uparrow$}} \\
\midrule
ResNet-50~\cite{he2016deep}
& $0.6882 \pm 0.0632$
& $0.8882 \pm 0.0174$
& $0.9674 \pm 0.0268$
& $0.3755 \pm 0.1176$
& $0.2755 \pm 0.1167$ \\
RepViT~\cite{wang2023repvit}
& $0.4275 \pm 0.0528$
& $0.7161 \pm 0.0276$
& $0.8403 \pm 0.0216$
& $0.3924 \pm 0.0933$
& $0.9486 \pm 0.0078$\\
LSNet~\cite{wang2025lsnet}
& $0.7581 \pm 0.0452$
& $0.9040 \pm 0.0142$
& $0.9551 \pm 0.0134$
& $0.4180 \pm 0.1410$
& $0.9449 \pm 0.0113$\\
UltraFedFM~\cite{jiang2025pretraining}
& $0.8531 \pm 0.0284$
& $0.9354 \pm 0.0114$
& $0.8422 \pm 0.0421$
& $0.4487 \pm 0.1452$
& $0.9669 \pm 0.0084$\\
MedGemma~\cite{sellergren2025medgemma}
& $0.8047 \pm 0.0430$
& $0.9201 \pm 0.0139$
& $0.9747 \pm 0.0084$
& $0.5537 \pm 0.1663$
& $0.9589 \pm 0.0096$\\
Qwen3-VL-8B-Instruct~\cite{bai2025qwen3}
& $0.7617 \pm 0.0511$
& $0.8787 \pm 0.0379$
& $0.9636 \pm 0.0106$
& $0.4112 \pm 0.1415$
& $0.9498 \pm 0.0096$\\
GPT-5~\cite{openai2025gpt5systemcard}
& $0.6627 \pm 0.0633$
& $0.6237 \pm 0.0666$
& $0.8920 \pm 0.0316$
& $0.3578 \pm 0.1089$
& $0.8311 \pm 0.0377$\\
Gemini-2.5-Pro~\cite{comanici_gemini_2025}
& $0.6205 \pm 0.0587$
& $0.4914 \pm 0.0841$
& $0.8462 \pm 0.0446$
& $0.3924 \pm 0.1527$
& $0.8403 \pm 0.0362$\\
\rowcolor{lightgray}
\textbf{ThyroidXAgent}
& \textbf{0.8545 $\pm$ 0.0600}
& \textbf{0.9653 $\pm$ 0.0078}
& \textbf{0.9752 $\pm$ 0.0089}
& \textbf{0.5863 $\pm$ 0.1380}
& \textbf{0.9711 $\pm$ 0.0006}\\
\bottomrule
\end{tabular}
\end{threeparttable}
\end{table*}
\clearpage

\begin{table*}[p]
\centering
\begin{threeparttable}

\caption{Performance on malignant thyroid lesion stratification. AUROC and AUPRC with 95\% confidence intervals are reported for lateral lymph-node metastasis prediction and follicular versus papillary thyroid carcinoma subtype classification. Training data comprised 4{,}756 malignant images from ZJH-8K, of which 20 cases (183 images) were held out for validation. Em dashes indicate tasks that were not evaluated.}
\label{tab:Malignant_images_tasks_performance}

\footnotesize
\setlength{\tabcolsep}{5.5pt}
\renewcommand{\arraystretch}{1.08}

\begin{tabular}{lcccc}
\toprule
\multirow{2}{*}{\textbf{Method}}
& \multicolumn{2}{c}{\textbf{Lymph Node Metastasis}}
& \multicolumn{2}{c}{\textbf{FTC/PTC subtype}} \\
& AUROC $\uparrow$ & AUPRC $\uparrow$ & AUROC $\uparrow$ & AUPRC $\uparrow$ \\
\midrule
RepViT~\cite{wang2023repvit}
& $0.7905 \pm 0.0676$
& $0.8152 \pm 0.0638$
& $0.6419 \pm 0.0839$
& $0.6297 \pm 0.0942$\\
LSNet~\cite{wang2025lsnet}
& $0.5878 \pm 0.0865$
& $0.6301 \pm 0.0875$
& $0.4858 \pm 0.0925$
& $0.4845 \pm 0.0908$\\
UltraFedFM~\cite{jiang2025pretraining}
& $0.7757 \pm 0.0731$
& $0.7902 \pm 0.0845$
& $0.7365 \pm 0.0744$
& $0.7582 \pm 0.0824$\\
MedGemma~\cite{sellergren2025medgemma}
& $0.8403 \pm 0.0461$
& $0.8585 \pm 0.0566$
& $0.6598 \pm 0.0824$
& $0.6142 \pm 0.1056$\\
Qwen3-VL-8B-Instruct~\cite{bai2025qwen3}
& $0.8070 \pm 0.0632$
& $0.8055 \pm 0.0800$
& $0.6056 \pm 0.0866$
& $0.5539 \pm 0.1118$\\
GPT-5~\cite{openai2025gpt5systemcard}
& $0.8410 \pm 0.0575$
& $0.8629 \pm 0.0533$
& $0.1604 \pm 0.0706$
& $0.3638 \pm 0.0847$\\
Gemini-2.5-Pro~\cite{comanici_gemini_2025}
& $0.5414 \pm 0.0736$
& $0.5492 \pm 0.0915$
& $0.3324 \pm 0.0872$
& $0.4187 \pm 0.0837$\\
LLNM-Net~\cite{Shen2025NatCommunLLNM}
& $0.7665 \pm 0.0692$
& $0.7363 \pm 0.0849$
& --
& --\\
Tiger-Model~\cite{Dai2025NatCommunThyroidSubtype}
& --
& --
& $0.7136 \pm 0.0814$
& $0.7117 \pm 0.1101$\\
\rowcolor{lightgray}
\textbf{ThyroidXAgent}
& \textbf{0.8642 $\pm$ 0.0550}
& \textbf{0.8808 $\pm$ 0.0537}
& \textbf{0.8053 $\pm$ 0.0599}
& \textbf{0.7863 $\pm$ 0.0793}\\
\bottomrule
\end{tabular}
\end{threeparttable}
\end{table*}
\clearpage

\begin{table*}[p]
\centering

\caption{Effects of cumulative training-data integration on segmentation and classification. Models were trained on progressively expanded configurations and evaluated on independent test sets. Segmentation performance is reported as Dice coefficient (\%, $\uparrow$ higher is better) and HD95 (mm, $\downarrow$ lower is better); classification performance as AUROC and AUPRC ($\uparrow$ higher is better). Values are means $\pm$ 95\% confidence intervals across five independent runs. Dashes indicate that the test set was not applicable for the given task. Segmentation training configurations: dataset1 (TN3K), dataset2 (TN3K + ThyroidXL), dataset3 (TN3K + ThyroidXL + PKTN), dataset4 (TN3K + ThyroidXL + PKTN + TN5K). Classification training configurations: dataset1 (TN3K), dataset2 (TN3K + ThyroidXL), dataset3 (TN3K + ThyroidXL + TN5K).}
\label{tab:stacked_performance}

\footnotesize
\setlength{\tabcolsep}{5pt}
\renewcommand{\arraystretch}{1.08}

\resizebox{\textwidth}{!}{%
\begin{tabular}{lccccccc}
\toprule
\diagbox{\textbf{Train}}{\textbf{Test}}
& \textbf{TN3K}
& \textbf{ThyroidXL}
& \textbf{PKTN}
& \textbf{TN5K}
& \textbf{DDTI}
& \textbf{ZJH-8K}
& \textbf{RJH-7K} \\
\midrule
\multicolumn{8}{l}{\textit{Segmentation -- Dice (\%) $\uparrow$}} \\
\midrule
dataset1
& \textbf{82.76 $\pm$ 3.54}
& 81.97 $\pm$ 2.63
& 79.21 $\pm$ 2.71
& 72.18 $\pm$ 5.08
& 78.08 $\pm$ 3.09
& 94.57 $\pm$ 0.43
& 80.77 $\pm$ 0.44 \\
dataset2
& 81.63 $\pm$ 3.81
& 86.84 $\pm$ 1.90
& 81.73 $\pm$ 2.26
& 71.07 $\pm$ 5.35
& 76.72 $\pm$ 3.39
& \textbf{94.85 $\pm$ 0.40}
& 82.38 $\pm$ 0.42 \\
dataset3
& 80.81 $\pm$ 3.82
& 86.00 $\pm$ 2.31
& 81.91 $\pm$ 2.26
& 72.82 $\pm$ 4.91
& 84.81 $\pm$ 2.43
& 94.82 $\pm$ 0.39
& 91.44 $\pm$ 0.15 \\
dataset4
& 81.86 $\pm$ 3.70
& \textbf{86.97 $\pm$ 2.19}
& \textbf{83.28 $\pm$ 2.19}
& \textbf{82.57 $\pm$ 3.46}
& \textbf{84.86 $\pm$ 2.38}
& 94.77 $\pm$ 0.39
& \textbf{91.46 $\pm$ 0.15} \\
\midrule
\multicolumn{8}{l}{\textit{Segmentation -- HD95 (mm) $\downarrow$}} \\
\midrule
dataset1
& \textbf{13.49 $\pm$ 3.83}
& 8.34 $\pm$ 2.34
& 11.73 $\pm$ 3.07
& 11.37 $\pm$ 3.49
& 16.93 $\pm$ 3.12
& 2.30 $\pm$ 0.47
& 11.46 $\pm$ 0.50 \\
dataset2
& 15.92 $\pm$ 4.58
& 4.99 $\pm$ 1.58
& 9.72 $\pm$ 2.71
& 13.64 $\pm$ 4.28
& 18.54 $\pm$ 3.17
& 1.98 $\pm$ 0.41
& 9.65 $\pm$ 0.44 \\
dataset3
& 15.94 $\pm$ 4.34
& 5.46 $\pm$ 1.62
& 10.92 $\pm$ 3.52
& 11.07 $\pm$ 3.28
& 11.97 $\pm$ 3.98
& \textbf{1.93 $\pm$ 0.38}
& \textbf{1.87 $\pm$ 0.07} \\
dataset4
& 17.00 $\pm$ 5.34
& \textbf{4.74 $\pm$ 1.42}
& \textbf{8.89 $\pm$ 2.93}
& \textbf{4.64 $\pm$ 1.43}
& \textbf{9.91 $\pm$ 2.18}
& 2.07 $\pm$ 0.40
& 1.88 $\pm$ 0.07 \\
\midrule
\multicolumn{8}{l}{\textit{Classification -- AUROC $\uparrow$}} \\
\midrule
dataset1
& 0.7666 $\pm$ 0.03
& 0.8713 $\pm$ 0.01
& --
& 0.8272 $\pm$ 0.02
& 0.7244 $\pm$ 0.07
& 0.9924 $\pm$ 0.01
& -- \\
dataset2
& 0.7724 $\pm$ 0.04
& 0.9254 $\pm$ 0.01
& --
& 0.8151 $\pm$ 0.02
& 0.5762 $\pm$ 0.10
& 0.9932 $\pm$ 0.01
& -- \\
dataset3
& \textbf{0.7906 $\pm$ 0.03}
& \textbf{0.9288 $\pm$ 0.01}
& --
& \textbf{0.9515 $\pm$ 0.01}
& \textbf{0.7623 $\pm$ 0.07}
& \textbf{0.9937 $\pm$ 0.01}
& -- \\
\midrule
\multicolumn{8}{l}{\textit{Classification -- AUPRC $\uparrow$}} \\
\midrule
dataset1
& 0.6806 $\pm$ 0.06
& 0.8147 $\pm$ 0.03
& --
& 0.9151 $\pm$ 0.02
& 0.3578 $\pm$ 0.13
& 0.9951 $\pm$ 0.01
& -- \\
dataset2
& \textbf{0.7237 $\pm$ 0.05}
& 0.9140 $\pm$ 0.01
& --
& 0.9074 $\pm$ 0.01
& 0.3190 $\pm$ 0.14
& \textbf{0.9968 $\pm$ 0.01}
& -- \\
dataset3
& 0.7188 $\pm$ 0.05
& \textbf{0.9144 $\pm$ 0.01}
& --
& \textbf{0.9803 $\pm$ 0.01}
& \textbf{0.4029 $\pm$ 0.14}
& 0.9967 $\pm$ 0.01
& -- \\
\bottomrule
\end{tabular}%
}
\end{table*}
\clearpage



\begin{table}[p]
\centering
\caption{Per-centre Dice similarity coefficient (\%) for thyroid gland segmentation on the NHC-MISD-TUS external test set. Values are reported as point estimates with 95\% confidence intervals. Bold indicates the best result in each row.}
\label{tab:gland_dice}
\footnotesize
\setlength{\tabcolsep}{4pt}
\begin{tabular}{llrrrrr}
\toprule
\multirow{2}{*}{Center} & \multirow{2}{*}{$N$} & \multicolumn{1}{c}{ThyroidXAgent} & \multicolumn{1}{c}{MedSAM2} & \multicolumn{1}{c}{MedSegX} & \multicolumn{1}{c}{TransUNet} & \multicolumn{1}{c}{UltraFedFM} \\
\cmidrule(lr){3-3} \cmidrule(lr){4-4} \cmidrule(lr){5-5} \cmidrule(lr){6-6} \cmidrule(lr){7-7}
 & & Dice (\%) $\uparrow$ & Dice (\%) $\uparrow$ & Dice (\%) $\uparrow$ & Dice (\%) $\uparrow$ & Dice (\%) $\uparrow$ \\
\midrule
\textit{Overall} & 8,721 & 59.28\textsubscript{(58.70, 59.86)} & 56.80\textsubscript{(56.35, 57.24)} & 54.98\textsubscript{(54.57, 55.39)} & 53.56\textsubscript{(53.06, 54.12)} & 33.30\textsubscript{(32.73, 33.88)} \\
\midrule
THYB\_S\_ZJ24 & 1,174 & \textbf{80.64}\textsubscript{(79.84, 81.40)} & 52.96\textsubscript{(51.96, 54.03)} & 50.32\textsubscript{(49.25, 51.42)} & 72.91\textsubscript{(72.03, 73.79)} & 8.49\textsubscript{(7.59, 9.42)} \\
THYB\_S\_EN04 & 949 & \textbf{65.30}\textsubscript{(63.82, 66.78)} & 51.05\textsubscript{(49.46, 52.51)} & 59.83\textsubscript{(58.66, 60.96)} & 59.65\textsubscript{(58.11, 61.15)} & 35.33\textsubscript{(33.81, 36.91)} \\
THYB\_S\_BJ01 & 876 & 61.05\textsubscript{(59.14, 62.88)} & \textbf{61.22}\textsubscript{(59.93, 62.48)} & 51.36\textsubscript{(49.83, 52.83)} & 56.68\textsubscript{(54.95, 58.35)} & 27.42\textsubscript{(25.73, 29.23)} \\
THYB\_S\_SH01 & 832 & 43.33\textsubscript{(41.36, 45.33)} & \textbf{64.88}\textsubscript{(63.37, 66.28)} & 55.60\textsubscript{(54.14, 56.95)} & 42.72\textsubscript{(40.94, 44.52)} & 49.82\textsubscript{(48.11, 51.48)} \\
THYB\_S\_ZJ05 & 694 & \textbf{55.25}\textsubscript{(53.12, 57.40)} & 54.83\textsubscript{(53.19, 56.51)} & 53.77\textsubscript{(52.33, 55.13)} & 50.71\textsubscript{(48.70, 52.55)} & 30.49\textsubscript{(28.66, 32.35)} \\
THYB\_S\_SH05 & 669 & \textbf{61.58}\textsubscript{(59.74, 63.48)} & 59.92\textsubscript{(58.52, 61.38)} & 56.63\textsubscript{(55.27, 57.98)} & 50.26\textsubscript{(48.46, 52.13)} & 34.59\textsubscript{(32.66, 36.40)} \\
THYB\_S\_NX01 & 582 & 52.69\textsubscript{(50.28, 54.89)} & 45.59\textsubscript{(44.00, 47.18)} & 52.80\textsubscript{(51.32, 54.40)} & \textbf{54.11}\textsubscript{(51.83, 56.22)} & 32.64\textsubscript{(30.64, 34.63)} \\
THYB\_S\_ZJ06 & 460 & 48.58\textsubscript{(46.12, 50.98)} & \textbf{62.74}\textsubscript{(60.83, 64.42)} & 55.20\textsubscript{(53.35, 56.97)} & 42.03\textsubscript{(39.70, 44.34)} & 35.19\textsubscript{(32.95, 37.35)} \\
THYB\_S\_QX07 & 289 & \textbf{66.05}\textsubscript{(63.24, 68.91)} & 50.61\textsubscript{(48.09, 53.10)} & 59.57\textsubscript{(57.62, 61.64)} & 62.24\textsubscript{(59.24, 65.06)} & 46.38\textsubscript{(43.64, 49.40)} \\
THYB\_S\_AN01 & 243 & 36.90\textsubscript{(33.32, 40.25)} & \textbf{74.15}\textsubscript{(71.96, 76.34)} & 56.14\textsubscript{(53.14, 59.07)} & 36.30\textsubscript{(33.35, 39.72)} & 60.76\textsubscript{(57.91, 63.56)} \\
THYB\_S\_CQ03 & 238 & \textbf{63.69}\textsubscript{(60.50, 66.62)} & 55.01\textsubscript{(52.63, 57.49)} & 58.52\textsubscript{(56.01, 60.89)} & 46.06\textsubscript{(42.68, 49.56)} & 33.93\textsubscript{(30.82, 37.30)} \\
THYB\_S\_GZ02 & 233 & 59.92\textsubscript{(56.20, 63.38)} & \textbf{62.23}\textsubscript{(60.11, 64.35)} & 55.84\textsubscript{(53.35, 58.40)} & 54.40\textsubscript{(51.00, 58.00)} & 37.71\textsubscript{(34.20, 41.27)} \\
THYB\_S\_JX06 & 215 & 54.16\textsubscript{(50.51, 58.01)} & \textbf{63.46}\textsubscript{(60.77, 65.99)} & 55.71\textsubscript{(52.48, 58.52)} & 45.96\textsubscript{(42.07, 49.54)} & 46.71\textsubscript{(43.30, 50.23)} \\
THYB\_S\_JS02 & 190 & 36.44\textsubscript{(32.39, 40.43)} & \textbf{64.41}\textsubscript{(61.96, 66.78)} & 56.60\textsubscript{(54.36, 58.91)} & 34.57\textsubscript{(31.09, 38.09)} & 50.48\textsubscript{(46.08, 54.64)} \\
THYB\_S\_YN05 & 137 & 55.44\textsubscript{(50.24, 60.32)} & 52.03\textsubscript{(48.59, 55.63)} & \textbf{63.75}\textsubscript{(60.52, 66.75)} & 63.10\textsubscript{(58.34, 67.44)} & 14.28\textsubscript{(10.73, 18.14)} \\
THYB\_S\_EN02 & 120 & \textbf{55.75}\textsubscript{(50.96, 60.59)} & 39.06\textsubscript{(36.22, 42.01)} & 51.87\textsubscript{(48.88, 54.83)} & 38.76\textsubscript{(33.77, 43.55)} & 40.34\textsubscript{(35.64, 44.90)} \\
THYB\_S\_AH04 & 101 & \textbf{65.76}\textsubscript{(59.94, 70.88)} & 52.92\textsubscript{(49.38, 56.65)} & 61.94\textsubscript{(58.10, 65.51)} & 56.38\textsubscript{(50.58, 61.47)} & 47.68\textsubscript{(41.39, 54.39)} \\
THYB\_S\_GS03 & 94 & 50.96\textsubscript{(45.22, 56.84)} & \textbf{59.17}\textsubscript{(54.56, 63.60)} & 51.74\textsubscript{(47.42, 55.76)} & 48.24\textsubscript{(42.53, 53.76)} & 40.09\textsubscript{(34.46, 45.45)} \\
THYB\_S\_FJ03 & 85 & \textbf{66.63}\textsubscript{(60.76, 72.05)} & 46.90\textsubscript{(43.13, 50.64)} & 54.13\textsubscript{(50.57, 57.52)} & 50.83\textsubscript{(44.94, 56.45)} & 36.40\textsubscript{(30.87, 41.92)} \\
THYB\_S\_SD12 & 75 & \textbf{69.59}\textsubscript{(64.15, 74.62)} & 64.10\textsubscript{(60.34, 67.84)} & 59.41\textsubscript{(55.70, 63.10)} & 58.28\textsubscript{(52.76, 63.56)} & 33.57\textsubscript{(27.12, 39.83)} \\
THYB\_S\_JS01 & 72 & \textbf{56.33}\textsubscript{(49.82, 62.97)} & 49.38\textsubscript{(45.18, 53.56)} & 54.34\textsubscript{(50.59, 58.03)} & 49.09\textsubscript{(42.61, 55.51)} & 28.27\textsubscript{(23.18, 33.88)} \\
THYB\_S\_NM02 & 66 & \textbf{57.71}\textsubscript{(52.20, 63.08)} & 52.72\textsubscript{(47.87, 57.77)} & 52.81\textsubscript{(47.81, 57.29)} & 46.24\textsubscript{(40.41, 52.18)} & 36.74\textsubscript{(30.84, 42.37)} \\
THYB\_S\_JL04 & 54 & 43.92\textsubscript{(34.91, 52.81)} & \textbf{56.22}\textsubscript{(50.23, 62.17)} & 55.83\textsubscript{(50.74, 61.29)} & 40.85\textsubscript{(32.13, 49.16)} & 42.49\textsubscript{(34.52, 50.97)} \\
THYB\_S\_SH06 & 43 & 49.75\textsubscript{(41.79, 57.64)} & \textbf{51.33}\textsubscript{(45.78, 56.54)} & 45.09\textsubscript{(37.42, 51.92)} & 46.30\textsubscript{(37.76, 54.02)} & 22.32\textsubscript{(15.88, 29.32)} \\
THYB\_S\_BJ09 & 41 & 54.80\textsubscript{(46.03, 63.29)} & \textbf{55.54}\textsubscript{(49.59, 61.66)} & 55.00\textsubscript{(50.40, 59.85)} & 39.65\textsubscript{(31.89, 47.97)} & 52.67\textsubscript{(44.89, 60.73)} \\
THYB\_S\_SX04 & 41 & \textbf{61.37}\textsubscript{(55.03, 67.56)} & 56.50\textsubscript{(50.56, 62.42)} & 48.79\textsubscript{(42.56, 54.12)} & 44.04\textsubscript{(34.60, 52.15)} & 42.36\textsubscript{(33.90, 51.28)} \\
THYB\_S\_SC06 & 29 & 38.63\textsubscript{(29.91, 48.19)} & \textbf{69.92}\textsubscript{(62.86, 76.12)} & 58.20\textsubscript{(49.69, 65.60)} & 32.00\textsubscript{(25.56, 38.77)} & 59.45\textsubscript{(50.88, 67.48)} \\
THYB\_S\_XJ01 & 25 & \textbf{68.57}\textsubscript{(58.62, 77.48)} & 55.07\textsubscript{(48.48, 61.59)} & 57.03\textsubscript{(50.98, 62.37)} & 60.51\textsubscript{(51.17, 68.75)} & 14.00\textsubscript{(7.83, 21.61)} \\
THYB\_S\_YN01 & 25 & 55.83\textsubscript{(43.67, 66.82)} & \textbf{58.55}\textsubscript{(51.86, 65.06)} & 56.48\textsubscript{(48.32, 64.21)} & 46.59\textsubscript{(35.73, 57.93)} & 29.32\textsubscript{(19.14, 40.82)} \\
THYB\_S\_SD13 & 22 & 69.21\textsubscript{(57.84, 80.57)} & \textbf{72.83}\textsubscript{(63.53, 80.86)} & 63.22\textsubscript{(55.37, 71.01)} & 41.87\textsubscript{(31.62, 52.02)} & 46.04\textsubscript{(34.90, 56.40)} \\
THYB\_S\_GX01 & 19 & \textbf{64.80}\textsubscript{(52.35, 75.87)} & 53.73\textsubscript{(46.83, 61.09)} & 60.04\textsubscript{(51.68, 67.03)} & 61.17\textsubscript{(48.99, 71.80)} & 24.77\textsubscript{(12.11, 38.80)} \\
THYB\_S\_ZJ29 & 16 & \textbf{85.21}\textsubscript{(82.65, 87.63)} & 52.40\textsubscript{(43.77, 61.90)} & 67.08\textsubscript{(59.58, 74.31)} & 47.06\textsubscript{(32.00, 61.91)} & 3.30\textsubscript{(0.48, 7.60)} \\
THYB\_S\_HB07 & 8 & 31.68\textsubscript{(9.84, 58.05)} & \textbf{69.62}\textsubscript{(56.11, 82.55)} & 52.10\textsubscript{(37.92, 62.81)} & 19.55\textsubscript{(0.08, 43.94)} & 29.09\textsubscript{(16.83, 41.56)} \\
THYB\_S\_FJ01 & 2 & \textbf{67.14}\textsubscript{(64.04, 70.23)} & 44.00\textsubscript{(38.25, 49.75)} & 49.95\textsubscript{(43.08, 56.82)} & 45.18\textsubscript{(32.48, 57.89)} & 39.20\textsubscript{(29.38, 49.01)} \\
THYB\_S\_SD14 & 2 & 3.13\textsubscript{(2.55, 3.71)} & 46.73\textsubscript{(36.29, 57.18)} & \textbf{58.38}\textsubscript{(54.22, 62.54)} & 44.78\textsubscript{(31.59, 57.97)} & 37.57\textsubscript{(31.23, 43.91)} \\
\bottomrule
\end{tabular}
\end{table}
\clearpage


\begin{table}[p]
\centering
\caption{Per-centre 95\% Hausdorff distance (HD95, mm) for thyroid gland segmentation on the NHC-MISD-TUS external test set. Values are reported as point estimates with 95\% confidence intervals. Bold indicates the best result in each row.}
\label{tab:gland_hd95}
\footnotesize
\setlength{\tabcolsep}{4pt}
\begin{tabular}{llrrrrr}
\toprule
\multirow{2}{*}{Center} & \multirow{2}{*}{$N$} & \multicolumn{1}{c}{ThyroidXAgent} & \multicolumn{1}{c}{MedSAM2} & \multicolumn{1}{c}{MedSegX} & \multicolumn{1}{c}{TransUNet} & \multicolumn{1}{c}{UltraFedFM} \\
\cmidrule(lr){3-3} \cmidrule(lr){4-4} \cmidrule(lr){5-5} \cmidrule(lr){6-6} \cmidrule(lr){7-7}
 & & HD95 (mm) $\downarrow$ & HD95 (mm) $\downarrow$ & HD95 (mm) $\downarrow$ & HD95 (mm) $\downarrow$ & HD95 (mm) $\downarrow$ \\
\midrule
\textit{Overall} & 8,721 & 35.83\textsubscript{(35.14, 36.49)} & 65.71\textsubscript{(65.07, 66.38)} & 45.87\textsubscript{(45.36, 46.37)} & 40.65\textsubscript{(39.95, 41.35)} & 58.17\textsubscript{(57.35, 58.99)} \\
\midrule
THYB\_S\_ZJ24 & 1,174 & \textbf{14.74}\textsubscript{(13.90, 15.63)} & 75.31\textsubscript{(73.88, 76.74)} & 64.28\textsubscript{(62.83, 65.85)} & 22.39\textsubscript{(21.43, 23.43)} & 53.71\textsubscript{(50.40, 56.90)} \\
THYB\_S\_EN04 & 949 & \textbf{30.01}\textsubscript{(28.39, 31.72)} & 73.59\textsubscript{(71.54, 75.87)} & 42.31\textsubscript{(40.95, 43.81)} & 34.27\textsubscript{(32.43, 36.13)} & 63.80\textsubscript{(61.88, 65.55)} \\
THYB\_S\_BJ01 & 876 & \textbf{36.67}\textsubscript{(34.70, 38.76)} & 55.12\textsubscript{(53.38, 56.96)} & 41.33\textsubscript{(40.11, 42.83)} & 40.85\textsubscript{(38.94, 43.05)} & 60.27\textsubscript{(57.54, 63.41)} \\
THYB\_S\_SH01 & 832 & 50.89\textsubscript{(48.25, 53.54)} & 55.08\textsubscript{(52.59, 57.51)} & \textbf{39.63}\textsubscript{(38.18, 41.11)} & 51.90\textsubscript{(49.43, 54.61)} & 51.25\textsubscript{(49.37, 53.41)} \\
THYB\_S\_ZJ05 & 694 & \textbf{37.22}\textsubscript{(34.87, 39.57)} & 70.81\textsubscript{(68.16, 73.22)} & 49.93\textsubscript{(48.19, 51.64)} & 40.86\textsubscript{(38.46, 43.25)} & 64.30\textsubscript{(61.86, 67.00)} \\
THYB\_S\_SH05 & 669 & \textbf{36.52}\textsubscript{(34.23, 39.01)} & 58.39\textsubscript{(56.18, 60.42)} & 43.68\textsubscript{(42.20, 45.23)} & 44.98\textsubscript{(42.65, 47.48)} & 67.24\textsubscript{(64.77, 69.87)} \\
THYB\_S\_NX01 & 582 & 39.66\textsubscript{(37.22, 42.54)} & 82.07\textsubscript{(79.47, 84.56)} & 51.93\textsubscript{(49.68, 54.07)} & \textbf{39.06}\textsubscript{(36.74, 41.62)} & 57.51\textsubscript{(54.52, 60.33)} \\
THYB\_S\_ZJ06 & 460 & \textbf{43.86}\textsubscript{(40.98, 47.06)} & 53.93\textsubscript{(51.41, 56.72)} & 44.67\textsubscript{(42.75, 46.81)} & 48.71\textsubscript{(45.68, 51.91)} & 62.98\textsubscript{(60.17, 65.91)} \\
THYB\_S\_QX07 & 289 & \textbf{30.43}\textsubscript{(27.06, 33.67)} & 84.32\textsubscript{(81.13, 87.54)} & 37.11\textsubscript{(35.21, 39.31)} & 34.16\textsubscript{(30.75, 37.53)} & 58.07\textsubscript{(53.90, 61.53)} \\
THYB\_S\_AN01 & 243 & 59.89\textsubscript{(54.84, 65.27)} & 39.11\textsubscript{(35.71, 42.69)} & \textbf{36.76}\textsubscript{(34.09, 39.70)} & 57.02\textsubscript{(52.09, 61.87)} & 42.73\textsubscript{(39.24, 46.33)} \\
THYB\_S\_CQ03 & 238 & \textbf{36.80}\textsubscript{(33.17, 40.64)} & 67.69\textsubscript{(64.19, 71.21)} & 39.40\textsubscript{(36.37, 42.33)} & 49.95\textsubscript{(45.43, 54.40)} & 63.03\textsubscript{(58.68, 67.50)} \\
THYB\_S\_GZ02 & 233 & \textbf{37.00}\textsubscript{(33.22, 40.65)} & 50.49\textsubscript{(47.63, 53.64)} & 39.41\textsubscript{(37.30, 41.67)} & 40.19\textsubscript{(35.91, 44.49)} & 57.61\textsubscript{(53.08, 62.19)} \\
THYB\_S\_JX06 & 215 & 42.43\textsubscript{(38.02, 47.27)} & 54.75\textsubscript{(50.60, 59.00)} & \textbf{37.59}\textsubscript{(35.11, 40.19)} & 48.81\textsubscript{(44.42, 53.58)} & 52.03\textsubscript{(47.75, 56.31)} \\
THYB\_S\_JS02 & 190 & \textbf{49.35}\textsubscript{(44.19, 55.17)} & 52.35\textsubscript{(48.11, 56.54)} & 53.87\textsubscript{(50.49, 57.35)} & 56.88\textsubscript{(51.46, 62.82)} & 52.07\textsubscript{(47.01, 57.13)} \\
THYB\_S\_YN05 & 137 & 37.57\textsubscript{(32.03, 43.12)} & 73.49\textsubscript{(68.84, 77.86)} & \textbf{35.07}\textsubscript{(31.77, 38.50)} & 39.92\textsubscript{(34.22, 46.49)} & 50.39\textsubscript{(41.86, 58.71)} \\
THYB\_S\_EN02 & 120 & \textbf{37.62}\textsubscript{(32.53, 42.62)} & 100.8\textsubscript{(95.9, 105.8)} & 45.67\textsubscript{(42.65, 48.58)} & 44.64\textsubscript{(39.27, 50.38)} & 54.04\textsubscript{(48.46, 59.63)} \\
THYB\_S\_AH04 & 101 & \textbf{25.82}\textsubscript{(20.55, 31.83)} & 70.36\textsubscript{(64.77, 75.45)} & 33.52\textsubscript{(29.78, 37.56)} & 35.53\textsubscript{(29.34, 42.79)} & 52.02\textsubscript{(44.23, 60.59)} \\
THYB\_S\_GS03 & 94 & 44.87\textsubscript{(37.84, 51.56)} & 61.22\textsubscript{(54.74, 67.47)} & \textbf{44.51}\textsubscript{(40.21, 49.25)} & 48.16\textsubscript{(40.83, 55.61)} & 52.29\textsubscript{(45.50, 59.13)} \\
THYB\_S\_FJ03 & 85 & \textbf{27.04}\textsubscript{(21.49, 33.08)} & 83.26\textsubscript{(77.91, 88.32)} & 48.29\textsubscript{(43.55, 53.23)} & 40.21\textsubscript{(34.71, 46.20)} & 57.81\textsubscript{(51.88, 63.89)} \\
THYB\_S\_SD12 & 75 & \textbf{23.80}\textsubscript{(19.40, 28.64)} & 57.62\textsubscript{(52.08, 62.79)} & 36.03\textsubscript{(32.04, 39.97)} & 35.69\textsubscript{(30.37, 41.96)} & 51.62\textsubscript{(43.29, 59.76)} \\
THYB\_S\_JS01 & 72 & \textbf{40.00}\textsubscript{(32.29, 47.62)} & 79.39\textsubscript{(72.78, 85.69)} & 45.34\textsubscript{(40.63, 50.53)} & 40.73\textsubscript{(33.99, 47.88)} & 64.43\textsubscript{(55.62, 73.62)} \\
THYB\_S\_NM02 & 66 & 45.94\textsubscript{(38.55, 53.94)} & 63.04\textsubscript{(56.60, 69.43)} & \textbf{44.14}\textsubscript{(39.41, 49.57)} & 51.02\textsubscript{(42.31, 59.30)} & 62.80\textsubscript{(54.21, 70.52)} \\
THYB\_S\_JL04 & 54 & 53.99\textsubscript{(42.73, 65.60)} & 67.66\textsubscript{(59.00, 76.78)} & \textbf{39.38}\textsubscript{(34.57, 44.49)} & 50.55\textsubscript{(39.81, 61.07)} & 57.53\textsubscript{(46.74, 67.69)} \\
THYB\_S\_SH06 & 43 & 45.77\textsubscript{(34.95, 57.09)} & 63.35\textsubscript{(55.55, 71.99)} & 45.14\textsubscript{(38.78, 51.03)} & \textbf{41.91}\textsubscript{(30.31, 53.31)} & 63.26\textsubscript{(50.44, 76.60)} \\
THYB\_S\_BJ09 & 41 & \textbf{36.63}\textsubscript{(27.94, 47.32)} & 63.20\textsubscript{(55.22, 70.78)} & 42.93\textsubscript{(36.63, 49.75)} & 41.72\textsubscript{(31.74, 52.87)} & 44.38\textsubscript{(34.57, 53.98)} \\
THYB\_S\_SX04 & 41 & \textbf{38.80}\textsubscript{(31.32, 47.14)} & 66.23\textsubscript{(57.04, 75.13)} & 48.62\textsubscript{(44.02, 53.56)} & 53.31\textsubscript{(42.58, 66.15)} & 68.42\textsubscript{(55.93, 82.59)} \\
THYB\_S\_SC06 & 29 & 51.99\textsubscript{(36.85, 66.32)} & 45.67\textsubscript{(35.65, 56.18)} & \textbf{38.52}\textsubscript{(30.69, 46.92)} & 56.62\textsubscript{(43.06, 70.35)} & 43.15\textsubscript{(33.11, 53.56)} \\
THYB\_S\_XJ01 & 25 & \textbf{29.06}\textsubscript{(21.16, 38.25)} & 54.51\textsubscript{(46.58, 63.05)} & 44.52\textsubscript{(37.28, 52.45)} & 44.96\textsubscript{(34.41, 56.69)} & 52.44\textsubscript{(36.21, 68.45)} \\
THYB\_S\_YN01 & 25 & \textbf{35.59}\textsubscript{(22.18, 54.31)} & 62.96\textsubscript{(52.87, 73.71)} & 38.48\textsubscript{(32.35, 44.59)} & 54.30\textsubscript{(37.89, 71.41)} & 66.12\textsubscript{(49.33, 83.02)} \\
THYB\_S\_SD13 & 22 & \textbf{23.75}\textsubscript{(12.47, 36.80)} & 43.51\textsubscript{(30.68, 58.21)} & 36.56\textsubscript{(28.65, 44.41)} & 52.16\textsubscript{(38.57, 65.81)} & 66.39\textsubscript{(52.20, 84.55)} \\
THYB\_S\_GX01 & 19 & 33.19\textsubscript{(22.74, 43.33)} & 78.95\textsubscript{(67.13, 88.94)} & \textbf{29.49}\textsubscript{(22.85, 36.30)} & 38.98\textsubscript{(26.03, 54.00)} & 38.20\textsubscript{(20.84, 60.52)} \\
THYB\_S\_ZJ29 & 16 & \textbf{12.12}\textsubscript{(6.47, 20.65)} & 65.31\textsubscript{(54.30, 76.11)} & 22.34\textsubscript{(17.11, 28.07)} & 38.17\textsubscript{(22.88, 54.66)} & 45.26\textsubscript{(21.23, 69.87)} \\
THYB\_S\_HB07 & 8 & \textbf{30.47}\textsubscript{(7.26, 58.20)} & 44.74\textsubscript{(23.64, 68.97)} & 32.40\textsubscript{(23.17, 45.39)} & 39.33\textsubscript{(7.94, 74.12)} & 80.20\textsubscript{(66.41, 92.63)} \\
THYB\_S\_FJ01 & 2 & 58.89\textsubscript{(23.60, 94.19)} & 75.44\textsubscript{(66.29, 84.60)} & \textbf{33.92}\textsubscript{(27.53, 40.31)} & 47.04\textsubscript{(34.00, 60.08)} & 49.86\textsubscript{(33.24, 66.48)} \\
THYB\_S\_SD14 & 2 & 78.47\textsubscript{(78.16, 78.77)} & 98.68\textsubscript{(81.39, 115.97)} & 55.96\textsubscript{(43.32, 68.59)} & \textbf{34.84}\textsubscript{(19.69, 50.00)} & 40.44\textsubscript{(36.06, 44.82)} \\
\bottomrule
\end{tabular}
\end{table}
\clearpage


\begin{table}[p]
\centering
\caption{Per-centre Dice similarity coefficient (\%) for thyroid nodule segmentation on the NHC-MISD-TUS external test set. Values are reported as point estimates with 95\% confidence intervals. Bold indicates the best result in each row.}
\label{tab:nodule_dice}
\footnotesize
\setlength{\tabcolsep}{4pt}
\begin{tabular}{llrrrrr}
\toprule
\multirow{2}{*}{Center} & \multirow{2}{*}{$N$} & \multicolumn{1}{c}{ThyroidXAgent} & \multicolumn{1}{c}{MedSAM2} & \multicolumn{1}{c}{MedSegX} & \multicolumn{1}{c}{TransUNet} & \multicolumn{1}{c}{UltraFedFM} \\
\cmidrule(lr){3-3} \cmidrule(lr){4-4} \cmidrule(lr){5-5} \cmidrule(lr){6-6} \cmidrule(lr){7-7}
 & & Dice (\%) $\uparrow$ & Dice (\%) $\uparrow$ & Dice (\%) $\uparrow$ & Dice (\%) $\uparrow$ & Dice (\%) $\uparrow$ \\
\midrule
\textit{Overall} & 6,384 & 82.31\textsubscript{(81.78, 82.83)} & 23.20\textsubscript{(22.58, 23.86)} & 28.82\textsubscript{(28.11, 29.53)} & 72.99\textsubscript{(72.29, 73.68)} & 58.87\textsubscript{(58.04, 59.68)} \\
\midrule
THYB\_S\_EN04 & 946 & \textbf{82.92}\textsubscript{(81.77, 84.00)} & 15.54\textsubscript{(14.29, 16.78)} & 22.25\textsubscript{(20.68, 23.66)} & 74.43\textsubscript{(72.85, 75.95)} & 56.67\textsubscript{(54.69, 58.62)} \\
THYB\_S\_SH01 & 834 & \textbf{88.78}\textsubscript{(88.01, 89.61)} & 39.13\textsubscript{(37.10, 41.01)} & 46.16\textsubscript{(44.13, 48.29)} & 84.48\textsubscript{(83.49, 85.58)} & 71.64\textsubscript{(69.99, 73.33)} \\
THYB\_S\_ZJ05 & 690 & \textbf{82.45}\textsubscript{(81.21, 83.64)} & 15.47\textsubscript{(13.92, 17.01)} & 16.32\textsubscript{(14.69, 17.96)} & 67.34\textsubscript{(65.14, 69.41)} & 55.66\textsubscript{(52.98, 58.07)} \\
THYB\_S\_SH05 & 665 & \textbf{85.53}\textsubscript{(84.17, 86.86)} & 20.11\textsubscript{(18.43, 21.73)} & 29.48\textsubscript{(27.34, 31.47)} & 75.56\textsubscript{(73.63, 77.41)} & 60.40\textsubscript{(57.93, 62.74)} \\
THYB\_S\_NX01 & 557 & \textbf{77.41}\textsubscript{(75.39, 79.40)} & 16.92\textsubscript{(15.32, 18.56)} & 23.51\textsubscript{(21.59, 25.42)} & 68.50\textsubscript{(66.26, 70.91)} & 47.56\textsubscript{(44.76, 50.35)} \\
THYB\_S\_ZJ06 & 460 & \textbf{84.52}\textsubscript{(82.82, 86.14)} & 25.61\textsubscript{(23.25, 27.95)} & 28.94\textsubscript{(26.49, 31.44)} & 75.20\textsubscript{(73.03, 77.38)} & 61.29\textsubscript{(58.63, 63.89)} \\
THYB\_S\_ZJ24 & 321 & \textbf{45.53}\textsubscript{(41.66, 48.73)} & 5.47\textsubscript{(4.61, 6.45)} & 4.17\textsubscript{(3.42, 5.08)} & 20.51\textsubscript{(17.52, 23.78)} & 14.92\textsubscript{(12.15, 17.89)} \\
THYB\_S\_QX07 & 280 & \textbf{85.51}\textsubscript{(83.52, 87.35)} & 22.27\textsubscript{(19.68, 24.88)} & 28.59\textsubscript{(25.55, 31.77)} & 77.89\textsubscript{(74.95, 80.54)} & 68.19\textsubscript{(65.16, 71.49)} \\
THYB\_S\_AN01 & 243 & \textbf{91.60}\textsubscript{(89.81, 92.94)} & 53.29\textsubscript{(49.81, 56.91)} & 61.68\textsubscript{(58.00, 65.00)} & 88.34\textsubscript{(86.28, 90.14)} & 78.19\textsubscript{(75.20, 81.15)} \\
THYB\_S\_CQ03 & 186 & \textbf{82.72}\textsubscript{(79.86, 85.39)} & 18.16\textsubscript{(15.15, 21.46)} & 27.06\textsubscript{(23.17, 31.10)} & 72.57\textsubscript{(68.81, 76.37)} & 55.25\textsubscript{(50.23, 60.01)} \\
THYB\_S\_JS02 & 184 & \textbf{87.67}\textsubscript{(85.58, 89.59)} & 34.39\textsubscript{(30.61, 38.22)} & 35.40\textsubscript{(30.90, 39.94)} & 82.96\textsubscript{(80.24, 85.39)} & 71.70\textsubscript{(67.73, 75.66)} \\
THYB\_S\_JX06 & 167 & \textbf{87.50}\textsubscript{(84.81, 89.82)} & 35.52\textsubscript{(31.67, 39.88)} & 45.33\textsubscript{(40.68, 50.25)} & 83.78\textsubscript{(80.89, 86.39)} & 73.76\textsubscript{(70.40, 77.15)} \\
THYB\_S\_BJ01 & 142 & \textbf{69.73}\textsubscript{(64.15, 74.65)} & 16.56\textsubscript{(13.36, 19.94)} & 21.83\textsubscript{(17.92, 25.72)} & 58.88\textsubscript{(53.18, 64.45)} & 46.92\textsubscript{(41.12, 52.74)} \\
THYB\_S\_EN02 & 107 & \textbf{84.38}\textsubscript{(81.60, 86.98)} & 11.99\textsubscript{(9.13, 15.33)} & 19.36\textsubscript{(15.55, 23.76)} & 73.29\textsubscript{(69.08, 77.46)} & 50.71\textsubscript{(44.52, 57.00)} \\
THYB\_S\_GZ02 & 92 & \textbf{89.21}\textsubscript{(87.74, 90.46)} & 27.23\textsubscript{(22.31, 32.19)} & 30.27\textsubscript{(24.62, 36.27)} & 84.00\textsubscript{(81.49, 86.01)} & 73.15\textsubscript{(68.35, 77.38)} \\
THYB\_S\_GS03 & 91 & \textbf{85.66}\textsubscript{(81.56, 88.93)} & 28.49\textsubscript{(22.48, 34.72)} & 35.22\textsubscript{(28.72, 41.50)} & 78.93\textsubscript{(74.06, 83.12)} & 64.75\textsubscript{(58.31, 70.89)} \\
THYB\_S\_JS01 & 66 & \textbf{85.58}\textsubscript{(81.87, 88.54)} & 13.65\textsubscript{(10.66, 16.77)} & 18.99\textsubscript{(15.28, 23.17)} & 76.18\textsubscript{(70.10, 81.21)} & 58.55\textsubscript{(50.67, 65.99)} \\
THYB\_S\_FJ03 & 64 & \textbf{79.73}\textsubscript{(74.98, 83.85)} & 10.49\textsubscript{(7.58, 13.92)} & 14.62\textsubscript{(10.01, 20.30)} & 64.17\textsubscript{(55.49, 71.69)} & 50.24\textsubscript{(41.88, 59.00)} \\
THYB\_S\_NM02 & 50 & \textbf{83.34}\textsubscript{(75.81, 89.18)} & 21.78\textsubscript{(15.16, 29.35)} & 27.11\textsubscript{(20.14, 34.56)} & 76.31\textsubscript{(67.18, 83.75)} & 61.40\textsubscript{(52.34, 69.28)} \\
THYB\_S\_SH06 & 31 & \textbf{70.85}\textsubscript{(59.24, 79.88)} & 4.60\textsubscript{(3.47, 5.86)} & 8.28\textsubscript{(5.59, 11.59)} & 55.16\textsubscript{(39.98, 68.39)} & 46.54\textsubscript{(33.59, 59.77)} \\
THYB\_S\_AH04 & 30 & \textbf{91.71}\textsubscript{(89.82, 93.31)} & 39.78\textsubscript{(30.88, 48.49)} & 48.60\textsubscript{(37.90, 58.61)} & 84.52\textsubscript{(77.22, 89.78)} & 79.09\textsubscript{(70.70, 84.79)} \\
THYB\_S\_SC06 & 29 & \textbf{90.16}\textsubscript{(88.07, 91.95)} & 44.20\textsubscript{(32.95, 53.60)} & 49.38\textsubscript{(38.45, 58.61)} & 88.29\textsubscript{(85.51, 90.65)} & 76.57\textsubscript{(67.27, 84.35)} \\
THYB\_S\_SX04 & 27 & \textbf{89.20}\textsubscript{(86.29, 91.70)} & 17.84\textsubscript{(12.89, 22.84)} & 16.51\textsubscript{(11.20, 23.50)} & 87.47\textsubscript{(84.52, 89.90)} & 68.19\textsubscript{(57.72, 77.31)} \\
THYB\_S\_BJ09 & 22 & \textbf{86.20}\textsubscript{(81.55, 90.25)} & 30.06\textsubscript{(19.75, 40.78)} & 34.98\textsubscript{(22.71, 47.67)} & 81.62\textsubscript{(74.86, 87.53)} & 70.13\textsubscript{(63.36, 77.01)} \\
THYB\_S\_YN01 & 22 & \textbf{87.24}\textsubscript{(79.41, 92.13)} & 23.20\textsubscript{(13.43, 34.02)} & 28.13\textsubscript{(18.31, 39.88)} & 80.58\textsubscript{(70.28, 87.54)} & 44.07\textsubscript{(29.99, 57.81)} \\
THYB\_S\_SD13 & 18 & \textbf{92.74}\textsubscript{(91.15, 94.06)} & 52.54\textsubscript{(39.34, 64.38)} & 61.58\textsubscript{(48.47, 73.53)} & 90.83\textsubscript{(88.24, 92.78)} & 77.31\textsubscript{(70.93, 82.80)} \\
THYB\_S\_JL04 & 16 & \textbf{81.27}\textsubscript{(67.59, 93.13)} & 36.60\textsubscript{(22.73, 50.94)} & 48.61\textsubscript{(31.53, 65.36)} & 76.87\textsubscript{(62.21, 89.36)} & 70.49\textsubscript{(53.00, 86.26)} \\
THYB\_S\_SD12 & 16 & \textbf{83.71}\textsubscript{(69.71, 93.91)} & 53.71\textsubscript{(39.32, 67.16)} & 62.86\textsubscript{(47.03, 78.03)} & 79.72\textsubscript{(62.56, 93.40)} & 74.86\textsubscript{(58.02, 88.45)} \\
THYB\_S\_YN05 & 12 & \textbf{91.81}\textsubscript{(88.17, 94.81)} & 59.02\textsubscript{(42.58, 71.76)} & 70.91\textsubscript{(51.90, 85.39)} & 91.01\textsubscript{(87.57, 93.88)} & 68.71\textsubscript{(47.83, 85.32)} \\
THYB\_S\_XJ01 & 8 & \textbf{82.71}\textsubscript{(73.69, 90.34)} & 13.18\textsubscript{(4.03, 24.95)} & 14.95\textsubscript{(4.71, 29.10)} & 56.32\textsubscript{(33.11, 79.31)} & 45.53\textsubscript{(20.28, 71.40)} \\
THYB\_S\_GX01 & 4 & \textbf{85.84}\textsubscript{(83.15, 88.54)} & 20.06\textsubscript{(9.70, 30.41)} & 38.09\textsubscript{(22.52, 49.68)} & 84.85\textsubscript{(78.42, 89.87)} & 70.97\textsubscript{(60.58, 83.12)} \\
THYB\_S\_FJ01 & 2 & 79.74\textsubscript{(78.18, 81.31)} & 6.47\textsubscript{(4.46, 8.48)} & 12.32\textsubscript{(8.26, 16.37)} & \textbf{81.94}\textsubscript{(77.73, 86.14)} & 46.04\textsubscript{(36.60, 55.49)} \\
THYB\_S\_HB07 & 2 & 21.28\textsubscript{(18.58, 23.99)} & 1.90\textsubscript{(1.48, 2.31)} & 1.86\textsubscript{(1.61, 2.11)} & \textbf{23.49}\textsubscript{(0.00, 46.97)} & 17.79\textsubscript{(11.76, 23.82)} \\
\bottomrule
\end{tabular}
\end{table}
\clearpage


\begin{table}[p]
\centering
\caption{Per-centre 95\% Hausdorff distance (HD95, mm) for thyroid nodule segmentation on the NHC-MISD-TUS external test set. Values are reported as point estimates with 95\% confidence intervals. Bold indicates the best result in each row.}
\label{tab:nodule_hd95}
\footnotesize
\setlength{\tabcolsep}{4pt}
\begin{tabular}{llrrrrr}
\toprule
\multirow{2}{*}{Center} & \multirow{2}{*}{$N$} & \multicolumn{1}{c}{ThyroidXAgent} & \multicolumn{1}{c}{MedSAM2} & \multicolumn{1}{c}{MedSegX} & \multicolumn{1}{c}{TransUNet} & \multicolumn{1}{c}{UltraFedFM} \\
\cmidrule(lr){3-3} \cmidrule(lr){4-4} \cmidrule(lr){5-5} \cmidrule(lr){6-6} \cmidrule(lr){7-7}
 & & HD95 (mm) $\downarrow$ & HD95 (mm) $\downarrow$ & HD95 (mm) $\downarrow$ & HD95 (mm) $\downarrow$ & HD95 (mm) $\downarrow$ \\
\midrule
\textit{Overall} & 6,384 & 9.41\textsubscript{(8.89, 9.93)} & 101.2\textsubscript{(100.2, 102.0)} & 86.96\textsubscript{(86.08, 87.84)} & 10.62\textsubscript{(10.14, 11.12)} & 27.04\textsubscript{(26.28, 27.75)} \\
\midrule
THYB\_S\_EN04 & 946 & \textbf{6.95}\textsubscript{(6.05, 7.91)} & 113.0\textsubscript{(111.0, 114.9)} & 91.23\textsubscript{(89.41, 92.99)} & 9.27\textsubscript{(8.31, 10.37)} & 33.50\textsubscript{(31.47, 35.76)} \\
THYB\_S\_SH01 & 834 & \textbf{6.41}\textsubscript{(5.50, 7.36)} & 79.94\textsubscript{(77.11, 82.81)} & 65.42\textsubscript{(63.21, 67.68)} & 8.32\textsubscript{(7.35, 9.27)} & 21.31\textsubscript{(19.64, 22.92)} \\
THYB\_S\_ZJ05 & 690 & \textbf{5.00}\textsubscript{(4.31, 5.76)} & 108.5\textsubscript{(106.1, 110.6)} & 106.8\textsubscript{(104.6, 109.0)} & 7.73\textsubscript{(6.88, 8.61)} & 21.22\textsubscript{(19.59, 23.13)} \\
THYB\_S\_SH05 & 665 & \textbf{7.60}\textsubscript{(6.34, 8.89)} & 101.8\textsubscript{(99.4, 104.1)} & 81.02\textsubscript{(78.74, 83.33)} & 10.17\textsubscript{(8.84, 11.51)} & 28.23\textsubscript{(25.93, 30.37)} \\
THYB\_S\_NX01 & 557 & \textbf{12.95}\textsubscript{(10.96, 14.86)} & 112.7\textsubscript{(110.1, 115.1)} & 92.51\textsubscript{(89.99, 94.97)} & 13.02\textsubscript{(11.18, 14.80)} & 30.17\textsubscript{(27.56, 32.82)} \\
THYB\_S\_ZJ06 & 460 & \textbf{7.62}\textsubscript{(6.22, 9.24)} & 92.99\textsubscript{(90.07, 95.87)} & 88.02\textsubscript{(85.09, 91.28)} & 10.88\textsubscript{(9.34, 12.66)} & 26.09\textsubscript{(23.54, 28.75)} \\
THYB\_S\_ZJ24 & 321 & 45.03\textsubscript{(39.74, 50.67)} & 142.9\textsubscript{(139.7, 145.9)} & 140.9\textsubscript{(137.7, 143.8)} & \textbf{30.45}\textsubscript{(26.09, 34.77)} & 45.71\textsubscript{(40.11, 51.74)} \\
THYB\_S\_QX07 & 280 & \textbf{6.69}\textsubscript{(5.08, 8.55)} & 110.2\textsubscript{(106.4, 113.9)} & 85.79\textsubscript{(82.50, 89.13)} & 8.11\textsubscript{(6.54, 10.16)} & 22.53\textsubscript{(19.71, 25.68)} \\
THYB\_S\_AN01 & 243 & \textbf{4.70}\textsubscript{(3.51, 6.11)} & 60.30\textsubscript{(55.67, 64.85)} & 49.32\textsubscript{(45.45, 53.68)} & 7.19\textsubscript{(5.65, 9.00)} & 16.62\textsubscript{(14.15, 19.21)} \\
THYB\_S\_CQ03 & 186 & \textbf{8.81}\textsubscript{(5.97, 12.38)} & 105.0\textsubscript{(100.0, 109.4)} & 81.35\textsubscript{(77.26, 85.23)} & 12.51\textsubscript{(9.28, 16.26)} & 33.57\textsubscript{(28.12, 38.94)} \\
THYB\_S\_JS02 & 184 & \textbf{6.68}\textsubscript{(4.75, 8.90)} & 80.55\textsubscript{(75.49, 85.50)} & 84.81\textsubscript{(78.76, 90.71)} & 8.78\textsubscript{(6.74, 10.90)} & 24.40\textsubscript{(20.23, 28.77)} \\
THYB\_S\_JX06 & 167 & \textbf{7.25}\textsubscript{(5.01, 9.88)} & 82.29\textsubscript{(76.67, 87.66)} & 63.70\textsubscript{(58.48, 68.68)} & 9.18\textsubscript{(6.44, 12.50)} & 20.12\textsubscript{(16.76, 23.56)} \\
THYB\_S\_BJ01 & 142 & 20.72\textsubscript{(15.43, 26.10)} & 101.8\textsubscript{(96.3, 107.7)} & 92.30\textsubscript{(87.28, 97.40)} & \textbf{20.10}\textsubscript{(15.47, 25.01)} & 34.87\textsubscript{(29.55, 40.13)} \\
THYB\_S\_EN02 & 107 & \textbf{4.49}\textsubscript{(3.07, 6.59)} & 128.9\textsubscript{(123.9, 133.4)} & 92.21\textsubscript{(87.64, 96.59)} & 8.45\textsubscript{(6.20, 11.49)} & 29.38\textsubscript{(24.47, 34.60)} \\
THYB\_S\_GZ02 & 92 & \textbf{4.51}\textsubscript{(3.43, 5.77)} & 84.00\textsubscript{(78.53, 89.58)} & 81.50\textsubscript{(74.63, 88.26)} & 7.53\textsubscript{(5.65, 9.73)} & 17.57\textsubscript{(13.41, 22.16)} \\
THYB\_S\_GS03 & 91 & \textbf{7.80}\textsubscript{(4.49, 11.81)} & 89.69\textsubscript{(81.70, 97.29)} & 75.21\textsubscript{(68.30, 82.38)} & 8.70\textsubscript{(5.54, 12.57)} & 24.98\textsubscript{(19.70, 31.13)} \\
THYB\_S\_JS01 & 66 & \textbf{5.69}\textsubscript{(3.49, 8.71)} & 111.8\textsubscript{(106.2, 117.2)} & 94.67\textsubscript{(89.47, 99.84)} & 9.84\textsubscript{(6.27, 14.06)} & 27.35\textsubscript{(19.68, 35.58)} \\
THYB\_S\_FJ03 & 64 & \textbf{6.99}\textsubscript{(3.86, 10.72)} & 115.9\textsubscript{(109.6, 122.4)} & 107.7\textsubscript{(100.0, 115.2)} & 8.87\textsubscript{(5.71, 12.92)} & 22.36\textsubscript{(16.53, 28.70)} \\
THYB\_S\_NM02 & 50 & 6.75\textsubscript{(3.95, 10.41)} & 95.61\textsubscript{(85.83, 104.48)} & 82.78\textsubscript{(75.37, 90.43)} & \textbf{6.49}\textsubscript{(4.04, 9.79)} & 21.73\textsubscript{(16.43, 27.52)} \\
THYB\_S\_SH06 & 31 & 16.01\textsubscript{(6.79, 25.90)} & 115.0\textsubscript{(110.3, 120.7)} & 94.85\textsubscript{(88.68, 100.91)} & \textbf{7.08}\textsubscript{(3.20, 12.90)} & 28.90\textsubscript{(17.04, 42.21)} \\
THYB\_S\_AH04 & 30 & \textbf{3.09}\textsubscript{(2.14, 4.32)} & 81.37\textsubscript{(70.67, 92.67)} & 66.34\textsubscript{(54.18, 78.80)} & 9.13\textsubscript{(4.11, 16.74)} & 18.64\textsubscript{(11.53, 27.65)} \\
THYB\_S\_SC06 & 29 & \textbf{4.72}\textsubscript{(2.94, 6.80)} & 71.23\textsubscript{(58.43, 85.23)} & 61.37\textsubscript{(51.00, 72.75)} & 5.60\textsubscript{(3.63, 7.90)} & 20.14\textsubscript{(10.14, 31.55)} \\
THYB\_S\_SX04 & 27 & \textbf{3.54}\textsubscript{(2.39, 4.80)} & 106.5\textsubscript{(98.0, 114.6)} & 108.8\textsubscript{(98.9, 118.0)} & 4.05\textsubscript{(3.12, 5.09)} & 25.37\textsubscript{(14.31, 39.60)} \\
THYB\_S\_BJ09 & 22 & \textbf{5.56}\textsubscript{(3.29, 8.33)} & 85.51\textsubscript{(71.97, 98.12)} & 77.91\textsubscript{(63.39, 92.08)} & 7.06\textsubscript{(3.88, 11.19)} & 18.84\textsubscript{(11.45, 26.82)} \\
THYB\_S\_YN01 & 22 & \textbf{4.61}\textsubscript{(2.12, 8.49)} & 96.72\textsubscript{(83.10, 109.97)} & 73.60\textsubscript{(64.25, 81.65)} & 4.98\textsubscript{(3.42, 6.90)} & 40.06\textsubscript{(27.32, 54.75)} \\
THYB\_S\_SD13 & 18 & \textbf{3.32}\textsubscript{(2.22, 4.57)} & 61.73\textsubscript{(47.66, 76.68)} & 45.00\textsubscript{(32.04, 58.04)} & 4.51\textsubscript{(2.74, 6.60)} & 21.77\textsubscript{(15.46, 28.42)} \\
THYB\_S\_JL04 & 16 & 14.24\textsubscript{(2.17, 31.43)} & 88.05\textsubscript{(70.97, 104.04)} & 58.06\textsubscript{(41.17, 75.76)} & \textbf{12.51}\textsubscript{(3.43, 23.13)} & 16.36\textsubscript{(6.05, 28.72)} \\
THYB\_S\_SD12 & 16 & 9.76\textsubscript{(3.28, 17.71)} & 57.71\textsubscript{(38.35, 78.90)} & 50.35\textsubscript{(31.21, 71.00)} & \textbf{8.08}\textsubscript{(2.22, 15.94)} & 18.16\textsubscript{(7.70, 30.26)} \\
THYB\_S\_YN05 & 12 & \textbf{3.22}\textsubscript{(2.05, 4.34)} & 61.17\textsubscript{(44.75, 80.17)} & 38.97\textsubscript{(24.66, 57.04)} & 4.86\textsubscript{(2.25, 9.08)} & 13.71\textsubscript{(6.45, 21.71)} \\
THYB\_S\_XJ01 & 8 & \textbf{6.99}\textsubscript{(1.71, 16.39)} & 111.3\textsubscript{(92.8, 131.6)} & 99.22\textsubscript{(85.22, 115.90)} & 20.73\textsubscript{(4.35, 42.34)} & 24.55\textsubscript{(5.64, 47.07)} \\
THYB\_S\_GX01 & 4 & 5.57\textsubscript{(3.00, 8.47)} & 101.0\textsubscript{(81.2, 123.7)} & 78.07\textsubscript{(56.31, 92.33)} & \textbf{5.12}\textsubscript{(2.62, 8.46)} & 15.24\textsubscript{(6.58, 25.12)} \\
THYB\_S\_FJ01 & 2 & 8.32\textsubscript{(4.00, 12.65)} & 112.3\textsubscript{(104.0, 120.7)} & 86.49\textsubscript{(73.93, 99.04)} & \textbf{4.41}\textsubscript{(2.83, 6.00)} & 56.71\textsubscript{(26.29, 87.13)} \\
THYB\_S\_HB07 & 2 & 30.80\textsubscript{(18.03, 43.57)} & 109.0\textsubscript{(98.6, 119.4)} & 104.4\textsubscript{(95.2, 113.5)} & \textbf{4.70}\textsubscript{(0.00, 9.39)} & 19.30\textsubscript{(8.60, 30.00)} \\
\bottomrule
\end{tabular}
\end{table}
\clearpage


\begin{table}[p]
\centering
\caption{Per-centre AUROC for benign versus malignant thyroid nodule classification on the NHC-MISD-TUS external test set. Values are reported as point estimates with 95\% confidence intervals. Bold indicates the best result in each row.}
\label{tab:binary_auroc}
\footnotesize
\setlength{\tabcolsep}{4pt}
\begin{tabular}{llrrrr}
\toprule
\multirow{2}{*}{Center} & \multirow{2}{*}{$N$} & \multicolumn{1}{c}{ThyroidXAgent} & \multicolumn{1}{c}{BiomedCLIP} & \multicolumn{1}{c}{MedSigLIP} & \multicolumn{1}{c}{UltraFedFM} \\
\cmidrule(lr){3-3} \cmidrule(lr){4-4} \cmidrule(lr){5-5} \cmidrule(lr){6-6}
 & & AUROC $\uparrow$ & AUROC $\uparrow$ & AUROC $\uparrow$ & AUROC $\uparrow$ \\
\midrule
\textit{Overall} & 4,999 & 0.819\textsubscript{(0.808, 0.831)} & 0.434\textsubscript{(0.418, 0.450)} & 0.520\textsubscript{(0.503, 0.535)} & 0.436\textsubscript{(0.420, 0.452)} \\
\midrule
THYB\_S\_SH01 & 792 & \textbf{0.896}\textsubscript{(0.873, 0.916)} & 0.327\textsubscript{(0.290, 0.368)} & 0.552\textsubscript{(0.517, 0.594)} & 0.331\textsubscript{(0.295, 0.368)} \\
THYB\_S\_ZJ05 & 647 & \textbf{0.811}\textsubscript{(0.773, 0.844)} & 0.526\textsubscript{(0.480, 0.576)} & 0.586\textsubscript{(0.546, 0.627)} & 0.494\textsubscript{(0.455, 0.533)} \\
THYB\_S\_EN04 & 626 & \textbf{0.846}\textsubscript{(0.808, 0.876)} & 0.345\textsubscript{(0.289, 0.404)} & 0.538\textsubscript{(0.472, 0.597)} & 0.528\textsubscript{(0.476, 0.584)} \\
THYB\_S\_SH05 & 539 & \textbf{0.719}\textsubscript{(0.672, 0.758)} & 0.331\textsubscript{(0.280, 0.381)} & 0.543\textsubscript{(0.492, 0.595)} & 0.370\textsubscript{(0.324, 0.423)} \\
THYB\_S\_ZJ06 & 452 & \textbf{0.811}\textsubscript{(0.768, 0.849)} & 0.468\textsubscript{(0.412, 0.520)} & 0.594\textsubscript{(0.541, 0.651)} & 0.539\textsubscript{(0.486, 0.592)} \\
THYB\_S\_QX07 & 279 & \textbf{0.870}\textsubscript{(0.814, 0.917)} & 0.457\textsubscript{(0.381, 0.537)} & 0.630\textsubscript{(0.534, 0.711)} & 0.428\textsubscript{(0.338, 0.520)} \\
THYB\_S\_ZJ24 & 269 & 0.590\textsubscript{(0.346, 0.832)} & 0.412\textsubscript{(0.270, 0.544)} & \textbf{0.673}\textsubscript{(0.493, 0.823)} & 0.463\textsubscript{(0.255, 0.678)} \\
THYB\_S\_AN01 & 236 & \textbf{0.862}\textsubscript{(0.784, 0.923)} & 0.241\textsubscript{(0.165, 0.319)} & 0.601\textsubscript{(0.519, 0.687)} & 0.292\textsubscript{(0.202, 0.390)} \\
THYB\_S\_CQ03 & 172 & \textbf{0.784}\textsubscript{(0.716, 0.849)} & 0.452\textsubscript{(0.366, 0.550)} & 0.521\textsubscript{(0.431, 0.608)} & 0.440\textsubscript{(0.350, 0.532)} \\
THYB\_S\_JS02 & 171 & \textbf{0.746}\textsubscript{(0.667, 0.815)} & 0.504\textsubscript{(0.417, 0.596)} & 0.675\textsubscript{(0.599, 0.750)} & 0.415\textsubscript{(0.328, 0.501)} \\
THYB\_S\_JX06 & 153 & \textbf{0.903}\textsubscript{(0.853, 0.951)} & 0.399\textsubscript{(0.315, 0.487)} & 0.616\textsubscript{(0.524, 0.702)} & 0.480\textsubscript{(0.390, 0.575)} \\
THYB\_S\_EN02 & 99 & \textbf{0.851}\textsubscript{(0.741, 0.941)} & 0.383\textsubscript{(0.248, 0.543)} & 0.749\textsubscript{(0.602, 0.878)} & 0.329\textsubscript{(0.166, 0.488)} \\
THYB\_S\_BJ01 & 93 & \textbf{0.749}\textsubscript{(0.635, 0.851)} & 0.245\textsubscript{(0.154, 0.345)} & 0.400\textsubscript{(0.235, 0.562)} & 0.460\textsubscript{(0.319, 0.609)} \\
THYB\_S\_GZ02 & 90 & \textbf{0.693}\textsubscript{(0.578, 0.795)} & 0.325\textsubscript{(0.198, 0.471)} & 0.499\textsubscript{(0.374, 0.614)} & 0.366\textsubscript{(0.223, 0.530)} \\
THYB\_S\_GS03 & 84 & \textbf{0.838}\textsubscript{(0.735, 0.924)} & 0.244\textsubscript{(0.125, 0.373)} & 0.443\textsubscript{(0.311, 0.566)} & 0.470\textsubscript{(0.328, 0.607)} \\
THYB\_S\_JS01 & 59 & \textbf{0.781}\textsubscript{(0.491, 1.000)} & 0.509\textsubscript{(0.069, 0.947)} & 0.281\textsubscript{(0.035, 0.552)} & 0.070\textsubscript{(0.017, 0.500)} \\
THYB\_S\_NM02 & 49 & 0.653\textsubscript{(0.497, 0.800)} & 0.408\textsubscript{(0.253, 0.567)} & \textbf{0.719}\textsubscript{(0.549, 0.875)} & 0.393\textsubscript{(0.202, 0.595)} \\
THYB\_S\_SX04 & 27 & \textbf{0.860}\textsubscript{(0.500, 1.000)} & 0.640\textsubscript{(0.231, 0.962)} & 0.500\textsubscript{(0.077, 0.924)} & 0.620\textsubscript{(0.400, 0.846)} \\
THYB\_S\_AH04 & 24 & \textbf{0.844}\textsubscript{(0.663, 0.979)} & 0.617\textsubscript{(0.368, 0.838)} & 0.586\textsubscript{(0.305, 0.838)} & 0.266\textsubscript{(0.076, 0.523)} \\
THYB\_S\_SC06 & 22 & \textbf{0.876}\textsubscript{(0.702, 1.000)} & 0.256\textsubscript{(0.050, 0.484)} & 0.612\textsubscript{(0.350, 0.839)} & 0.141\textsubscript{(0.009, 0.325)} \\
THYB\_S\_YN01 & 21 & \textbf{0.853}\textsubscript{(0.618, 1.000)} & 0.353\textsubscript{(0.000, 0.778)} & 0.500\textsubscript{(0.105, 0.895)} & 0.176\textsubscript{(0.000, 0.400)} \\
THYB\_S\_BJ09 & 20 & 0.586\textsubscript{(0.319, 0.849)} & 0.505\textsubscript{(0.213, 0.798)} & \textbf{0.636}\textsubscript{(0.341, 0.885)} & 0.404\textsubscript{(0.150, 0.687)} \\
THYB\_S\_SD13 & 18 & \textbf{0.875}\textsubscript{(0.636, 1.000)} & 0.536\textsubscript{(0.125, 1.000)} & 0.714\textsubscript{(0.415, 0.956)} & 0.304\textsubscript{(0.000, 0.623)} \\
THYB\_S\_SD12 & 15 & \textbf{0.929}\textsubscript{(0.500, 1.000)} & 0.143\textsubscript{(0.000, 0.500)} & 0.571\textsubscript{(0.356, 0.786)} & 0.071\textsubscript{(0.000, 0.500)} \\
THYB\_S\_JL04 & 12 & \textbf{0.407}\textsubscript{(0.000, 1.000)} & 0.074\textsubscript{(0.000, 0.446)} & 0.148\textsubscript{(0.000, 0.500)} & 0.296\textsubscript{(0.000, 0.727)} \\
THYB\_S\_YN05 & 12 & \textbf{0.500}\textsubscript{(0.500, 0.500)} & 0.500\textsubscript{(0.500, 0.500)} & 0.500\textsubscript{(0.500, 0.500)} & 0.500\textsubscript{(0.500, 0.500)} \\
THYB\_S\_XJ01 & 5 & \textbf{1.000}\textsubscript{(0.500, 1.000)} & 0.667\textsubscript{(0.000, 1.000)} & 0.333\textsubscript{(0.000, 1.000)} & 0.833\textsubscript{(0.250, 1.000)} \\
THYB\_S\_GX01 & 4 & \textbf{0.500}\textsubscript{(0.500, 0.500)} & 0.500\textsubscript{(0.500, 0.500)} & 0.500\textsubscript{(0.500, 0.500)} & 0.500\textsubscript{(0.500, 0.500)} \\
THYB\_S\_SH06 & 4 & \textbf{1.000}\textsubscript{(0.500, 1.000)} & 0.000\textsubscript{(0.000, 0.500)} & 0.000\textsubscript{(0.000, 0.500)} & 0.500\textsubscript{(0.000, 1.000)} \\
THYB\_S\_FJ01 & 2 & \textbf{0.500}\textsubscript{(0.500, 0.500)} & 0.500\textsubscript{(0.500, 0.500)} & 0.500\textsubscript{(0.500, 0.500)} & 0.500\textsubscript{(0.500, 0.500)} \\
THYB\_S\_FJ03 & 2 & \textbf{0.500}\textsubscript{(0.500, 0.500)} & 0.500\textsubscript{(0.500, 0.500)} & 0.500\textsubscript{(0.500, 0.500)} & 0.500\textsubscript{(0.500, 0.500)} \\
THYB\_S\_NX01 & 1 & \textbf{0.500}\textsubscript{(0.500, 0.500)} & 0.500\textsubscript{(0.500, 0.500)} & 0.500\textsubscript{(0.500, 0.500)} & 0.500\textsubscript{(0.500, 0.500)} \\
THYB\_S\_HB07 & 0 & -- & -- & -- & -- \\
THYB\_S\_SD14 & 0 & -- & -- & -- & -- \\
THYB\_S\_ZJ29 & 0 & -- & -- & -- & -- \\
\bottomrule
\end{tabular}
\end{table}
\clearpage


\begin{table}[p]
\centering
\caption{Per-centre AUPRC for benign versus malignant thyroid nodule classification on the NHC-MISD-TUS external test set. Values are reported as point estimates with 95\% confidence intervals. Bold indicates the best result in each row.}
\label{tab:binary_auprc}
\footnotesize
\setlength{\tabcolsep}{4pt}
\begin{tabular}{llrrrr}
\toprule
\multirow{2}{*}{Center} & \multirow{2}{*}{$N$} & \multicolumn{1}{c}{ThyroidXAgent} & \multicolumn{1}{c}{BiomedCLIP} & \multicolumn{1}{c}{MedSigLIP} & \multicolumn{1}{c}{UltraFedFM} \\
\cmidrule(lr){3-3} \cmidrule(lr){4-4} \cmidrule(lr){5-5} \cmidrule(lr){6-6}
 & & AUPRC $\uparrow$ & AUPRC $\uparrow$ & AUPRC $\uparrow$ & AUPRC $\uparrow$ \\
\midrule
\textit{Overall} & 4,999 & 0.823\textsubscript{(0.807, 0.838)} & 0.450\textsubscript{(0.433, 0.467)} & 0.519\textsubscript{(0.499, 0.539)} & 0.465\textsubscript{(0.447, 0.483)} \\
\midrule
THYB\_S\_SH01 & 792 & \textbf{0.859}\textsubscript{(0.817, 0.893)} & 0.323\textsubscript{(0.290, 0.356)} & 0.450\textsubscript{(0.404, 0.504)} & 0.340\textsubscript{(0.304, 0.380)} \\
THYB\_S\_ZJ05 & 647 & \textbf{0.746}\textsubscript{(0.690, 0.798)} & 0.460\textsubscript{(0.412, 0.518)} & 0.519\textsubscript{(0.469, 0.580)} & 0.445\textsubscript{(0.393, 0.504)} \\
THYB\_S\_EN04 & 626 & \textbf{0.969}\textsubscript{(0.957, 0.978)} & 0.796\textsubscript{(0.758, 0.840)} & 0.851\textsubscript{(0.812, 0.888)} & 0.871\textsubscript{(0.837, 0.904)} \\
THYB\_S\_SH05 & 539 & \textbf{0.832}\textsubscript{(0.787, 0.865)} & 0.539\textsubscript{(0.496, 0.589)} & 0.680\textsubscript{(0.630, 0.734)} & 0.555\textsubscript{(0.509, 0.610)} \\
THYB\_S\_ZJ06 & 452 & \textbf{0.811}\textsubscript{(0.752, 0.864)} & 0.502\textsubscript{(0.442, 0.573)} & 0.597\textsubscript{(0.526, 0.670)} & 0.554\textsubscript{(0.490, 0.626)} \\
THYB\_S\_QX07 & 279 & \textbf{0.575}\textsubscript{(0.432, 0.706)} & 0.163\textsubscript{(0.107, 0.235)} & 0.270\textsubscript{(0.173, 0.385)} & 0.158\textsubscript{(0.107, 0.245)} \\
THYB\_S\_ZJ24 & 269 & \textbf{0.164}\textsubscript{(0.026, 0.423)} & 0.032\textsubscript{(0.014, 0.054)} & 0.068\textsubscript{(0.027, 0.137)} & 0.051\textsubscript{(0.017, 0.137)} \\
THYB\_S\_AN01 & 236 & \textbf{0.588}\textsubscript{(0.430, 0.756)} & 0.110\textsubscript{(0.082, 0.147)} & 0.195\textsubscript{(0.141, 0.261)} & 0.121\textsubscript{(0.087, 0.172)} \\
THYB\_S\_CQ03 & 172 & \textbf{0.876}\textsubscript{(0.819, 0.922)} & 0.603\textsubscript{(0.510, 0.706)} & 0.637\textsubscript{(0.544, 0.735)} & 0.559\textsubscript{(0.474, 0.651)} \\
THYB\_S\_JS02 & 171 & \textbf{0.734}\textsubscript{(0.629, 0.822)} & 0.502\textsubscript{(0.414, 0.621)} & 0.649\textsubscript{(0.544, 0.750)} & 0.459\textsubscript{(0.369, 0.561)} \\
THYB\_S\_JX06 & 153 & \textbf{0.897}\textsubscript{(0.839, 0.946)} & 0.351\textsubscript{(0.276, 0.438)} & 0.545\textsubscript{(0.422, 0.671)} & 0.409\textsubscript{(0.314, 0.522)} \\
THYB\_S\_EN02 & 99 & \textbf{0.969}\textsubscript{(0.937, 0.992)} & 0.812\textsubscript{(0.710, 0.908)} & 0.939\textsubscript{(0.888, 0.983)} & 0.775\textsubscript{(0.669, 0.899)} \\
THYB\_S\_BJ01 & 93 & \textbf{0.930}\textsubscript{(0.880, 0.970)} & 0.692\textsubscript{(0.583, 0.825)} & 0.743\textsubscript{(0.634, 0.858)} & 0.764\textsubscript{(0.661, 0.884)} \\
THYB\_S\_GZ02 & 90 & \textbf{0.850}\textsubscript{(0.760, 0.921)} & 0.558\textsubscript{(0.460, 0.695)} & 0.702\textsubscript{(0.581, 0.825)} & 0.569\textsubscript{(0.457, 0.697)} \\
THYB\_S\_GS03 & 84 & \textbf{0.921}\textsubscript{(0.847, 0.970)} & 0.578\textsubscript{(0.459, 0.706)} & 0.692\textsubscript{(0.559, 0.816)} & 0.674\textsubscript{(0.553, 0.808)} \\
THYB\_S\_JS01 & 59 & \textbf{0.990}\textsubscript{(0.966, 1.000)} & 0.964\textsubscript{(0.885, 1.000)} & 0.950\textsubscript{(0.862, 1.000)} & 0.921\textsubscript{(0.808, 1.000)} \\
THYB\_S\_NM02 & 49 & \textbf{0.810}\textsubscript{(0.662, 0.920)} & 0.594\textsubscript{(0.441, 0.781)} & 0.777\textsubscript{(0.611, 0.941)} & 0.562\textsubscript{(0.420, 0.729)} \\
THYB\_S\_SX04 & 27 & \textbf{0.988}\textsubscript{(0.959, 1.000)} & 0.960\textsubscript{(0.873, 1.000)} & 0.931\textsubscript{(0.792, 1.000)} & 0.965\textsubscript{(0.895, 1.000)} \\
THYB\_S\_AH04 & 24 & \textbf{0.746}\textsubscript{(0.430, 0.967)} & 0.463\textsubscript{(0.222, 0.800)} & 0.547\textsubscript{(0.203, 0.819)} & 0.258\textsubscript{(0.130, 0.450)} \\
THYB\_S\_SC06 & 22 & \textbf{0.873}\textsubscript{(0.653, 1.000)} & 0.390\textsubscript{(0.229, 0.609)} & 0.704\textsubscript{(0.409, 0.898)} & 0.356\textsubscript{(0.209, 0.573)} \\
THYB\_S\_YN01 & 21 & \textbf{0.968}\textsubscript{(0.901, 1.000)} & 0.748\textsubscript{(0.537, 0.990)} & 0.832\textsubscript{(0.615, 0.993)} & 0.737\textsubscript{(0.498, 0.947)} \\
THYB\_S\_BJ09 & 20 & 0.683\textsubscript{(0.415, 0.910)} & 0.687\textsubscript{(0.406, 0.895)} & \textbf{0.699}\textsubscript{(0.431, 0.925)} & 0.559\textsubscript{(0.292, 0.806)} \\
THYB\_S\_SD13 & 18 & \textbf{0.567}\textsubscript{(0.200, 1.000)} & 0.446\textsubscript{(0.059, 1.000)} & 0.415\textsubscript{(0.111, 0.900)} & 0.196\textsubscript{(0.056, 0.415)} \\
THYB\_S\_SD12 & 15 & \textbf{0.500}\textsubscript{(0.000, 1.000)} & 0.077\textsubscript{(0.000, 0.177)} & 0.143\textsubscript{(0.000, 0.365)} & 0.071\textsubscript{(0.000, 0.150)} \\
THYB\_S\_JL04 & 12 & \textbf{0.491}\textsubscript{(0.083, 1.000)} & 0.187\textsubscript{(0.083, 0.379)} & 0.199\textsubscript{(0.083, 0.404)} & 0.241\textsubscript{(0.083, 0.610)} \\
THYB\_S\_YN05 & 12 & \textbf{0.000}\textsubscript{(0.000, 0.000)} & 0.000\textsubscript{(0.000, 0.000)} & 0.000\textsubscript{(0.000, 0.000)} & 0.000\textsubscript{(0.000, 0.000)} \\
THYB\_S\_XJ01 & 5 & \textbf{1.000}\textsubscript{(1.000, 1.000)} & 0.806\textsubscript{(0.333, 1.000)} & 0.589\textsubscript{(0.200, 1.000)} & 0.917\textsubscript{(0.417, 1.000)} \\
THYB\_S\_GX01 & 4 & \textbf{1.000}\textsubscript{(1.000, 1.000)} & 1.000\textsubscript{(1.000, 1.000)} & 1.000\textsubscript{(1.000, 1.000)} & 1.000\textsubscript{(1.000, 1.000)} \\
THYB\_S\_SH06 & 4 & \textbf{1.000}\textsubscript{(0.000, 1.000)} & 0.417\textsubscript{(0.000, 1.000)} & 0.417\textsubscript{(0.000, 1.000)} & 0.583\textsubscript{(0.000, 1.000)} \\
THYB\_S\_FJ01 & 2 & \textbf{1.000}\textsubscript{(1.000, 1.000)} & 1.000\textsubscript{(1.000, 1.000)} & 1.000\textsubscript{(1.000, 1.000)} & 1.000\textsubscript{(1.000, 1.000)} \\
THYB\_S\_FJ03 & 2 & \textbf{1.000}\textsubscript{(1.000, 1.000)} & 1.000\textsubscript{(1.000, 1.000)} & 1.000\textsubscript{(1.000, 1.000)} & 1.000\textsubscript{(1.000, 1.000)} \\
THYB\_S\_NX01 & 1 & \textbf{1.000}\textsubscript{(1.000, 1.000)} & 1.000\textsubscript{(1.000, 1.000)} & 1.000\textsubscript{(1.000, 1.000)} & 1.000\textsubscript{(1.000, 1.000)} \\
THYB\_S\_HB07 & 0 & -- & -- & -- & -- \\
THYB\_S\_SD14 & 0 & -- & -- & -- & -- \\
THYB\_S\_ZJ29 & 0 & -- & -- & -- & -- \\
\bottomrule
\end{tabular}
\end{table}
\clearpage

\clearpage

\begin{table*}[p]
\centering
\caption{Performance of preprocessing and executor tools in ThyroidXAgent. Held-out test results are reported for preprocessing tools, including image normalization, nodule-presence triage and anatomical-context parsing, and for executor-stage tools, including measurement support, gland localization, lymph-node screening, gland captioning and nodule-feature extraction. Anatomical-context parsing and nodule-feature extraction are additionally reported at the class level. For the binary margin and shape classifiers, AUROC and AUPRC are reported once across the paired class rows. AP, average precision; MAE, mean absolute error; MSE, mean squared error; MAPE, mean absolute percentage error.}
\label{tab:auxiliary_tools}
\scriptsize
\setlength{\tabcolsep}{2.3pt}
\renewcommand{\arraystretch}{1.08}

\resizebox{\textwidth}{!}{%
\begin{tabular}{
|>{\centering\arraybackslash}p{0.100\textwidth}
|>{\raggedright\arraybackslash}p{0.105\textwidth}
|>{\raggedright\arraybackslash}p{0.130\textwidth}
|>{\raggedright\arraybackslash}p{0.140\textwidth}
|>{\centering\arraybackslash}p{0.060\textwidth}
|>{\centering\arraybackslash}p{0.055\textwidth}
|>{\centering\arraybackslash}p{0.055\textwidth}
|>{\centering\arraybackslash}p{0.070\textwidth}
|>{\centering\arraybackslash}p{0.080\textwidth}
|>{\centering\arraybackslash}p{0.075\textwidth}
|>{\centering\arraybackslash}p{0.075\textwidth}
|>{\centering\arraybackslash}p{0.075\textwidth}|}

\hline
\multicolumn{1}{|c|}{\textbf{Agent stage}}
& \multicolumn{11}{c|}{\textbf{Preprocessing tools}} \\
\hline

\multicolumn{1}{|c|}{}
& \multicolumn{1}{c|}{\textbf{Tool group}}
& \multicolumn{2}{c|}{\textbf{Tool}}
& \multicolumn{1}{c|}{\textbf{Train}}
& \multicolumn{1}{c|}{\textbf{Val}}
& \multicolumn{1}{c|}{\textbf{Test}}
& \multicolumn{2}{c|}{\textbf{Primary result}}
& \multicolumn{3}{c|}{\textbf{Secondary result}} \\
\hline

\multirow{10}{=}{\centering Preprocessing}
& Image normalization
& \multicolumn{2}{l|}{Ultrasound ROI cropping}
& 152
& 17
& 79
& \multicolumn{2}{l|}{Dice, 0.9822; IoU, 0.9658}
& \multicolumn{3}{l|}{Precision, 0.9904; recall, 0.9749; pixel accuracy, 0.9829} \\
\cline{2-12}

& Case triage
& \multicolumn{2}{l|}{Nodule-presence detection}
& 82,312
& 10,982
& 16,467
& \multicolumn{2}{l|}{Accuracy, 0.9830; F1, 0.9749}
& \multicolumn{3}{l|}{AUROC, 0.9981; AP, 0.9961; sensitivity, 0.9848; specificity, 0.9821} \\
\cline{2-12}

& \multicolumn{11}{c|}{\textbf{Anatomical context parsing}} \\
\cline{2-12}

& \multicolumn{1}{c|}{\textbf{Tool}}
& \multicolumn{1}{c|}{\textbf{Class}}
& \multicolumn{1}{c|}{\textbf{Train}}
& \multicolumn{1}{c|}{\textbf{Val}}
& \multicolumn{1}{c|}{\textbf{Test}}
& \multicolumn{1}{c|}{\textbf{Total}}
& \multicolumn{1}{c|}{\textbf{Precision}}
& \multicolumn{1}{c|}{\textbf{Recall}}
& \multicolumn{1}{c|}{\textbf{F1}}
& \multicolumn{1}{c|}{\textbf{AUROC}}
& \multicolumn{1}{c|}{\textbf{AUPRC}} \\
\cline{2-12}

& \multirow{6}{*}{\makecell[l]{Thyroid-region\\classification}}
& Left-lobe lateral view
& 780 & 110 & 159 & 1,049
& 0.6643 & 0.5975 & 0.6291 & 0.8521 & 0.7185 \\
\cline{3-12}

& & Right-lobe lateral view
& 946 & 133 & 199 & 1,278
& 0.6460 & 0.7337 & 0.6871 & 0.8327 & 0.7368 \\
\cline{3-12}

& & Bilateral thyroid view
& 120 & 19 & 30 & 169
& 0.8929 & 0.8333 & 0.8621 & 0.9896 & 0.8911 \\
\cline{3-12}

& & Left-lobe transverse view
& 267 & 52 & 41 & 360
& 0.7045 & 0.7561 & 0.7294 & 0.9567 & 0.8198 \\
\cline{3-12}

& & Right-lobe transverse view
& 272 & 61 & 79 & 412
& 0.8088 & 0.6962 & 0.7483 & 0.9553 & 0.8585 \\
\cline{3-12}

& & Neck region
& 267 & 21 & 12 & 300
& 1.0000 & 0.9167 & 0.9565 & 0.9974 & 0.9524 \\
\hline

\multicolumn{1}{|c|}{\textbf{Agent stage}}
& \multicolumn{11}{c|}{\textbf{Executor-stage tools}} \\
\hline

\multicolumn{1}{|c|}{}
& \multicolumn{1}{c|}{\textbf{Tool group}}
& \multicolumn{2}{c|}{\textbf{Tool}}
& \multicolumn{1}{c|}{\textbf{Train}}
& \multicolumn{1}{c|}{\textbf{Val}}
& \multicolumn{1}{c|}{\textbf{Test}}
& \multicolumn{2}{c|}{\textbf{Primary result}}
& \multicolumn{3}{c|}{\textbf{Secondary result}} \\
\hline

\multirow{20}{=}{\centering Executor}
& Measurement support
& \multicolumn{2}{l|}{Spacing prediction}
& 5,288
& 661
& 662
& \multicolumn{2}{l|}{MAE, 0.0131; \(R^2\), 0.8520}
& \multicolumn{3}{l|}{MSE, \(5.66\times10^{-4}\); MAPE, 21.39\%} \\
\cline{2-12}

& Gland localization
& \multicolumn{2}{l|}{Gland segmentation}
& 335
& -
& 90
& \multicolumn{2}{l|}{Dice, 0.8006; IoU, 0.6866}
& \multicolumn{3}{l|}{Precision, 0.8025; recall, 0.8339} \\
\cline{2-12}

& Neck-region screening
& \multicolumn{2}{l|}{Cervical lymph-node detection}
& 251
& -
& 49
& \multicolumn{2}{l|}{Accuracy, 0.7959; F1, 0.7368}
& \multicolumn{3}{l|}{AUROC, 0.8163} \\
\cline{2-12}

& Gland description
& \multicolumn{2}{l|}{Gland captioning}
& 22,782
& 200
& 612
& \multicolumn{2}{l|}{BLEU-4, 0.5898; METEOR, 0.4582}
& \multicolumn{3}{l|}{ROUGE$_L$, 0.7450; CIDEr, 2.7736} \\
\cline{2-12}

& \multicolumn{11}{c|}{\textbf{Nodule feature extraction}} \\
\cline{2-12}

& \multicolumn{1}{c|}{\textbf{Tool family}}
& \multicolumn{1}{c|}{\textbf{Feature classifier}}
& \multicolumn{1}{c|}{\textbf{Class}}
& \multicolumn{1}{c|}{\textbf{Train}}
& \multicolumn{1}{c|}{\textbf{Val}}
& \multicolumn{1}{c|}{\textbf{Test}}
& \multicolumn{1}{c|}{\textbf{Total}}
& \multicolumn{1}{c|}{\textbf{Specificity}}
& \multicolumn{1}{c|}{\textbf{Sensitivity}}
& \multicolumn{1}{c|}{\textbf{AUROC}}
& \multicolumn{1}{c|}{\textbf{AUPRC}} \\
\cline{2-12}

& \multirow{14}{*}{\makecell[l]{Nodule-feature\\classification}}
& \multirow{3}{*}{Composition}
& Cystic
& 1,824 & 227 & 228 & 2,279
& 0.8397 & 0.8553 & 0.9166 & 0.9073 \\
\cline{4-12}

& & & Mixed cystic and solid
& 883 & 108 & 110 & 1,101
& 0.9358 & 0.3636 & 0.8257 & 0.5943 \\
\cline{4-12}

& & & Solid
& 1,391 & 173 & 177 & 1,741
& 0.8018 & 0.7966 & 0.9011 & 0.8033 \\
\cline{3-12}

& & \multirow{4}{*}{Echogenicity}
& Anechoic
& 1,691 & 212 & 213 & 2,116
& 0.8603 & 0.8967 & 0.9397 & 0.9160 \\
\cline{4-12}

& & & Hyperechoic
& 218 & 27 & 27 & 272
& 0.9847 & 0.2593 & 0.8474 & 0.3591 \\
\cline{4-12}

& & & Hypoechoic
& 1,416 & 173 & 173 & 1,762
& 0.8173 & 0.6705 & 0.8391 & 0.7514 \\
\cline{4-12}

& & & Isoechoic
& 580 & 73 & 72 & 725
& 0.9274 & 0.5417 & 0.8804 & 0.5440 \\
\cline{3-12}

& & \multirow{3}{*}{Echogenic foci}
& Macrocalcifications
& 1,165 & 145 & 146 & 1,456
& 0.8107 & 0.4726 & 0.7279 & 0.5601 \\
\cline{4-12}

& & & None
& 2,150 & 270 & 270 & 2,690
& 0.5130 & 0.8037 & 0.7431 & 0.7483 \\
\cline{4-12}

& & & Punctate echogenic foci
& 668 & 83 & 84 & 835
& 0.9423 & 0.1310 & 0.6411 & 0.2615 \\
\cline{3-12}

& & \multirow{2}{*}{Margin}
& Ill-defined
& 1,177 & 147 & 148 & 1,472
& 0.8200 & 0.7432
& \multirow{2}{*}{0.8702}
& \multirow{2}{*}{0.8847} \\
\cline{4-10}

& & & Smooth
& 1,602 & 200 & 200 & 2,002
& 0.7432 & 0.8200 & & \\
\cline{3-12}

& & \multirow{2}{*}{Shape}
& Taller-than-wide
& 81 & 10 & 9 & 100
& 0.9872 & 0.5556
& \multirow{2}{*}{0.9174}
& \multirow{2}{*}{0.9899} \\
\cline{4-10}

& & & Wider-than-tall
& 602 & 79 & 78 & 759
& 0.5556 & 0.9872 & & \\
\hline

\end{tabular}%
}
\end{table*}
\clearpage

\begin{table*}[p]
\centering
\begin{threeparttable}

\caption{Lexical performance of thyroid ultrasound report generation across datasets. BLEU-1 to BLEU-4, METEOR and ROUGE$_L$ are reported on the SMU-HMC, KMVE and ZJH-TS test sets. Values are means $\pm$ the half-width of the bootstrap 95\% percentile confidence interval.}
\label{tab:report_generation_nlg_ci}

\footnotesize
\setlength{\tabcolsep}{5.2pt}
\renewcommand{\arraystretch}{1.04}

\begin{tabular}{lcccccc}
\toprule
\textbf{Model}
& \textbf{BLEU-1}
& \textbf{BLEU-2}
& \textbf{BLEU-3}
& \textbf{BLEU-4}
& \textbf{METEOR}
& \textbf{ROUGE$_L$} \\
\midrule

\rowcolor{gray!20}
\multicolumn{7}{c}{\textbf{SMU-HMC Testset} ($n=400$)} \\

GPT-4o~\cite{openai_gpt4_2024}
& 0.3500$\pm$0.0100
& 0.2535$\pm$0.0080
& 0.1842$\pm$0.0066
& 0.1330$\pm$0.0057
& 0.3247$\pm$0.0047
& 0.3577$\pm$0.0077 \\

GPT-5~\cite{openai2025gpt5systemcard}
& 0.3836$\pm$0.0085
& 0.2749$\pm$0.0069
& 0.1965$\pm$0.0059
& 0.1374$\pm$0.0054
& 0.3254$\pm$0.0037
& 0.3732$\pm$0.0068 \\

Gemini2.5 Pro~\cite{comanici_gemini_2025}
& 0.3702$\pm$0.0094
& 0.2660$\pm$0.0081
& 0.1907$\pm$0.0070
& 0.1373$\pm$0.0063
& 0.3308$\pm$0.0038
& 0.3584$\pm$0.0071 \\

Qwen3.5 Plus~\cite{yang_qwen3_2025}
& 0.4483$\pm$0.0130
& 0.3623$\pm$0.0111
& 0.2959$\pm$0.0095
& 0.2427$\pm$0.0081
& \textbf{0.3628$\pm$0.0040}
& 0.5147$\pm$0.0088 \\

Claude-Sonnet-4.6
& 0.3326$\pm$0.0099
& 0.2543$\pm$0.0080
& 0.1968$\pm$0.0067
& 0.1528$\pm$0.0056
& 0.3417$\pm$0.0035
& 0.3782$\pm$0.0074 \\

MedGemma~\cite{sellergren2025medgemma}
& 0.0457$\pm$0.0072
& 0.0343$\pm$0.0056
& 0.0265$\pm$0.0044
& 0.0207$\pm$0.0035
& 0.1736$\pm$0.0051
& 0.0829$\pm$0.0075 \\

LLaVA-Med~\cite{li_llavamed_2023}
& 0.1670$\pm$0.0077
& 0.0581$\pm$0.0066
& 0.0290$\pm$0.0040
& 0.0159$\pm$0.0024
& 0.1439$\pm$0.0053
& 0.1482$\pm$0.0073 \\

KMVE~\cite{li_ultrasound_2024}
& 0.1743$\pm$0.0131
& 0.1110$\pm$0.0082
& 0.0684$\pm$0.0052
& 0.0398$\pm$0.0035
& 0.1719$\pm$0.0063
& 0.2212$\pm$0.0039 \\

\rowcolor{lightgray}
\textbf{ThyroidXAgent}
& \textbf{0.5924$\pm$0.0143}
& \textbf{0.4806$\pm$0.0139}
& \textbf{0.4006$\pm$0.0137}
& \textbf{0.3381$\pm$0.0135}
& 0.3627$\pm$0.0088
& \textbf{0.5422$\pm$0.0122} \\

\midrule

\rowcolor{gray!20}
\multicolumn{7}{c}{\textbf{KMVE Testset} ($n=492$)} \\

GPT-4o~\cite{openai_gpt4_2024}
& 0.4467$\pm$0.0154
& 0.3423$\pm$0.0143
& 0.2683$\pm$0.0118
& 0.2136$\pm$0.0102
& 0.2799$\pm$0.0116
& 0.3850$\pm$0.0148 \\

GPT-5~\cite{openai2025gpt5systemcard}
& 0.4149$\pm$0.0073
& 0.2992$\pm$0.0066
& 0.2190$\pm$0.0059
& 0.1668$\pm$0.0062
& 0.3042$\pm$0.0070
& 0.4128$\pm$0.0089 \\

Gemini2.5 Pro~\cite{comanici_gemini_2025}
& 0.5065$\pm$0.0117
& 0.3851$\pm$0.0103
& 0.2981$\pm$0.0095
& 0.2394$\pm$0.0092
& 0.2863$\pm$0.0083
& 0.4742$\pm$0.0111 \\

Qwen3.5 Plus~\cite{yang_qwen3_2025}
& 0.5425$\pm$0.0135
& 0.4243$\pm$0.0126
& 0.3415$\pm$0.0121
& 0.2783$\pm$0.0120
& 0.2952$\pm$0.0084
& 0.5066$\pm$0.0118 \\

Claude-Sonnet-4.6
& 0.4331$\pm$0.0119
& 0.3380$\pm$0.0112
& 0.2638$\pm$0.0107
& 0.2089$\pm$0.0106
& 0.3284$\pm$0.0079
& 0.4639$\pm$0.0115 \\

MedGemma~\cite{sellergren2025medgemma}
& 0.0374$\pm$0.0195
& 0.0299$\pm$0.0175
& 0.0252$\pm$0.0161
& 0.0219$\pm$0.0151
& 0.1075$\pm$0.0119
& 0.1862$\pm$0.0205 \\

LLaVA-Med~\cite{li_llavamed_2023}
& 0.2842$\pm$0.0149
& 0.2135$\pm$0.0124
& 0.1595$\pm$0.0099
& 0.1244$\pm$0.0088
& 0.2138$\pm$0.0101
& 0.3939$\pm$0.0127 \\

\rowcolor{lightgray}
\textbf{ThyroidXAgent}
& \textbf{0.6357$\pm$0.0171}
& \textbf{0.5606$\pm$0.0157}
& \textbf{0.5008$\pm$0.0151}
& \textbf{0.4535$\pm$0.0151}
& \textbf{0.3672$\pm$0.0099}
& \textbf{0.5880$\pm$0.0120} \\

\midrule

\rowcolor{gray!20}
\multicolumn{7}{c}{\textbf{ZJH-TS Testset} ($n=150$)} \\

GPT-4o~\cite{openai_gpt4_2024}
& 0.3402$\pm$0.0273
& 0.2447$\pm$0.0206
& 0.1788$\pm$0.0155
& 0.1332$\pm$0.0119
& 0.2495$\pm$0.0163
& 0.3679$\pm$0.0219 \\

GPT-5~\cite{openai2025gpt5systemcard}
& 0.4736$\pm$0.0158
& 0.3499$\pm$0.0130
& 0.2539$\pm$0.0108
& 0.1789$\pm$0.0095
& 0.3332$\pm$0.0060
& 0.4790$\pm$0.0098 \\

Gemini2.5 Pro~\cite{comanici_gemini_2025}
& 0.1932$\pm$0.0112
& 0.1307$\pm$0.0080
& 0.0887$\pm$0.0058
& 0.0601$\pm$0.0045
& 0.2765$\pm$0.0052
& 0.2536$\pm$0.0091 \\

Qwen3.5 Plus~\cite{yang_qwen3_2025}
& 0.4964$\pm$0.0182
& 0.3965$\pm$0.0162
& 0.3196$\pm$0.0145
& 0.2581$\pm$0.0131
& \textbf{0.3508$\pm$0.0076}
& 0.5314$\pm$0.0125 \\

Claude-Sonnet-4.6
& 0.4480$\pm$0.0181
& 0.3518$\pm$0.0157
& 0.2792$\pm$0.0135
& 0.2229$\pm$0.0118
& 0.3455$\pm$0.0073
& 0.4864$\pm$0.0127 \\

MedGemma~\cite{sellergren2025medgemma}
& 0.1236$\pm$0.0220
& 0.0958$\pm$0.0174
& 0.0750$\pm$0.0138
& 0.0591$\pm$0.0110
& 0.2268$\pm$0.0094
& 0.1592$\pm$0.0211 \\

LLaVA-Med~\cite{li_llavamed_2023}
& 0.2318$\pm$0.0194
& 0.1316$\pm$0.0141
& 0.0891$\pm$0.0101
& 0.0606$\pm$0.0074
& 0.1658$\pm$0.0106
& 0.2479$\pm$0.0201 \\

KMVE~\cite{li_ultrasound_2024}
& 0.1682$\pm$0.0146
& 0.1041$\pm$0.0090
& 0.0598$\pm$0.0054
& 0.0289$\pm$0.0038
& 0.1648$\pm$0.0069
& 0.2244$\pm$0.0038 \\

\rowcolor{lightgray}
\textbf{ThyroidXAgent}
& \textbf{0.5051$\pm$0.0228}
& \textbf{0.4137$\pm$0.0206}
& \textbf{0.3447$\pm$0.0188}
& \textbf{0.2909$\pm$0.0174}
& 0.3293$\pm$0.0119
& \textbf{0.5480$\pm$0.0160} \\

\bottomrule
\end{tabular}

\end{threeparttable}
\end{table*}
\clearpage

\begin{table*}[p]
\centering
\begin{threeparttable}

\caption{Clinical semantic performance of thyroid ultrasound report generation across datasets. False discovery rate (FDR), feature accuracy, lesion-level F1 score, completeness, consistency and ThyClinScore are reported on the SMU-HMC, KMVE and ZJH-TS test sets. Values are means $\pm$ the half-width of the bootstrap 95\% percentile confidence interval. Lower FDR indicates better performance.}
\label{tab:report_generation_clinical_ci}

\footnotesize
\setlength{\tabcolsep}{5.2pt}
\renewcommand{\arraystretch}{1.04}

\begin{tabular}{lcccccc}
\toprule
\textbf{Model}
& \textbf{FDR}\,$\downarrow$
& \textbf{Feat Acc}
& \textbf{F1 Score}
& \textbf{Complete.}
& \textbf{Consist.}
& \textbf{ThyClin} \\
\midrule

\rowcolor{gray!20}
\multicolumn{7}{c}{\textbf{SMU-HMC Testset} ($n=400$)} \\

GPT-4o~\cite{openai_gpt4_2024}
& 0.7189$\pm$0.0362
& 0.5555$\pm$0.0351
& 0.2526$\pm$0.0332
& 0.8208$\pm$0.0132
& 0.4023$\pm$0.0182
& 0.4105$\pm$0.0172 \\

GPT-5~\cite{openai2025gpt5systemcard}
& 0.4170$\pm$0.0417
& 0.5644$\pm$0.0351
& 0.4390$\pm$0.0402
& 0.8585$\pm$0.0077
& 0.4980$\pm$0.0166
& 0.4882$\pm$0.0182 \\

Gemini2.5 Pro~\cite{comanici_gemini_2025}
& 0.6797$\pm$0.0388
& 0.5586$\pm$0.0385
& 0.2965$\pm$0.0366
& 0.8789$\pm$0.0107
& 0.4040$\pm$0.0187
& 0.4280$\pm$0.0176 \\

Qwen3.5 Plus~\cite{yang_qwen3_2025}
& 0.7426$\pm$0.0295
& 0.5804$\pm$0.0373
& 0.2656$\pm$0.0290
& 0.9620$\pm$0.0063
& 0.4809$\pm$0.0142
& 0.4883$\pm$0.0137 \\

Claude-Sonnet-4.6
& 0.8040$\pm$0.0305
& 0.5580$\pm$0.0464
& 0.2049$\pm$0.0298
& 0.8786$\pm$0.0108
& 0.3886$\pm$0.0147
& 0.4015$\pm$0.0144 \\

MedGemma~\cite{sellergren2025medgemma}
& 0.8943$\pm$0.0229
& 0.5784$\pm$0.0434
& 0.1027$\pm$0.0202
& 0.9657$\pm$0.0055
& 0.3353$\pm$0.0119
& 0.3697$\pm$0.0131 \\

LLaVA-Med~\cite{li_llavamed_2023}
& \textbf{0.1625$\pm$0.0363}
& 0.4100$\pm$0.2125
& 0.2582$\pm$0.0426
& 0.4576$\pm$0.0227
& 0.1387$\pm$0.0166
& 0.1752$\pm$0.0150 \\

KMVE~\cite{li_ultrasound_2024}
& 0.4813$\pm$0.0488
& 0.5944$\pm$0.1233
& 0.2103$\pm$0.0390
& 0.5995$\pm$0.0085
& 0.2013$\pm$0.0144
& 0.2265$\pm$0.0145 \\

\rowcolor{lightgray}
\textbf{ThyroidXAgent}
& 0.2238$\pm$0.0377
& \textbf{0.6238$\pm$0.0358}
& \textbf{0.5467$\pm$0.0436}
& \textbf{0.9691$\pm$0.0060}
& \textbf{0.5016$\pm$0.0166}
& \textbf{0.5189$\pm$0.0212} \\

\midrule

\rowcolor{gray!20}
\multicolumn{7}{c}{\textbf{KMVE Testset} ($n=492$)} \\

GPT-4o~\cite{openai_gpt4_2024}
& 0.5843$\pm$0.0427
& 0.6608$\pm$0.0691
& 0.2137$\pm$0.0350
& 0.6180$\pm$0.0113
& 0.4275$\pm$0.0231
& 0.3344$\pm$0.0188 \\

GPT-5~\cite{openai2025gpt5systemcard}
& 0.4133$\pm$0.0423
& 0.6967$\pm$0.0492
& 0.3749$\pm$0.0402
& 0.6506$\pm$0.0038
& 0.4504$\pm$0.0189
& 0.3853$\pm$0.0189 \\

Gemini2.5 Pro~\cite{comanici_gemini_2025}
& 0.3211$\pm$0.0407
& 0.5093$\pm$0.0418
& \textbf{0.4648$\pm$0.0401}
& 0.6270$\pm$0.0045
& 0.4135$\pm$0.0213
& 0.3810$\pm$0.0187 \\

Qwen3.5 Plus~\cite{yang_qwen3_2025}
& 0.4553$\pm$0.0422
& 0.6154$\pm$0.0448
& 0.3823$\pm$0.0397
& 0.6525$\pm$0.0050
& 0.4035$\pm$0.0230
& 0.3674$\pm$0.0193 \\

Claude-Sonnet-4.6
& 0.5803$\pm$0.0432
& 0.6605$\pm$0.0578
& 0.2721$\pm$0.0375
& \textbf{0.6707$\pm$0.0043}
& 0.3779$\pm$0.0211
& 0.3362$\pm$0.0182 \\

MedGemma~\cite{sellergren2025medgemma}
& 0.1877$\pm$0.0327
& 0.6820$\pm$0.0463
& 0.2868$\pm$0.0381
& 0.6622$\pm$0.0033
& 0.5172$\pm$0.0240
& 0.3932$\pm$0.0203 \\

LLaVA-Med~\cite{li_llavamed_2023}
& \textbf{0.0000$\pm$0.0000}\tnote{*}
& N/A\tnote{*}
& 0.2541$\pm$0.0386
& 0.6508$\pm$0.0007
& \textbf{0.5669$\pm$0.0225}
& 0.3928$\pm$0.0208 \\

\rowcolor{lightgray}
\textbf{ThyroidXAgent}
& 0.4858$\pm$0.0386
& \textbf{0.7366$\pm$0.0355}
& 0.3623$\pm$0.0358
& 0.6416$\pm$0.0029
& 0.5654$\pm$0.0217
& \textbf{0.4407$\pm$0.0194} \\

\midrule

\rowcolor{gray!20}
\multicolumn{7}{c}{\textbf{ZJH-TS Testset} ($n=150$)} \\

GPT-4o~\cite{openai_gpt4_2024}
& 0.4583$\pm$0.0706
& 0.5510$\pm$0.0468
& 0.2706$\pm$0.0523
& 0.7290$\pm$0.0305
& 0.3415$\pm$0.0294
& 0.3139$\pm$0.0260 \\

GPT-5~\cite{openai2025gpt5systemcard}
& 0.5201$\pm$0.0658
& 0.5263$\pm$0.0487
& 0.3771$\pm$0.0531
& 0.9534$\pm$0.0112
& 0.4735$\pm$0.0243
& 0.4472$\pm$0.0248 \\

Gemini2.5 Pro~\cite{comanici_gemini_2025}
& 0.5450$\pm$0.0689
& 0.5507$\pm$0.0547
& 0.3157$\pm$0.0523
& 0.7859$\pm$0.0183
& 0.3910$\pm$0.0292
& 0.3557$\pm$0.0244 \\

Qwen3.5 Plus~\cite{yang_qwen3_2025}
& 0.5781$\pm$0.0558
& 0.5464$\pm$0.0465
& 0.3902$\pm$0.0485
& 0.9575$\pm$0.0086
& \textbf{0.4767$\pm$0.0219}
& 0.4520$\pm$0.0220 \\

Claude-Sonnet-4.6
& 0.6186$\pm$0.0571
& 0.5748$\pm$0.0455
& 0.3405$\pm$0.0496
& 0.8226$\pm$0.0176
& 0.4298$\pm$0.0233
& 0.3861$\pm$0.0219 \\

MedGemma~\cite{sellergren2025medgemma}
& 0.6537$\pm$0.0644
& 0.5463$\pm$0.0498
& 0.2959$\pm$0.0520
& 0.9437$\pm$0.0106
& 0.3868$\pm$0.0225
& 0.3807$\pm$0.0238 \\

LLaVA-Med~\cite{li_llavamed_2023}
& \textbf{0.2733$\pm$0.0733}
& \textbf{0.6595$\pm$0.1077}
& 0.1241$\pm$0.0454
& 0.6136$\pm$0.0494
& 0.1505$\pm$0.0308
& 0.1812$\pm$0.0275 \\

KMVE~\cite{li_ultrasound_2024}
& 0.7911$\pm$0.0617
& 0.6288$\pm$0.1490
& 0.0445$\pm$0.0253
& 0.6252$\pm$0.0085
& 0.1581$\pm$0.0162
& 0.1650$\pm$0.0124 \\

\rowcolor{lightgray}
\textbf{ThyroidXAgent}
& 0.2833$\pm$0.0600
& 0.5820$\pm$0.0408
& \textbf{0.4889$\pm$0.0570}
& \textbf{0.9641$\pm$0.0090}
& 0.4564$\pm$0.0228
& \textbf{0.4676$\pm$0.0268} \\

\bottomrule
\end{tabular}

\begin{tablenotes}[flushleft]
\footnotesize
\item[*] On the KMVE dataset, LLaVA-Med collapsed and predicted
``no abnormality'' for all test samples. Therefore, Feat Acc is N/A
and FDR is 0 because no positive predictions were made.
\end{tablenotes}

\end{threeparttable}
\end{table*}
\clearpage

\begin{table*}[p]
\centering
\begin{threeparttable}

\caption{Static-pipeline metrics used for report-generation radar plots. Conventional language-generation metrics and clinical semantic metrics are reported for the SMU-HMC, KMVE and ZJH-TS test sets used in Fig.~\ref{fig:thyroidxagent_report_generation}g. Values are means $\pm$ the half-width of the 95\% confidence interval. Lower FDR indicates better performance.}
\label{tab:static_rule_controller_radar_metrics}

\scriptsize
\setlength{\tabcolsep}{4.2pt}
\renewcommand{\arraystretch}{1.10}

\begin{tabular}{lccccccc}
\toprule
\rowcolor{gray!20}
\multicolumn{8}{c}{\textbf{Conventional natural-language generation metrics}} \\
\midrule
\textbf{Dataset}
& \textbf{\(n\)}
& \textbf{BLEU-1}
& \textbf{BLEU-2}
& \textbf{BLEU-3}
& \textbf{BLEU-4}
& \textbf{METEOR}
& \textbf{ROUGE$_L$} \\
\midrule
SMU-HMC
& 400
& 0.4586$\pm$0.0149
& 0.3817$\pm$0.0139
& 0.3271$\pm$0.0136
& 0.2849$\pm$0.0133
& 0.3209$\pm$0.0073
& 0.4725$\pm$0.0115 \\
KMVE
& 492
& 0.3266$\pm$0.0069
& 0.2105$\pm$0.0048
& 0.1306$\pm$0.0034
& 0.0624$\pm$0.0034
& 0.2724$\pm$0.0051
& 0.2750$\pm$0.0045 \\
ZJH-TS
& 150
& 0.4242$\pm$0.0234
& 0.3527$\pm$0.0206
& 0.2986$\pm$0.0186
& 0.2534$\pm$0.0172
& 0.2916$\pm$0.0101
& 0.4994$\pm$0.0140 \\
\bottomrule
\end{tabular}

\vspace{0.8em}

\begin{tabular}{lccccccc}
\toprule
\rowcolor{gray!20}
\multicolumn{8}{c}{\textbf{Clinical semantic metrics}} \\
\midrule
\textbf{Dataset}
& \textbf{\(n\)}
& \textbf{FDR}\,$\downarrow$
& \textbf{Feat Acc}
& \textbf{F1 Score}
& \textbf{Complete.}
& \textbf{Consist.}
& \textbf{ThyClin} \\
\midrule
SMU-HMC
& 400
& 0.3162$\pm$0.0431
& 0.6227$\pm$0.0382
& 0.5060$\pm$0.0453
& 0.7821$\pm$0.0064
& 0.4325$\pm$0.0161
& 0.4293$\pm$0.0192 \\
KMVE
& 492
& 0.5894$\pm$0.0432
& 0.5616$\pm$0.0631
& 0.2644$\pm$0.0375
& 0.7665$\pm$0.0059
& 0.3391$\pm$0.0181
& 0.3346$\pm$0.0174 \\
ZJH-TS
& 150
& 0.4433$\pm$0.0717
& 0.5517$\pm$0.0516
& 0.3894$\pm$0.0603
& 0.8127$\pm$0.0120
& 0.3734$\pm$0.0201
& 0.3648$\pm$0.0239 \\
\bottomrule
\end{tabular}

\end{threeparttable}
\end{table*}
\clearpage

\begin{table*}[p]
\centering
\begin{threeparttable}

\caption{Ablation of tool integration for thyroid ultrasound report generation. Segmentation, classification, captioning and measurement tools were added cumulatively, and performance was evaluated using conventional language-generation metrics. Values are means $\pm$ the half-width of the 95\% confidence interval.}
\label{tab:report_generation_tool_ablation}

\footnotesize
\setlength{\tabcolsep}{5.0pt}
\renewcommand{\arraystretch}{1.08}

\begin{tabular}{lcccccc}
\toprule
\textbf{Configuration}
& \textbf{BLEU-1}
& \textbf{BLEU-2}
& \textbf{BLEU-3}
& \textbf{BLEU-4}
& \textbf{METEOR}
& \textbf{ROUGE$_L$} \\
\midrule

\rowcolor{gray!20}
\multicolumn{7}{c}{\textbf{SMU-HMC Testset} ($n=400$)} \\
Segmentation only
& 0.0897$\pm$0.0095
& 0.0692$\pm$0.0075
& 0.0562$\pm$0.0063
& 0.0459$\pm$0.0053
& 0.1521$\pm$0.0047
& 0.2603$\pm$0.0092 \\
- Classification
& 0.1877$\pm$0.0169
& 0.1431$\pm$0.0130
& 0.1148$\pm$0.0104
& 0.0929$\pm$0.0085
& 0.1849$\pm$0.0072
& 0.2870$\pm$0.0106 \\
- Captioning
& 0.4503$\pm$0.0163
& 0.3724$\pm$0.0150
& 0.3179$\pm$0.0142
& 0.2764$\pm$0.0139
& 0.3052$\pm$0.0081
& 0.4601$\pm$0.0126 \\
- Measurement (full)
& 0.5924$\pm$0.0143
& 0.4806$\pm$0.0139
& 0.4006$\pm$0.0137
& 0.3381$\pm$0.0135
& 0.3627$\pm$0.0088
& 0.5422$\pm$0.0122 \\

\midrule
\rowcolor{gray!20}
\multicolumn{7}{c}{\textbf{KMVE Testset} ($n=492$)} \\
Segmentation only
& 0.6130$\pm$0.0191
& 0.5456$\pm$0.0186
& 0.4907$\pm$0.0182
& 0.4462$\pm$0.0186
& 0.3658$\pm$0.0103
& 0.5747$\pm$0.0126 \\
- Classification
& 0.6315$\pm$0.0189
& 0.5637$\pm$0.0184
& 0.5081$\pm$0.0180
& 0.4633$\pm$0.0183
& 0.3751$\pm$0.0102
& 0.5799$\pm$0.0127 \\
- Captioning
& 0.6357$\pm$0.0172
& 0.5606$\pm$0.0160
& 0.5008$\pm$0.0152
& 0.4535$\pm$0.0153
& 0.3672$\pm$0.0099
& 0.5880$\pm$0.0122 \\
- Measurement (full)
& 0.6357$\pm$0.0172
& 0.5606$\pm$0.0160
& 0.5008$\pm$0.0152
& 0.4535$\pm$0.0153
& 0.3672$\pm$0.0099
& 0.5880$\pm$0.0122 \\

\midrule
\rowcolor{gray!20}
\multicolumn{7}{c}{\textbf{ZJH-TS Testset} ($n=150$)} \\
Segmentation only
& 0.0964$\pm$0.0150
& 0.0799$\pm$0.0123
& 0.0674$\pm$0.0105
& 0.0567$\pm$0.0091
& 0.1500$\pm$0.0073
& 0.3348$\pm$0.0129 \\
- Classification
& 0.2456$\pm$0.0246
& 0.1990$\pm$0.0197
& 0.1648$\pm$0.0163
& 0.1356$\pm$0.0137
& 0.2053$\pm$0.0105
& 0.4003$\pm$0.0142 \\
- Captioning
& 0.4092$\pm$0.0252
& 0.3409$\pm$0.0220
& 0.2891$\pm$0.0197
& 0.2461$\pm$0.0180
& 0.2848$\pm$0.0115
& 0.4903$\pm$0.0162 \\
- Measurement (full)
& 0.5051$\pm$0.0227
& 0.4137$\pm$0.0205
& 0.3447$\pm$0.0188
& 0.2909$\pm$0.0174
& 0.3293$\pm$0.0118
& 0.5480$\pm$0.0158 \\
\bottomrule
\end{tabular}

\begin{tablenotes}[flushleft]
\footnotesize
\item The KMVE dataset retains only the findings section and does not provide original measurement values. To match the original evaluation protocol, only the generated findings section was evaluated and measurement values were masked; therefore, the captioning and full configurations have identical KMVE scores.
\end{tablenotes}

\end{threeparttable}
\end{table*}
\clearpage

\end{appendices}

\clearpage
\bibliography{ref}

\end{document}